# Applications of Machine Learning to Optimizing Polyolefin Manufacturing

Liu, Y. A., & Sharma, N.

Abstract

This is a preprint version of the chapter of our book -Liu, Y. A., & Sharma, N. (2023). *Integrated Process Modeling, Advanced Control and Data Analytics for Optimizing Polyolefin Manufacturing*. Wiley-VCH GmbH. lease cite the original work [169,170] if referenced.

This chapter covers the applications of machine learning (ML) to optimizing chemical and polymer processes, particularly polyolefin manufacturing. Our goal is to prepare an overview of ML for university students and faculty, and practicing engineers and scientists who are new to the field, and also for those who are knowledgeable, but wish to know the new developments and application literature in chemical and polymer processes.

Section 10.1 presents an introduction, beginning in Section 10.1.1 with the historical developments of artificial intelligence (AI) and machine learning (ML) in chemical process industries (CPIs), and suggests the time for actively adopting AI and ML in CPIs has arrived. Section 10.1.2 continues with three key components of ML applications, namely data, representations and learning, and explains the concepts of supervised learning, semi-supervised learning, unsupervised learning, and reinforcement learning. Section 10.1.3 suggests resources for the readers to get started with ML, including reference books, list of top free online Python training courses, Python reference libraries, and books with ML principles and coding examples. Appendix B of the book gives an introduction to Python for chemical engineers.

Section 10.2 gives an overview of selected ML methods and applications to regression and classification problems. Section 10.2.1 covers supervised learning methods for regression applications. We discuss linear regression, polynomial regression, underfitting, overfitting and regularization, Ridge linear regression and Lasso linear regression, bias-variance tradeoff, and performance evaluation metrics for regression problems. Section 10.2.2 discusses supervised learning methods for classification applications. We cover logistic regression (classification), radial basis function network, and K-means clustering, P-nearest neighbor algorithm, support vector machine classification and regression, decision trees for classification and regression, and performance evaluation metrics for classification problems. Section 10.2.3 covers unsupervised learning for dimensionality reduction, outlier detection and clustering applications. We have already introduced principal component analysis (PCA) in Chapter 9.  In this section, we introduce kernel PCA, K-means clustering, hierarchical clustering, density-based spatial clustering of applications with noise (DBSCAN), and Gaussian mixture model, together with their similarities and differences.

Section 10.3 presents enhanced learning by ensemble methods, including bagging, boosting and stacking, and introduces the popular methods of random forest, AdaBoost, XGBoost, among others.

Section 10.4 discusses enhanced learning by deep neural networks. Section 10.4.1 first reviews the key concepts, parameters and training of a multilayer perceptron (MLP), focusing on its limitations and required changes when applied to deep neural networks (DNNs). We explain the vanishing and exploding gradient problems in the popular gradient descent training algorithm, and discuss techniques to improve the performance of DNNs, such as batch normalization, regularization by weight decay and dropout, and fast optimizer. Section 10.4.2 covers deep learning by recurrent neural networks (RNNs) for modeling time-dependent processes, including backpropagation through time, long short-term memory (LSTM) RNNs, gated recurrent units (GRUs), and bidirectional RNNs. Section 10.4.3 introduces convolutional neural networks (CNNs). Section 10.4.4 presents the transformer neural network that finds growing applications to chemistry, drug discovery and chemical processes. Section 10.5 discusses the general guidelines to choose appropriate ML algorithms for specific applications.

Sections 10.6 presents a workshop to predict the HDPE melt index using random forest and extreme gradient boosting (XGBoost) ensemble learning models. Section 10.7 covers a workshop to predict the HDPE melt index using a deep neural network. Section 10.8 presents a workshop on time-dependent RNNs to predict the HDPE melt index in dynamic operations, using LSTM and GRU RNNs. Section 10.9 presents a workshop for polymer property prediction based on molecular structure both transformer network and convolutional neural networks. Section 10.10 covers a workshop for melt index using automated ML. Section 10.11 presents an example demonstrating the limitations of stand-alone data-based machine learning model, which leads to Chapter 11, focusing on integrating first-principle-based model with a data-based machine learning model to improve both the interpolative and extrapolative accuracies of the resulting model. Section 10.12 gives a detailed bibliography for further studies and software implementation by the readers.

Overall, this chapter represents a comprehensive introduction to the literature on the fundamentals, practice and hands-on workshops of applying ML techniques to chemical and polymer processes, particular polyolefin manufacturing. We are not aware of any published references with a similar coverage.

A growing trend is to integrate science-guided fundamental models with data-based machine learning in a hybrid science-guided machine learning (SGML) approach to modeling chemical processes. We recommend our readers to continue studying Chapter 11 about the hybrid SGML approach and its applications to modeling chemical processes and optimizing polyolefin manufacturing.

## 10.1 Introduction

### 10.1.1 The Time for AI, Particularly Machine Learning, in Chemical Industries Has Finally Arrived

Machine learning (ML) is the most important subfield of artificial intelligence, focusing on the science and art of programming computers so they can learn from data [1, p. 2]. Alan Turing [2] first put forward the concept of AI and explained how computers or machines could learn. In mid-1980s to early 1990's, when chemical engineers began to get involved in AI research and practice, the term "machine learning" did not appear in most textbooks [3,4] and research monographs [5]. The focuses then were on rule-based expert systems [3] and neural networks [4].

Venkatasubramanian [25] gives an excellent perspective of the evolution of AI in chemical engineering. He divides the historical development up to the present into three phases: *phase I- expert system era* (~1983 to ~1995); *phase II –neural network er*a (~1990 to ~2008); and *phase III - data science and deep*

*learning era* (~2005 to present). He attributes the lack of impact of AI during phases I and II to: (1) the problems we were attacking are extremely challenging even today; (2) the lack of powerful computing, storage, communication and programming environments; (3) the lack of sufficient data for ML model development; and (4) the lack of inexpensive and free resources.

It is truly amazing that new developments in ML over the past two decades have touched every aspect of chemical industries. In fact, it would not be possible for us to adequately review the broad ranges of reported and potential applications of ML in chemical industries in this short chapter. We trust that a study of representative published reviews, such as references [15 to 32], arranged in year of publication from 2003 to 2022, should convince the reader that the time for AI, particularly ML, in chemical industries has finally arrived.

In addition to Venkatasubramanian [25], we suggest the reader to study Dobbelaere, et. al [26] who present a SWOT (strengths, weaknesses, opportunities and threats) analysis of ML in chemical engineering; Niket and Liu [31] on a science-guided machine learning (SGML) approach to modeling chemical processes; and Chiang et al. [32] on a holistic view of how the chemical industry is transforming digitally towards AI at scale.

### 10.1.2 Data, Representation and Learning

### 10.1.2.a  Data

The first key component of an ML application is *data*, which could consist of numeric data, text data, image data, etc. [26]. *In this chapter, we focus on numerical data only*. Qin [16] discusses the 4Vs of big data in chemical process industries:

*(1) Volume*: There are massive process data available from process operation databases due to digital control systems.

*(2) Velocity*: Most of the process data are time-sensitive, and thus dynamic process monitoring and fault diagnosis using bid data analytics becomes increasingly important.

*(3) Variety:* Depending on the manufacturing process, there are different variety of process data available like numeric data, and non-numeric data such as texts, images, audio, video, etc.

*(4) Veracity*: precision, trust, correctness, reliability, etc. of the data are important. Establishing trust in big data and conclusions based on them is a continuing challenge in the process industries.

Our task for ML is to develop good models from datasets. A dataset generally consists of *feature vectors*, where each feature vector describes an object (e.g., polymer product quality) by using a set of *features* (e.g., melt index MI, density RHO, weight-average molecular weight MWW, number-average molecular weight MWN, polydispersity index PDI...). A feature vector is also called *an instance*, and a dataset is often called *a sample*.  We designate a dataset containing both input (independent variables, denoted by X) and output (dependent variables, denoted by Y) data as *a labeled dataset*, and a dataset containing only the input (X) data as *an unlabeled dataset.*

Two basic practices of preparing the dataset for ML are normalization and standardization. *Normalization* involves converting an actual range of values which a numerical feature (independent variable) can take into a standard range of value, typically between -1 to +1, or between 0 to 1. We demonstrate this application in our neural network example in Section 10.4.1. *Standardization* (also called *z-score normalization*) converting the given data to a zero mean, and scaled by their standard deviation. We show the details in Section A.1.7 of Appendix A of this book.

**10.1.2b  Representation**

The second component of an ML application is to develop the appropriate representation of the data inputs for ML model building, which depends much on the domain of application [26]. In our view, this component is the most challenging part of developing a ML model.

As we will discuss more in Section 10.4 below, deep neural networks (DNNs) find growing applications in predicting physical properties, chemical reactions, catalyst synthesis, and drug design, etc. [21, 22, 24, 96].

Developing effective molecular representations as inputs to DNNs in such applications continues to attract much attention. For example, based on molecular graph theory, SMILES (Simplified Molecular Input Line Entry System) [122] uses linear string of characters to represent atoms, bonds, branches, cyclic structures, disconnected structures, and aromaticity with coding rules. RInChI (Reaction International Chemical Identifier) [123] presents a machine-readable character string based on the InChI (IUPAC International Chemical Identifier) algorithm suitable for data storage and indexing. SMILES2Vec [124] is a DNN that automatically learns features (independent variables) from SMILES to predict chemical properties without the need for additional explicit feature engineering. The representations in these prior reports [122 to 124] consider little chemical information beyond atom type and connectivity, and ignore some important characteristics such as polarity and non-covalent interactions.

Overcoming some of these limitations, Maginn and his team [137] demonstrate in 2022 the use of sigma profiles as a universal molecular descriptor for data representation in DNNs. The resulting ML model accurately correlate and predicts a wide range of thermophysical properties of 1432 chemical components contained in our freely available, VT-2005 sigma profile database [138]. Note that the sigma profile is the probability distribution of a molecular surface segment having a specific charge density. Our database enables the public to apply the sigma profiles to predict thermophysical properties without doing tedious calculations of solvation thermodynamics and computational quantum mechanics. Maginn et al. [137] further demonstrate the ability of the resulting DNN model to incorporate thermodynamic conditions (such as temperature) as additional inputs to broaden the applicability of the model.

We refer the reader to articles describing the current data representations of ML model inputs in different applications, for examples: catalyst design and discovery [20,22]; fermentation and biochemical engineering [29,30,73]; quality predictions in polymerization processes [48 to 52, 74, 154];drug discovery [53,149], air pollution forecasting [75], solubility prediction [76], chemical product engineering [28], fault diagnosis [42, 81,82, 116], process monitoring [43], classification applications

[46,117,136], process control [106, 108 to 111], soft sensor development[44, 83, 115], quality structure-property relationships (QSPR) [92,126], composite manufacturing [94 ,96, 103], and image classification [102].

### 10.1.2c Learning

Learning (training) is the process for generating models from data. A learning algorithm is the computational method used to carry out the learning. A learner is a learned or trained model.

We use available data to train, validate and test a ML model, such as a regression (prediction) model. To do this, we typically divide the available dataset of independent variables (feature data) and dependent variable (labeled data) into training data, validation data and test data.

- *Training data* refer to those data that we use to build the model and find the model parameter values.
- *Validation data* represent the new data that we use to check the accuracy of the developed model in predicting the dependent variable values (labeled data) that the model has not seen during training. We use the validation data to choose an appropriate learning algorithm and to find the best values of hyperparameters. We refer to properties in the learning algorithm that must be set *before* training as *hyperparameters*, such as the learning rate in the backpropagation algorithm in training a neural network [see Eq. (10.46), Section 10.4.1b)]. Hyperparameter is different from a *parameter* that the algorithm is learning during training; for example, those parameters are the weights and biases in a neural network model (see Figure 10.22).
- *Test data* are those new data that we use to assess the validated model before delivering it to the client or putting in in production.

Heuristically, we use of 70% of the dataset for training, 15% for validation and 15% for testing (in a 70:15:15 ratio). Alternatively, some users choose an 80:10:10 ratio. In the era of big data, datasets could have millions of examples; in such cases, it could be reasonable to use a 95:2.5:2.5 ratio [7, p. 49].

Learning can be supervised, semi-supervised, unsupervised and reinforcement.

### (1) Supervised Learning

In the context of process data analytics, the majority of ML applications use supervised learning techniques, which basically means that both *independent variables or features* (expressed as a feature vector X)*,* and *dependent variables or labels* (expressed as a label vector Y) are available in the data. In this book, we limit each label within the label vector Y as a real number or one of a finite set of classes. For process applications, independent variables (X) are the process input variables and operating conditions like feed flows, temperature, pressure etc.; dependent variables (Y) are the process outputs and product quality measurements like concentrations, molecular weights, density, etc. Denote $x_i$ and $y_i$ as i-th scalar components of feature vector X and label vector Y. The task of supervised learning is [1, p. 653]: *Given a training set of N sample input-output pairs, $(x_1, y_1), (x_2, y_2),\ldots (x_N, y_N)$, where each pair was generated by an unknown function $y = f(x)$, discover a function $h(x)$ that approximates the true function f.*

The resulting function is a learned model, often called *a hypothesis*. How good is a learned model or hypothesis? It depends on how is well it does in predicting the outputs or labels for inputs or features that it has not seen before. We say that a learned model or hypothesis h(x) generalizes well if it accurately predicts the outputs of test data.

The *bias* refers to the tendency of a learned model or hypothesis to deviate from the expected value when averaged over different training data. The *variance* represents the amount of change in the learned model or hypothesis due to fluctuation in the training data.

Often there is *a bias-variance tradeoff*, that is, a choice between more complex, low-bias hypotheses that fit training data well with simpler, low-variance hypotheses that may generalize better. We illustrate this aspect more in Figure 10.2 and Section 10.2.1d(3).

In supervised learning applications like process monitoring and control, and soft sensors, we use *regression models* to fit an empirical model for any of process outputs/product quality as a function of process inputs. For some applications, we might also need to use *classification models*, for example, to classify some product batches into different categories given the product data labels.

Most supervised learning algorithms dealing with process data analytics can be classified into two major categories of *predictive and causal models*. Popular machine learning models are known for predictive modeling. Neural networks (NN) are one of the most popular algorithms that are known to give the highest prediction accuracy. They are made up of interconnected nodes (neurons or processing elements) that process information by its dynamic state to external inputs [4]. These are known to be kind of black box models which do not require much feature engineering and can predict the output with high accuracy. As we will discuss in Section 10.2, there are other conventional algorithms like ridge regression [39], support vector machines [44 to 54], decision tree [55-60], etc. which can be used for predictive modeling.

## (2) Unsupervised Learning

Unsupervised Learning methods are useful when data of process independent variables or features (X) are available as inputs, but process dependent variables or labels (Y) are not available as outputs. The unsupervised learning algorithm creates a model that takes the input vector X, and transforms it into another vector or into a value that can be used to solve a practical problem. For example, in *clustering*, the model returns the id of the cluster of each feature vector in the dataset. In *dimensionality reduction*, the output of the model is a feature vector that has fewer features than the input X. In *outlier detection*, the output is a real number that indicates how a scalar component of the feature vector is different from a typical sample in the dataset [7].

For example, principal component analysis, covered in Sections 9.1 and 9.2, is a well-known unsupervised learning method for dimensionality reduction and outlier detection.

## (3) Semi-Supervised Learning

In semi-supervised learning, the dataset contains both labeled examples (with X and Y) and unlabeled examples (with X only). Usually, the quantity of unlabeled examples is much higher than the number of labeled examples.  We use a few labeled examples to mine more information from a large collection of unlabeled examples. This happens to be the case of polymer manufacturing, where the output quality measurements (Y) are measured at lower frequency compared to the input process variables (X).

The goal of a semi-supervised learning algorithm is identical to that of the supervise learning algorithm. By using many unlabeled examples, we hope to help the learning algorithm to produce a better model than using labeled examples alone. In Section 10.2.3, we will give examples of some semi-supervised learning algorithms.

**(4) Reinforcement Learning**

This description follows [7, 8]. Reinforcement learning is a subfield of ML where the learning system, called *an agent* (e.g., an algorithm or a decision-maker), "lives" in a prescribed environment (e.g., process plant) and is capable of observing the "state" (e.g., temperature, pressure, etc.) of that environment as a vector of independent variables or features. The agent can execute sequential actions (e.g., a controller) in every state, and different actions bring different rewards (e.g., product yields) in return (or in the terminology of psychology, reinforcements) and could also move the machine to another state of the environment. Some actions bring penalties as negative rewards. The goal of a reinforcement learning algorithm is to learn a "policy" (e.g., controller action), which is similar to a learned model or hypothesis f(x) as in supervised learning. Specifically, a policy takes the feature vector of a state as input and outputs an optimal action to execute in that state. This action is optimal if it maximizes the expected average reward or the cumulative reward. We note that in reinforcement learning, the decision making is sequential and the goal is long-term.

According to Geron [8], many robots implement reinforcement learning algorithms to learn how to walk. DeepMind's AlphaGo program is an example of reinforcement learning. It made the headlines in May 2017 when it beat the world champion Ke Jie in the game Go. It learned its winning "policy" by analyzing millions of games, and then play many games against itself. During the game, the learning was turned off; AlphaGo was just playing the "policy" it had learned.

In this book, we focus on short-term or one-shot decision making, where input examples are independent of one another and the predictions made before the decision making. Thus, we will not focus on reinforcement learning.

Hoskins and Himmelblau [106] published the first article on process control via neural networks and reinforcement learning in 1992. This article inspired a number of subsequent studies of reinforcement learning applied to process control of polymerization reactors [109 to 111]. Reviews on reinforcement learning for process control are available [107,108].

### 10.1.3 Suggested Resources to Get Started with Machine Learning

### 10.1.3a Reference Books on AI and ML

We recommend Russel and Norvig [1] as a modern reference for AI and ML, despite that the book is huge with 1115 pages. The book by Turning and Haugeland [2] is an early reference on AI. Our textbooks [3] and [4], as well as the excellent reference by Stephanopoulos and Han [5] cover the development of intelligent systems, including expert systems and neural networks within phase I- expert system era (~1983 to ~1995) and phase II –neural network era (~1990 to ~2008). Haykins [6] is a comprehensive book on neural networks and learning machines, although the book does not cover the latest development of deep learning since 2009. We recommend Burkov's *The Hundred-Page Machine Learning Book* [7], which is short, explaining most of the modern ML concepts in easy-to-understand terms.

### 10.1.3b Basic Training of Python

We assume the reader have some Python programming experience. If not, get started with http://learnpython.org. Try the official tutorials on Python.Org (https://docs.python.org/3/tutorial/), and study Appendix B of this book, "Introduction to Python for Chemical Engineers".

There are many suggested lists of top free online Python training courses. See, for example: (1) Top 10 free Python training courses: https://www.bestcolleges.com/bootcamps/guides/learn-python-free/ (2) Ten 10 best online Python classes of 2022: https://www.intelligent.com/best-online-courses/python-classes/; (3) Top 10 websites to learn Python programming for free in 2022: https://medium.com/javarevisited/10-free-python-tutorials-and-courses-from-google-microsoft-and-coursera-for-beginners-96b9ad20b4e6; (4) 14 great free online courses for learning Python: https://www.onlinecoursereport.com/free/learning-python/

The reader should become familiar with Python's main scientific libraries, particularly NumPy (https://numpy.org/), pandas (https://pandas.pydata.org/), Matplotlib –visualization with Python (https://matplotlib.org/).

### 10.1.3c   Books with ML Principles and Coding Examples

We highly recommend the 2022 book by Geron [8]. It covers the concepts, tools and techniques to build ML models using Scikit-Learn, Keras, and TensorFlow. Another 2022 book by Marsland [9] introduces ML concepts well, covering a wide range of topics with coding examples. Grus [10] covers the basic ML principals with examples of Python implementations. Chollet [11] is a practical book explaining many mathematical concepts well with coding examples, from the author of the excellent Keras library. The book by Raschka and Mirjalili [12] is also a good introduction of ML and leverages Python open source libraries. Lastly, references [13,14] describe Scikit-learn and TensorFlow libraries.

### 10.2 An Overview of Relevant Machine Learning Concepts and Models

This section introduces the basic concepts and applications of common ML algorithms. Our goal is to cover the essential concepts to enable our reader to study the literature and to choose potential algorithms for specific applications. We present common ML methods following the categories of supervised and unsupervised learning, together with their applications to problems of regression (prediction), classification, clustering, dimensionality reduction and outlier detection, as we discussed previously in Section 10.1.2c (we do not cover semi-supervised and reinforcement learning).

While neural networks are an important supervised learning method for regression, prediction and classification applications, we delay our discussion of new developments of neural networks to Section 10.4 (enhanced learning by deep neural networks).

In this section, we limit our discussion of neural networks to radial-basis-function networks (RBFNs), the most popular neural network for classification and clustering applications. This follows because the key concepts of RBFNs, such as clustering, nearest neighbors, and decision boundaries, are also important to other supervised and unsupervised learning methods for classification and clustering applications, such as support vector machines, K-nearest neighbors, and K-means clustering.

### 10.2.1 Supervised Learning Methods for Regression Applications

### 10.2.1a Linear Regression

This section follows [1, pp. 679-680]. We represent a linear regression by the following equation:

$$y_p = a_0 + a_1x_1 + a_2x_2 + \ldots\ldots + a_nx_n = [a_0\ a_1\ a_2\ldots.a_n] * \begin{bmatrix} x_0 \\ x_1 \\ x_2 \\ \vdots \\ x_n \end{bmatrix} = \mathbf{a^T} \cdot \mathbf{x} = f_a(\mathbf{x}) \qquad (10.1)$$

In the equation, $y_p$ is the predicted value of the dependent variable, $a_0$ is the bias term, $a_j$ is the j-th model parameter, $x_0$ is a pseudo-independent variable and is always equal to 1, and $x_j$ is the j-th independent variable. $\mathbf{a}$ is the model's parameter vector containing the bias term $a_0$ plus the weight factors $a_1$ to $a_n$ (the model parameters are also called *weight factors*). $\mathbf{x}$ is the independent variable vector. $f_a(\mathbf{x})$ is the resulting regression model or hypothesis.

We extend Eq. (10.1) to an m-dimensional vector of predicted value of dependent variable $\mathbf{y_p}$; a mx(n+1)-dimensional data matrix $\mathbf{X}$, that is, the matrix of one (n+1)-dimensional example per row of the bias term ($x_0$) plus n independent variables ($x_1\ x_2\ldots\ldots x_n$); and a (n+1)-dimensional model parameter vector $\mathbf{a}$ containing the bias term $a_0$ plus the feature parameters or weight factors $a_1$ to $a_n$. We write:

$$\mathbf{y_p} = [y_{p,1}\ y_{p,2}\ldots\ldots y_{p,m}]^\top = \begin{bmatrix} x_{10} & \cdots & x_{1n} \\ \vdots & \ddots & \vdots \\ x_{m0} & \cdots & x_{mn} \end{bmatrix} * \begin{bmatrix} a_0 \\ a_1 \\ a_2 \\ \vdots \\ a_n \end{bmatrix} = \mathbf{X}* \mathbf{a} \qquad (10.2)$$

Let the vector of outputs for the training examples be denoted $\mathbf{y} = [y_1\ y_2\ldots\ldots y_m]^\top$. We try to minimize a MSE (minimum squared error) cost or loss function to find the model parameter vector $\mathbf{a}$, denoted by $\mathbf{a^*}$, that minimizes the MSE loss function, denoted by L($\mathbf{a}$):

$$\text{Min L}(\mathbf{a}) = \text{Min MSE}(\mathbf{a}) = \text{Min} \| \mathbf{y_p} - \mathbf{y} \|^2 = \text{Min}\| \mathbf{X}*\mathbf{a} - \mathbf{y} \|^2 \qquad (10.3)$$

Setting the gradient with respect to $\mathbf{a}$ to zero gives:

$$\nabla_a\ L(\mathbf{a}) = \nabla_a\ \text{MSE}(\mathbf{a}) = 2\ \mathbf{X}^\top (\mathbf{X}*\mathbf{a} - \mathbf{y}) = 0 \qquad (10.4)$$

Rearranging, we find the minimum-loss weight factor vector **a\***as:

$$\mathbf{a^*} = \mathbf{(X^TX)^{-1}X^Ty} \tag{10.5}$$

Equation (10.5) is known as *the normal equation*. We call $\mathbf{(X^TX)^{-1}X^T}$ the *pseudoinverse* of the data matrix **X**. We note that not every matrix has an inverse, but every matrix has a pseudoinverse, even non-square matrices. It's easy to compute the pseudoinverse using the singular value decomposition (SVD). See Section A.2.5.3 of Appendix A of this book.   Reference [8, p. 113] shows how to implement the solution in Python.   Additionally, Code B.1 at the end of Appendix B of this book gives the Python implementation of the linear regression model.

For a more complete discussion of multivariate linear regression, please refer to the free online book by Dunn [37, Chap. 4] or to any textbook on multivariate statistical analysis such as Johnson and Wichern [38, Chap. 7].

### 10.2.1b  Performance Evaluation Metrics for Regression Models

For regression applications, we evaluate the ML model performance based on the following metrics.

(a) *Mean squared error (MSE)*: It measures how close the predictions are to the actual target values (labeled values). Let $y_i$ = original observation, $y_p$ = model-predicted value, and n = number of observation. We write:

$$MSE = \frac{1}{n}\sum_{i=1}^{n}(y_i - y_p)^2 \tag{10.6}$$

(b) *Root mean squared error (RMSE)*: It creates a single value to summarize the model error. By squaring the difference, this metric ignores the difference between over-prediction and under-prediction.

$$RMSE = \sqrt{MSE} \tag{10.7}$$

(c) *% Normalized RMSE*: Let $y_m$ = mean value of observations. We write:

$$y_m = \frac{1}{n}\sum_{1}^{n}y_i \qquad nRMSE = \frac{RMSE}{y_m}x100 \tag{10.8}$$

(d) *Coefficient of determination ($R^2$)*: It is a goodness-of-fit measure for regression models. This statistic indicates the percentage of the variance in the dependent variable that the independent variables explain collectively. R-squared measures the strength of the relationship between the model and the dependent variable on a convenient scale of 0 to 1.

### 10.2.1c  Polynomial Regression, Underfitting, Overfitting and Regularization

We modify a figure from [8, p. 130] to illustrate the concepts of underfitting, overfitting and regularization. Figure 10.1 shows that the simple, linear model (straight line) *underfits* the training data, *resulting in large deviations (bias)* between the model line and the training data (generated by a quadratic algebraic equation in the form of y = 0.3$x^2$ -0.3x +0.3 + noise).

The quadratic-equation model fits the training data best with smaller deviations, and the 6-order polynomial-equation model (in the form of y = $a_6x^6$ +$a_5x^5$ +……$a_2x^2$ +$a_1x^1$ +$a_0$) is severely overfitting the training data (that is, trying to fit the training data perfectly). By *overfitting*, we mean that the model fits well on the training data, but does not predict accurately new validation and test data, which the model has not seen, that is, it does not *generalize* well. In other words, we say that a model is overfitting the training data when it pays too much attention to a particular dataset it is trained on, causing it to perform poorly on unseen data [1, p. 655]. To reduce the generalization error with new validation data, we should feed the model with more training data until the validation error reaches the training error [8, p. 133].

Constraining the model to make it simpler and reduce the risk of overfitting is called *regularization* [1, p. 671; 8, p. 27]. In particular, regularization attempts to limit the complexity of a model, as we want to find the right balance between fitting the data perfectly and keeping the model simple enough to ensure that it generalizes well with new validation and test data.

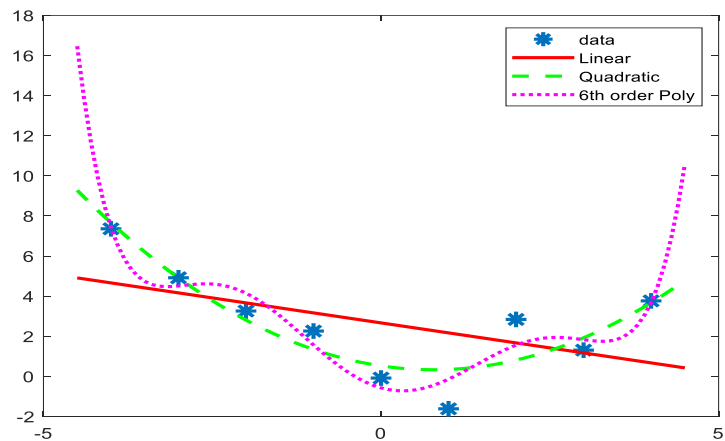

Figure 10.1   A comparison of good fitting with underfitting by a linear-equation (straight line) model, good fitting by a quadratic-equation model, and overfitting by a 6$^{th}$-order polynomial-equation model.

### 10.2.1.d   Regularized Linear Regression Models: Ridge Linear Regression and Lasso Linear Regression

Regularization is a good way to reduce overfitting by constraining the model, that is, by lowering the degrees of freedom a model has, which will make it harder to overfit the data. An example to regularize a polynomial regression model is to reduce the number of polynomial degrees [8, pp. 134-135].

For linear regression models, regularization typically involves constraining the model parameters or weight factors.  Two well-known regularized linear regression models are Ridge regression [8, p.135; 39] and Lasso (Least Absolute Shrinkage and Selection Operator) regression [8, p.137; 40,41].

### (1) Ridge Linear Regression

Also called Tikhonov regularization, Ridge linear regression adds a regularization term, $\alpha\sum_{i=0}^{n} a_i^2$ to the cost function, Eq. (10.3):

$$\text{Min } L(\mathbf{a}, \alpha) = \text{Min MSE}(\mathbf{a}, \alpha) = \text{Min } \| \mathbf{y_p} - \mathbf{y} \|^2 = \text{Min}\{\| \mathbf{X^*a} - \mathbf{y} \|^2 \ + \ \alpha(\frac{1}{2}) \sum_{i=1}^{n} a_i^2 \ \} \qquad (10.9)$$

where α is a hyperparameter. We note in Eq. (10.9), the last term does not include the bias term $a_0$ and the sum starts at $i$ =1, not 0. We may also represent the last term using the notation for Euclidean norm or $L_2$ norm, $\|\mathbf{a}\|_2$. This gives:

$$\text{Min } L2(\mathbf{a}, \alpha) = \text{Min MSE}(\mathbf{a}, \alpha) = \text{Min}\{\|\mathbf{X*a} - \mathbf{y}\|^2 + \alpha\left(\tfrac{1}{2}\right)[\|\mathbf{a}\|_2]^2 \qquad (10.10)$$

Reference [7, p. 53] calls this *the L2 regularization* for Ridge linear regression. There is a closed-form solution to the Ridge linear regression [8, p. 136]:

$$\mathbf{a*} = (\mathbf{X^T X} + \alpha\mathbf{I})^{-1}\mathbf{X^T y} \qquad (10.11)$$

where $\mathbf{I}$ is an (n+1)x(n+1) identify matrix with 0 in the top-left cell, corresponding to the bias term. The same reference shows how to implement the solution in Python.

**(2) Lasso Linear Regression**

Lasso linear regression replaces the Euclidean or L2 norm in the loss function, Eq. (10.10), by a simple vector or L1 norm:

$$\text{Min } L1(\mathbf{a}, \alpha) = \text{Min MSE}(\mathbf{a}, \alpha) = \text{Min}\{\|\mathbf{X*a} - \mathbf{y}\|^2 + \alpha\sum_{i=1}^{n}|a_i| \qquad (10.12)$$

Reference [7, p. 53] calls this *the L1 regularization* for Lasso linear regression. The hyperparameter for Lasso linear regression is typically between 0 to 1. This regularization method tends to eliminate those model parameters or weight factors $a_i$ of the least important features or independent variable $x_i$, that is, to set $a_i$ to zero [8, p. 137] . This results in a sparse regression model with few nonzero model parameters or weight factors. Reference [8, p. 139] shows how to implement the solution in Python.

**(3) Bias, Variance, and Noise (Irreducible Error): The Bias-Variance Tradeoff**

Applying a machine learning model to new validation data that the model has not seen previously results in a model's generalization error, which is commonly expressed as the sum of three components, namely, *bias, variance, and noise (irreducible error)* [8, p. 133]:

(1) *bias*: this error typically results from having a wrong model assumption, such as using a linear model while the actual data were generated from a quadratic-equation model. A model that underfits the data leads to a high bias value.

(2*) variance*: this error typically results from the model's excessive sensitivity to small variations in the training data. A model with a high degree of freedom, such as a sixth-order polynomial model with a bias term $a_0$ and six model parameters $a_1$ to $a_6$, is likely to lead to a high variance and to overfit he data.

(3) *noise* (irreducible error): this error results from the noisiness of the data itself.  To lower this error, we must clean up the data by fixing the data sources, such as removing the outliers and fixing the broken sensors.

Increasing a model's complexity (such as going from a linear regression model to a six-order polynomial regression model) typically increases the model's variance and reduces its bias. By contrast, reducing a model's complexity increases its bias and reduces its variance. This corresponds to the well-known bias-variance tradeoff, as illustrated in Figure 10.2.

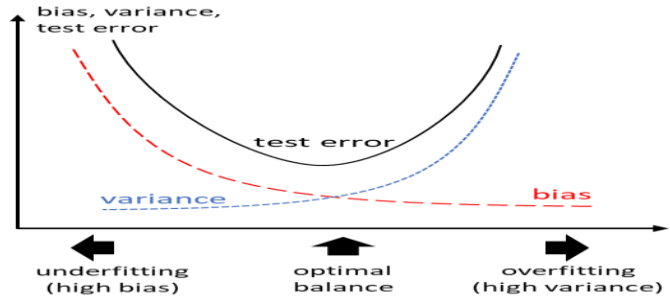

Figure 10.2   An illustration of the bias-variance tradeoff in ML models

**(4) Effect of Ridge Regularization on Linear and Polynomial Regression Models**

We adopt an example from [8, p. 136] to illustrate both linear and polynomial regressions using different levels of Ridge regularization, and explain the concept of bias-variance tradeoff in machine learning models. Figure 10.3 shows the effect of imposing different levels of Ridge regularization on the variance and bias of a linear regression model (left) and a polynomial regression model (right). We see that increasing the level of Ridge regularization through increasing the value of the hyperparameter $\alpha$ in Eq. (10.9) for the linear regression model on the left figure from $\alpha = 0$ to 1 to 10 and to 100 results in less extreme, and more reasonable predictions, thus reducing the variance of the model from the data, while increasing its bias [8, p. 136]. Likewise, increasing the level of Ridge regularization for the polynomial regression model on the right figure gives a similar result.

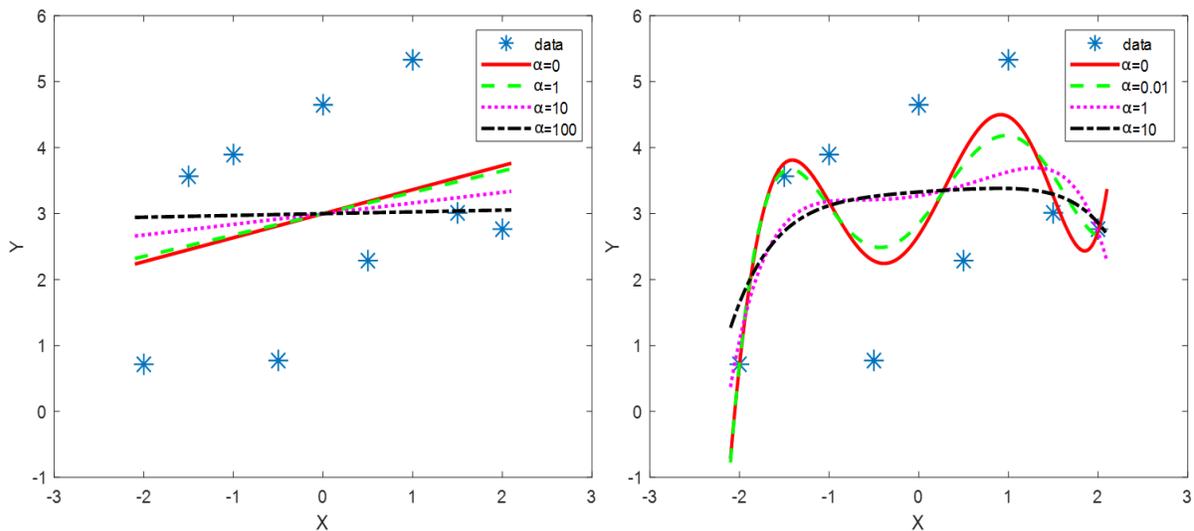

Figure 10.3 A linear regression model (left) and a polynomial regression model (right) with Ridge regularization with increasing values of hyperparameter $\alpha$:  increasing $\alpha$ value results in a smaller variance and a larger bias.

**10.2.2 Supervised Learning Methods for Classification Applications**

**10.2.2a Logistic Regression**

Why do we include a "regression" model within a section for "classification" model? Logistic regression is commonly used for classification, as it can output a value that corresponds to the probability of belonging to a class (e.g., 10% of being spam).

Using the same notations as in Eqs. (10.1) to (10.3), we write the estimated probability $\hat{p}$ as:

$$\hat{p} = f_a(\mathbf{x}) = \sigma(\mathbf{a^T \cdot x}) \qquad (10.13)$$

where σ represents the sigmoid activation function:

$$\sigma(x_i) = \frac{1}{1 + e^{-x_i}} \qquad (10.14)$$

See Table 10.7 and Figure 10.23 below for more about the sigmoid function, which has an output value bounded between 0 and 1. We pause to mention that Eqs. (10.13) and (10.14) are similar to the basic operation of a node or neuron in a neural network, as illustrated previously in Figure 8.101 in Section 8.3.2a. In Section 10.4.1a below, we give details of a neural network operation.

Based on Eq. (10.13), if the estimated probability ($\hat{p}$) is greater than 50%, then the model predicts that the instance belongs to the positive class, labeled "1". Otherwise, the model predicts that the instance belongs to the negative class, labeled "0" [8, p. 142]. Once the model applies Eqs. (10.13) and (10.14) to estimate the probability that an instance **x** belongs to the positive class, we can write the prediction of the output $y_p$ :

$$y_p = 0, \text{ if } \hat{p} < 0.5$$

$$= 1, \text{ if } \hat{p} \geq 0.5 \qquad (10.15)$$

Figure 10.4 illustrates the logistic regression's prediction moving from negative class region ($y_p = 0$) to positive class region ($y_p = 1$), resulting in *a decision boundary* and *a transition region*. Understanding the decision boundary is important to classification models discussed below.

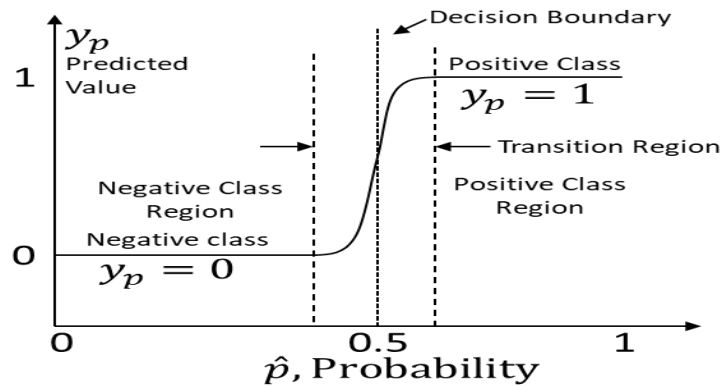

Figure 10.4   An illustration of classification regions, decision boundary and transition regions.

How do we train a logistic regression model to find the model parameters or weight factors, [$a_0$ $a_1$ $a_2$....$a_n$] or $\mathbf{a^T}$ so that the model estimates high probabilities of positive instances ($y_p = 1$) and low probabilities of negative instances ($y_p = 0$)? For a single training instance, we define a cost function C($\mathbf{a}$) as [8, p. 144]:

$$C(\mathbf{a}) = -\log(\hat{p}) \quad \text{if } y_p = 1$$

$$= -\log(1 - \hat{p}) \quad \text{if } y_p = 0 \qquad (10.16)$$

We take the average cost over all training instances as the cost function over the whole training set. This gives [8, p. 144]:

$$\text{Min } L(\mathbf{a}) = -\frac{1}{m}\sum_{i=1}^{m}[y_{p,i}\log(\hat{p}_i) + (1 - y_{p,i})\log(1 - \hat{p}_i)] \qquad (10.17)$$

Contrary to linear regression, there is no closed-form solution to the logistic regression problem defined by Eq. (10.17). We typically use a numerical procedure based on a gradient descent optimization. The gradient or the partial derivative of this cost function with respect to the model parameter or weight factor $a_j$ is:

$$\nabla_a L(a) = \frac{\partial}{\partial a_j}L(\mathbf{a}) = \frac{1}{m}\sum_{i=1}^{m}(\sigma(\boldsymbol{a}^T\boldsymbol{x_i}) - y_{p,i})x_{j,i} \qquad (10.18)$$

This equation computes the prediction error and multiplies it by the j-th feature or independent variable value $x_{j,i}$, and then it computes the average error over all training instances. Once we have the gradient vector containing all the partial derivatives, we can apply the gradient descent algorithm. Reference [8, p. 145] shows how to implement this solution in Python.

With the advances in neural computing since 1990, neural networks have essentially replaced logistic regression for classification applications. In Section 10.4.1a, we present a detailed example of multilayer perception network applied to classification problems.

### 10.2.2b Radial Basis Neural Network (RBFN)

RBFN is among one of the several online lists of top 10 ML algorithms that beginners should know in 2022 [164 to 166]. It includes some key concepts of clustering, nearest neighbors, and decision boundary that are common to several other algorithms for both regression and classification applications. We update our previous discussion of RBFN [4, pp. 115-120] below.

### (1) RBFN Architecture

Figure 10.5 illustrates the architecture of a RBFN [4], with N nodes (or neurons) in the input layer, L nodes in the hidden layer and M nodes in the output layer.

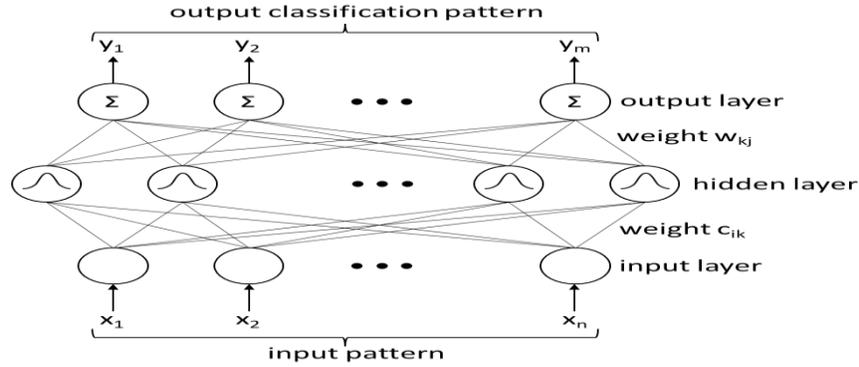

Figure 10.5 The architecture of a RBFN. Adapted from [4] with permission.

*The input layer* uses a direct transfer function, so that an input vector **x** with elements $x_i$ (i= 1 to N) gives the same vector **x** as its output vector.

The most important processing step in a RBFN is *the hidden layer*. There are L nodes (k = 1 to L) that are *radially symmetric*. Each hidden node has three key components:

(1) *A centroid* [8, p. 240] *or a cluster center vector* **$c_k$** in the input space, consisting of cluster centers with elements $c_{ik}$ (i = 1 to N), which are stored as weight factors between the input and hidden layers (see Figure 10.5). Figure 10.7 below illustrates three cluster centers **$c_k$** (k= 1 to 3) in a two-dimensional input space, which we will discuss below.

(2) *A distance measure* to determine how far an input vector **x** with elements $x_i$ (i= 1 to N) is from an assumed cluster center vector **$c_k$**. We quantify this distance $I_k$ (k= 1 to L) between the two vectors **x** and **$c_k$** by their Euclidean norm or *$L_2$* norm:

$$I_k = \| \mathbf{x} - \mathbf{c_k} \| = [\textstyle\sum_{1}^{N}(x_i - c_{ik})^2]^{1/2} \qquad (10.19)$$

(3) *A transfer function* which converts the Euclidean norm, Eq. (10.16), to give an output for each node. A popular choice is the Gaussian transfer function, illustrated in Figure 10.6, which transforms $I_k$ (k= 1 to L) to an output from the k-th node, $v_k$ (k = 1 o L), assuming a width of $\sigma_k$ (k = 1 to L):

$$v_k = \exp(-I_k^2/\sigma_k^2) \qquad (10.20)$$

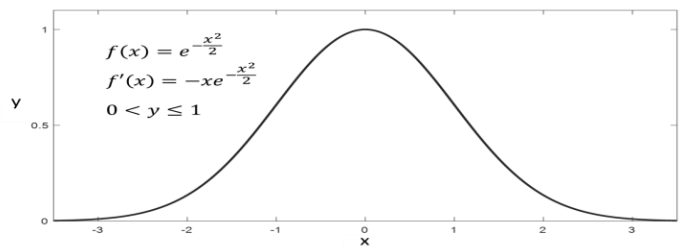

Figure 10.6 The Gaussian transfer function.

We see that the hidden layer processes the output from the input layer through a distance calculation, Eq. (10.19) and a transfer function, Eq. (10.20). To develop a RBFN network, we assume initial values of the cluster center vector $\mathbf{c_k}$ and the Gaussian transfer function width $\sigma_k$, and use the training data to update their values, as we discuss below.

We now turn to *the output layer* of a RBFN. In Figure 10.5, we see the weight factors $w_{kj}$ (k =1 to L; j = 1 to M) between the k-th node in the hidden layer and the j-th node of the output layer. We find the output $y_j$ from the output layer using a standard transfer function, as shown in Figure 8.101 previously and illustrated in details in Section 10.4.1a below.

## (2) K-Means Clustering Algorithm for Finding the Cluster Center Vector, $\mathbf{c_k}$

The K-means clustering algorithm begins with *an initialization step*. We assume a set of cluster center (or centroid) vectors $\mathbf{c_k}$ (k =1 to L) for the L nodes in the hidden layer, with elements $c_{ik}$ (i = 1 to N, k= 1 to L), stored as weight factors between the input and hidden layers. We also assume that there are T training examples available to the input layer with N nodes, and represent them as T training vectors $\mathbf{x}$(t) with elements $x_{it}$ (i = 1 to N; t = 1 to T).

The algorithm continues with *an iterative step* to find a desirable set of L center vectors $\mathbf{c_k}$ (k =1 to L) that minimizes the sum of squares of the distance between T training vector $\mathbf{x}$(t) and their nearest L centers $\mathbf{c_k}$ (k =1 to L).

Figure 10.7 illustrates three cluster centers (or centroids) $\mathbf{c_1}$ to $\mathbf{c_3}$, in a two-dimensional input space for a network having three nodes in the hidden layer (L = 3). In the figure, we see that input value $x_{1t}$ has a moderate output response, $v_3$, but it does not activate nodes 1 and 2 (no $v_1$ or $v_2$, respectively). Look at the cluster center $\mathbf{c_3}$. The activation of node 3 by the input value $x_{1t}$ generates an output response $v_3$, which reaches its maximum value at the center point and decreases as the distance between the input value $x_{1t}$ and the cluster center, $c_{3k}$, increases. We also note that input value $x_{2t}$ is not within any of the three cluster groups, and does not activate any output response associated with it.

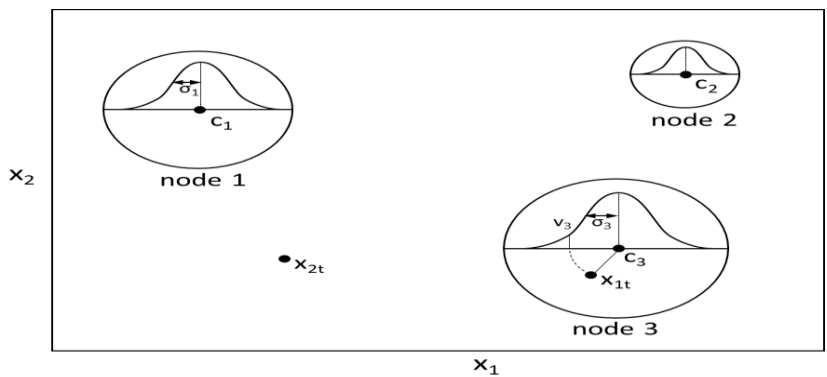

Figure 10.7   An illustration of three cluster centers (centroids) $\mathbf{c_1}$ to $\mathbf{c_3}$, in a two-dimensional input space for a RBFN having three nodes in the hidden layer (L = 3). Not the width of the Gaussian function $\sigma_k$ (k = 1 to 3) and two training examples $x_{1t}$ and $x_{2t}$. Adapted from [4] with permission.

With this background information, the K-means clustering algorithm involves the following iterative steps:

1) Read the next training vector **x**(t) with elements $x_{it}$ (I = 1 to N) into the input layer as t increases from 1 to T.

2) Modify only th cluster center **c**$_k$ (k= 1 to L) closest to the training vector **x**(t) with elements $x_{it}$ (I =1 to N) in the Euclidian distance:

$$c_k^{new} = c_k^{old} + α\,[\mathbf{x}(t) - c_k^{old}] \qquad\qquad (10.21)$$

In the equation,  α is a hyperparameter, the learning rate, which decreases as t increases from 1 to T.

3) Repeat the process for a fixed number of iterations.

### (3) P-Nearest Neighbor Algorithm for Finding the Width of the Gaussian Transfer Function, σ$_k$

We now present the P-nearest neighbor algorithm [1, p. 691; 4, p. 117; 8, p. 21] for finding the values of the width of the Gaussian transfer function, σ$_k$, in Eq. (10.20) and displayed in Figure 10.7. Consider a given cluster center vector **c**$_k$ (k= 1 to L) and assume $c_{k1}$, $c_{k2}$, ...., $c_{kp}$ (1≤k1, k2, …kp ≤ L) are the P nearest neighboring centers. We find the width of the Gaussian transfer function, σ$_k$, as the root-mean-squares (RMS) distance of the given cluster center vector **c**$_k$ to the P nearest neighboring centers:

$$σ_k = [\tfrac{1}{P}\textstyle\sum_{p=1}^{P} \|c_k - c_{kp}\|^2\,]^{1/2} \qquad\qquad (10.22)$$

In Appendix B of this book, code B.2 and Table B.1 at the end give the Python implementation of the P-nearest neighbor algorithm, together with a list of common parameters and their suggested values.

### (4) Weight Factors between the Hidden Layer and the Output Layer, w$_{kj}$

In Figure 10.5, we see that the hidden layer has L nodes and the output layer has M nodes, and the weight factors between the two layers are $w_{kj}$ (k = 1 to L; j = 1 to M). In Section 10.4.1a below, we illustrate the well-known *backpropagation algorithm* for training a multilayer neural network and describe its limitation and required changes to training a deep neural network. After completing the K-means clustering procedure, we train a RBFN using the backpropagation algorithm to find the weight factors $w_{kj}$ (k = 1 to L; j = 1 to M), and then compute the network output $y_j$ (j = 1 to M) using a sigmoid transfer function f( ) defined in Table 10.7, displayed in Figure 10.23, and illustrated in Section 10.1.4a:

$$y_j = f\left(\textstyle\sum_{k=1}^{L} w_{kj}\, v_k - T_j\right) = f\left(\textstyle\sum_{k=0}^{L} w_{kj}\, v_k\right) \qquad\qquad (10.23)$$

where $T_j$ is the internal threshold for node j in th output layer. See Section 10.4.1a, particularly Figures 10.20 and 10.22 about the internal threshold and its alternative representations by a bias node and the corresponding weight factor ( $v_0 = 1$, $w_{0j} = T_j$).

### (5) Insight and Experience for Training a RBFN and for Using the K-Means Clustering Algorithm

We summarize some insight and experience for training a RBFN [4, pp. 118-120, 129-130] and for applying the K-means clustering algorithm [8, pp. 247-249] in this section.

Training a RBFN using K-means clustering and P-nearest neighbor algorithms has two important features. First, we do not use any desired output to train the input connections to the hidden layer.

Second, for any training example represented by the t-th training vector $\mathbf{x}(t)$ with elements $x_{it}$ ($I$ = 1 to N), we iteratively modify only one cluster center $\mathbf{c}_k$ ($k$ = 1 to L) closest to the t-th training vector in the Euclidean distance. This means that we only change those stored weight factors $c_{ik}$ ($I$ = 1 to N; k =1 to L) between the input and hidden layers that are elements of one selected cluster center vector $\mathbf{c}_k$ ($k$ = 1 to L). Therefore, during the network training, we only evaluate only a small fraction of input nodes with cluster centers very close to the training input. This localized training speeds up the network training considerably.

Next, we typically determine the weight factors between the input and hidden layers, $c_{ik}$ ($I$ = 1 to N; k = 1 to L) using the K-means clustering during the first 2000 iterations. Over the same time period, we do not train the weight factors between the hidden and output layers, $w_{kj}$ ($k$ = 1 to L; j = 1 to M), by setting both the learning rate (see Eq. 10.41, Section 10.4.1a) and momentum coefficient (see Eq. 10.46, Section 10.4.1.b) to zero. We call this *the data-clustering phase.* This initial training gives the weight factors between the input and hidden layers, $c_{ik}$ ($I$ = 1 to N; k = 1 to L), which are fixed while we continue to train the weight factors between the hidden and output layers, $w_{kj}$ ($k$ = 1 to L; j = 1 to M).

It is important *to scale the input features* (i.e., prepropossing the dataset, such as mean-centered and scaled by standard deviation discussed in Sections A.1.5 to A.1.17 in Appendix A) before we run K-means, or the clusters may be very stretched and K-means will perform poorly. Scaling does not guarantee that all the clusters will be nice and spherical, but it generally improves things [8, p. 249].

Additionally, we need to specify the number of clusters when applying the K-means clustering algorithm, which can be challenging. Therefore, we need to run the algorithm several times with difference numbers of clusters in order to avoid suboptimal solutions. K-means algorithm does not perform well when the clusters have varying sizes, densities, and nonspherical shapes [8, p. 247]. Reference [8, pp. 239-252] shows how to implement the K-means clustering algorithm in Python. A Google search will also give many Python coding examples for implementing the RBFN.

### 10.2.2c Support Vector Machine (SVM) Classification and Regression

In the early 2000s, SVM was the most popular algorithm for supervised learning applications, especially for those who do not have specialized prior knowledge about a domain. Currently, that position has been taken over by enhanced learning by ensemble methods (Section 10.3) and by deep neural networks (Section 10.4) [1, p. 692]. SVM has found significant applications to soft sensing modeling [44] and process fault diagnosis [46], to prediction of polymer density and melt index in polyolefin manufacturing [48 to 52], and to novel drug discovery [53].   SVM is among many online lists of top 10 machine learning algorithms that beginners should know in 2022 [164 to 166]. Two tutorial articles [45,47] are available.

### (1) Linear SVM Classifier

We follow [1, pp. 692-696; 7, pp. 3-7, 30-34; 8, pp. 153 to 172] to discuss the fundamentals and practice of SVM. We begin with some graphical illustrations that demonstrate the key concepts of the SVM. Following the concept of decision boundary in Figure 10.4, we show in Figure 10.8 the decision boundaries (lines 1 to 3) of three potential linear classifiers. We note that lines 1 and 3 separate the dataset into two classes, but there is essentially a minimum margin of separation between the two separated classes, making these classifiers impractical. The model that is represented by the decision boundary of line 2 performs poorly that it does not even separate the two classes properly.

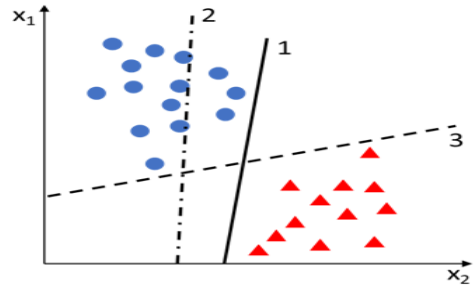

Figure 10.8 Decision boundaries (lines 1 to 3) of three potential linear classifiers

In Figure 10.9, we show a better linear classifier, where there is a distinct decision boundary separating the two classes of the dataset with sufficient margins of separation

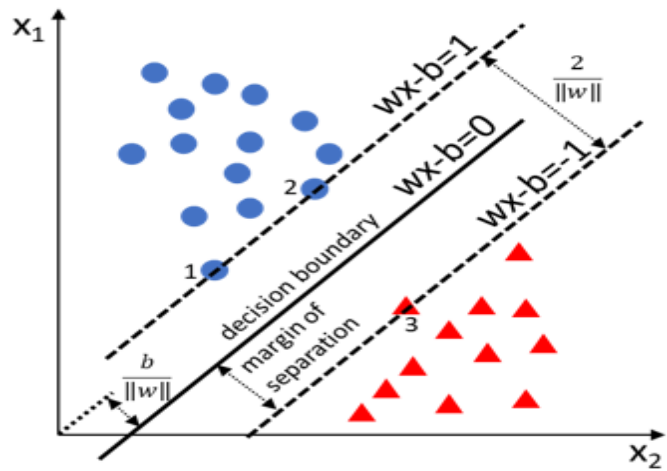

Figure 10.9   A SVM model for a two-dimensional feature vectors illustrating the distillation boundary ($\mathbf{w}^T*\mathbf{x} - b = 0$), constraints, margin of separation, and support vectors 1 to 3

Given an input feature (independent variable) vector $\mathbf{x} = [\, x_1 \; x_2 \; \ldots. \; X_D\,]^T$, where D is the number of dimensions of the feature vector, the SVM algorithm puts all feature vectors on a D-dimensional "line" (actually a *hyperplane*) that separate examples with positive labels from examples with negative labels. We define *the decision boundary* with a real-valued weight factor vector $\mathbf{w} = [w_1 \; w_2 \; \ldots. \; w_D]^T$ and a real-valued "intercept " b:

$$\mathbf{w}^T*\mathbf{x} - b = 0 \qquad\qquad (10.24)$$

Assume that for each feature or independent variable $x_i$, there is a corresponding label or dependent variable $y_i$, which takes the value of -1 or +1, depending on the following constraints [7, p. 5]:

$$\mathbf{w}^T*\mathbf{x} - b \geq +1 \qquad \text{if } y_i = +1$$

$$\mathbf{w}^T * \mathbf{x} - b < -1 \qquad \text{if } y_i = -1 \qquad\qquad (10.25)$$

Figure 10.9 displays the two constraint "lines" (hyperplanes). We also see *the margin of separation* which is the distance between the closest examples of two classes as defined by the decision boundary. A large margin leads to a better generalization, that is, how well the model will classify new examples in the future. Geometrically, the constraint equations $\mathbf{w}^T * \mathbf{x} - b = 1$ and $\mathbf{w}^T * \mathbf{x} - b = -1$ represent two parallel hyperplanes.

We can think of the linear SVM classifier in Figure 10.9 as fitting the *widest* possible "street" bounded by two constraint "lines" (hyperplanes) between the classes, which is also called *large margin classification* [8, p. 153]. The width of the "street" is fully determined or "supported" by the instances or examples (labeled 1, 2 and 3 in Figure 10.9) located on the edge of the street, that is, on the two constraint "lines" (hyperplanes). We call instances 1 to 3 *the support vectors,* because they "hold up" the separating plane [1, p. 694].

As seen in Figure 10.9, the distance between these hyperplanes is $2/\|\mathbf{w}\|$, which suggests that the small the norm $\|\mathbf{w}\|$, the larger the distance between these hyperplanes. Therefore, to achieve a large margin classification, we need to minimize the Euclidean norm of $\mathbf{w}$, defined by $[\sum_{i=1}^{D} w_i^2]^{1/2}$. Minimizing $\mathbf{w}$ is equivalent to minimizing $\frac{1}{2}\mathbf{w}^T\mathbf{w}$ or $\frac{1}{2}\|\mathbf{w}\|^2$. Note that $\frac{1}{2}\|\mathbf{w}\|^2$ has a nice, simple derivative, which is just $\mathbf{w}$. By contrast, $\|\mathbf{w}\|$ is not differentiable when $\mathbf{w} = 0$. Using the squared norm form also makes it possible to use quadratic programming optimization that we will discuss below [7, p. 31; 8, p. 166].

We can now summarize the problem of developing a linear SVM classifier as follows [7, p. 31]:

$$\text{Min } \{\tfrac{1}{2}\|\mathbf{w}\|^2\} \text{ such that } y_i(\mathbf{w}^T * \mathbf{x} - b) -1 \geq 0, \quad i = 1, 2, \ldots.D \qquad (10.26)$$

In Figure 10.9, we strictly require that all instances must be off the "street" and on the right side of the constraint "lines" (hyperplanes) for the two classes. We call this *hard margin classification*. This type of classification will not work: (1) when there is no hyperplane that can perfectly separate positive examples from negative ones (that is, the data are not linearly separable); or (2) when the data contain noise (with outlier or examples with wrong labels). We discuss how to extend SVM to handle both cases in the following section that leads to the concept of *soft margin classification*.

## (2) Soft Margin Classification by SVM

Referring to Eqs. (10.10) and (10.12), we see that both Ridge and Lasso linear regressions incorporate a penalty term into the minimization objective function (loss function) to constrain the ML model by lowering the degrees of freedom that a model has.  We adopt the same approach by imbedding the constraint "lines" (hyperplanes), Eq. (10.25), into our minimization loss function. To do this, we need to become familiar with *the hinge loss function*, illustrated in Figure 10.10 [8, p. 173].

The hinge loss function is not differentiable at z =1; but just like for Lasso regression, we can still use the gradient descent algorithm for optimization using a "subderivative" of any value between 0 and 1 at t = 1 [8, p. 173].

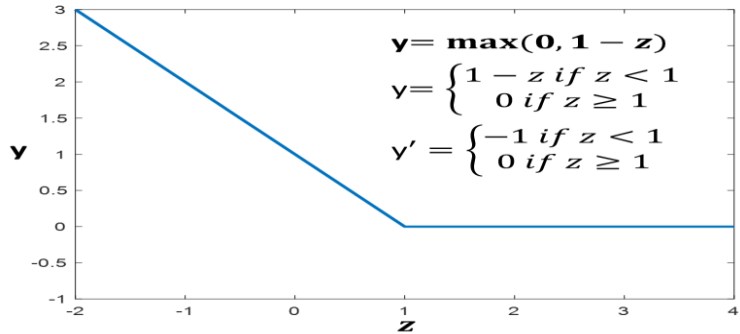

Figure 10.10 An illustration of hinge loss function

We follow [7, p. 31] in the discussion below. First, by imbedding our constraints, Eq. (10.24), into the hinge loss function: max [0, 1 - y$_i$($\mathbf{w}$ᵀ*$\mathbf{x}$ − b)], we see that the hinge loss function is zero if the constraints of Eq. (10.25) are satisfied. This means that $\mathbf{w}$ᵀ*$\mathbf{x}$ lies on the correct side of the decision boundary. For data on the wrong side of the decision boundary, the function's value is proportional to the distance from the decision boundary. Next, we modify the minimization problem of Eq. (10.26) by including a penalty term based on the hinge loss function [7, p. 31]:

$$\text{Min } \{C\|\mathbf{w}\|^2 + \frac{1}{D}\sum_{i=1}^{D} \max [0, 1 - y_i(\mathbf{w}^T\mathbf{x} - b)]\} \quad \text{i= 1, 2, ....D} \qquad (10.27)$$

In the equation, the hyperparameter C characterizes the tradeoff between increasing the width of the decision boundary and ensuring that each input feature x$_i$ lies on the correct side of the decision boundary. We typically find the appropriate value of C through multiple trials. Eq. (10.27) minimizes the hinge loss and is called *the soft margin SVM classification*, while Eq. (10.26) defines *the hard margin SVM classification.*

For sufficiently high values of C, the second term in Eq. (10.27) becomes negligible. Consequently, the SVM algorithm will try to find the highest margin by completely ignoring misclassification. When the value of C becomes small and makes classification errors costlier, the SVM algorithm will make fewer mistakes in classification by sacrificing the size (or distance) of the margin that separates the closest examples of positive and negative classes. A large margin contributes to a better generalization, that is, how well the model will classify new examples in the future. Therefore, hyperparameter C regulates the tradeoff between classifying the training data well (minimizing empirical risk) and classifying future examples well (improving generalization).

There is an alternative way of defining the SVM soft margin classification problem by introducing *the slack variable S$_i$* for each instance. Specifically, S$_i$ measures how much the i-th instance is allowed to violate the margin. We find $\mathbf{w}$, b and *S$_i$* (I = 1, 2, ...D) to minimize [8, p. 166]:

$$\text{Min } \{\frac{1}{2}\|\mathbf{w}\|^2 + C\sum_{i=1}^{D} s_i\} \quad \text{such that } y_i(\mathbf{w}^T\mathbf{x} - b) \geq 1 - S_i \quad \text{and } S_i \geq 0 \quad \text{i = 1, 2, ....D} \qquad (10.28)$$

Here, we wish to make the slack variables as small as possible to reduce the margin violations, and make $\frac{1}{2}\|\mathbf{w}\|^2$ as small as possible to increase the margin using the hyperparameter C to quantify this tradeoff.

**(3) Dealing with Inherent Nonlinearity and the Kernel Trick**

We follow [1, pp. 693-696; 7, pp. 31-34; 8, pp. 153-172] to introduce the "kernel trick" for dealing with inherently nonlinear datasets that cannot be separately by a hyperplane in the original space. Basically, if we could transform a two-dimensional non-separable dataset into a three-dimensional space, we could hope that the dataset will become linearly separable in the higher-dimensional space. See Figure 10.11.

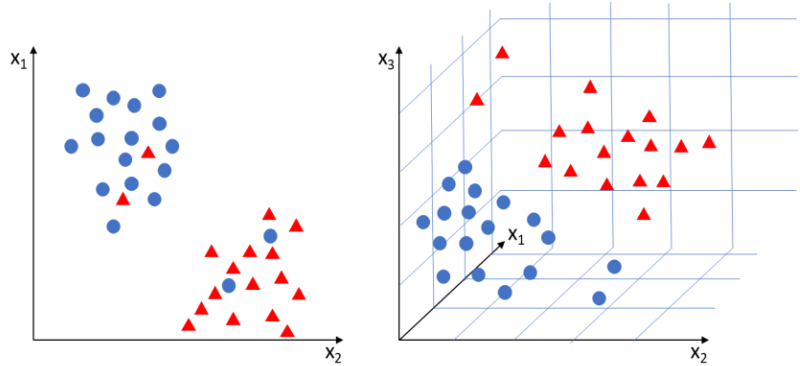

Figure 10.11 An illustration of transforming a linear two-dimensional non-separable case into linear separable case in a three-dimensional space

In SVMs, using a function to implicitly transform the original space into a higher dimensional space during the loss function optimization is called *the kernel trick*. For the example of two-dimensional training data displayed on the left of Figure 10.11, we could convert a 2-D feature vector $\mathbf{x}$ by a second-degree polynomial transformation $\phi(\mathbf{x})$ to a 3-D space [1, p. 694; 7, p. 31; 8, p. 169]:

$$\phi(\mathbf{x}) = \phi\left(\begin{pmatrix} x_1 \\ x_2 \end{pmatrix}\right) = \begin{pmatrix} x_1^2 \\ \sqrt{2}x_1 x_2 \\ x_2^2 \end{pmatrix} \tag{10.29}$$

Next, let us apply this transformation to two 2-D vectors of our training instances, $\boldsymbol{a}$ and $\boldsymbol{b}$:

$$K(\mathbf{a},\mathbf{b}) = \phi(\boldsymbol{a})^\top \phi(\boldsymbol{b}) = \begin{pmatrix} a_1^2 & \sqrt{2}\,a_1 a_2 & a_2^2 \end{pmatrix} \begin{pmatrix} b_1^2 \\ \sqrt{2}a_1 a_2 \\ a_2^2 \end{pmatrix} = a_1^2 b_1^2 + 2a_1 b_1 a_2 b_2 + a_2^2 b_2^2 = (a_1 b_1 + a_2 b_2)^2$$

$$= [(a_1\ a_2)^\top (b_1\ b_2)] = (\boldsymbol{a}^\top \mathbf{b})^2 \tag{10.30}$$

This result (or "trick") is very significant. It says that we do not need to transform the training instances $\boldsymbol{a}$ and $\boldsymbol{b}$ at all; just replace the dot products of transformed vectors, $\phi(\boldsymbol{a})^\top \phi(\boldsymbol{b})$, simply by $(\boldsymbol{a}^\top \mathbf{b})^2$. The result is strictly the same even if we had gone through the trouble of transforming the training instances, and then fitting a linear SVM algorithm. This trick makes the whole process much more computationally efficient [8, p. 169].

The function $K(\mathbf{a},\mathbf{b}) = (\boldsymbol{a}^\top \mathbf{b})^2$ represents a second-degree polynomial kernel. In ML, we refer to a function $\phi$ as *a kernel* if it is capable of computing the dot product $\phi(\boldsymbol{a})^\top \phi(\boldsymbol{b})$ based only on the original vectors of our training instances, **a** and **b**, without having to compute (or even know about) the transformation ϕ. Table 10.1 lists several commonly used kernels for SVMs [8, p. 171]:

Table 10.1 Commonly used kernels for SVMs

| Kernel name | Kernel form: K(**a, b**) |
|---|---|
| 1. Linear | $\boldsymbol{a}^\mathsf{T}\mathbf{b}$ |
| 2. Polynomial | $(\Upsilon\,\boldsymbol{a}^\mathsf{T}\mathbf{b} + r)^d$ |
| 3. Gaussian Radial Basis Function (RBF) | $\exp(-\,\Upsilon\|\mathbf{a} - \mathbf{b}\|^2)$ |

According to *Mercer's theorem* [8, p. 171; 44, 50,54], if a function K(**a, b**) respects a few mathematical conditions, called *Mercer's conditions* (e.g., K must be continuous symmetric in its arguments so that K(**a, b**) = K(**b, a**), etc.), then there exists a function ϕ that maps vectors of training instances **a** and **b** into another space (possibly with much higher dimensions) such that K(**a,b**) = ϕ(**a**)$^\mathsf{T}$ ϕ(**b**).  We can use K as a kernel because we know ϕ exists, even if we do not know what ϕ is. For the Gaussian RBF kernel, research has shown that ϕ maps each training instance to an infinite-dimensional space, so that it is truly significant that the kernel trick says that we do not need to actually perform the mapping!

## (4) Quadratic Programming, and Solution to Primal and Dual Problems of Soft Margin SVM Classification

The soft margin SVM classification problem, defined previously in Eq. (10.27), is to find **w**, b and slack variable $s_i$ (I = 1, 2, …D) to minimize

$$\text{Min } \{\tfrac{1}{2}\|\mathbf{w}\|^2 + C \textstyle\sum_{i=1}^{D} s_i \} \quad \text{such that } y_i(\mathbf{w}^{\mathsf{T}*}\mathbf{x} - b) \geq 1 - s_i \quad \text{and } s_i \geq 0 \quad i = 1, 2, \ldots.D \qquad (10.31)$$

Denoting a slack variable vector **s** = [ $s_1, s_2, \ldots.s_D$]$^\mathsf{T}$ , we can modify this minimization problem by incorporating the two linear constraints into the minimization loss function by introducing the Lagrange multipliers, **α** = [α$_1$, α$_2$, ….α$_D$]$^\mathsf{T}$ and **β** = [β$_1$, β$_2$, ….β$_D$]$^\mathsf{T}$ such that:

$$\text{Min } L(\mathbf{w}, b, \boldsymbol{s}, \boldsymbol{\alpha}, \boldsymbol{\beta}) = Min \{\tfrac{1}{2}\|\mathbf{w}\|^2 + C \textstyle\sum_{i=1}^{D} s_i + \sum_{i=1}^{D} \alpha_i \ [y_i(\mathbf{w}^{\mathsf{T}*}\mathbf{x} - b) -1 + s_i ] - \sum_{i=1}^{D} \beta_i \ s_i\} \qquad (10.32)$$

This is a constrained optimization problem with two linear constraints, known as *quadratic programming*, for which there are many "off-the-shelf" solvers available. Given this constrained optimization problem, for which the optimization literature calls it *a primal problem*, it is possible to develop a slightly different and closely related problem, called its *dual problem*. The solution to the dual problem typically gives a lower bound to the solution of the primal problem. Under some conditions, the dual problem can have the same solution as the primal problem.  Luckily, the current SVM minimization problem happens to meet these conditions; we can choose to solve the primal or the dual problem to find the same solution [8, p. 168].

For the primal problem defined by Eq. (10.32), interested readers may refer to [8, pp. 761-764; 54] for the details in developing a dual problem. We only write the resulting dual problem as follows:

$$\text{Min } \{\tfrac{1}{2}\textstyle\sum_{i=1}^{D}\sum_{j=1}^{D} y_i \ y_j \alpha_i \ \alpha_j \mathbf{x}_i^{\boldsymbol{T}}\mathbf{x_j} \ - \sum_{i=1}^{D} \alpha_i\}$$

subject to $\qquad\qquad \sum_{i=1}^{D} y_i \alpha_i = 0$ and $0 \leq \alpha_i \leq C$ $\qquad\qquad\qquad$ (10.33)

**(5) SVM for Regression and for Outlier Detection**

Referring to Figure 10.9, how do we apply the existing concepts of SVM for classification to regression? The answer is simple. Instead of finding the widest margin of separation, or the largest possible "street" between the two classes without violating the margins, SVM regression (i.e., SVR) tries to fit as many instances as possible *on the street,* that is, between the constraint lines (hyperplanes), while limiting margin violations (i.e., instances *off the street*). The basic idea behind SVR is to find the best fit line. In SVR, the best fit "line" is the hyperplane that has the maximum number of feature points.

Unlike other regression models that try to minimize the error between the real and predicted values, the SVR tries to fit the best "line" (hyperplane) within a threshold value. The threshold value is the distance between the hyperplane and boundary "line" (hyperplane).

Our preceding discussion of SVM has already provided sufficient background to enable the readers to understand many online resources on SVR principles and its implementation.

In Appendix B of this book, code B.7 and Table B.1 at the end give the Python implementation of the SVM algorithm, together with a list of common parameters and their suggested values.

Lastly, SVM can also be applied to outlier detection; see the online Scikit-Learn's documentation for more details.

**10.2.2d Decision Trees for Classification and Regression Problems**

**(1) Introduction to Decision Trees**

A decision tree has a structure similar to a conventional flow chart that uses a branching to illustrate every possible outcome of a decision. Each node with a tree represents a test on a specific variable, and each branch uses the outcome of that test.

Decision trees are versatile ML algorithms for both classification and regression applications. They are also the fundamental components of ensemble methods for enhanced learning, such as random forest that we will discuss in Section 10.4 [8, p. 175-186; 57]. One of the many qualities of decision trees is that they require very little data preparation. In fact, they require *no feature centering or scaling* at all [8, p. 176].

Figure 10.12 shows a decision tree corresponding to the dataset of Table 10.2. This example first appeared in Quinlan [56], and was repeated in many papers and online tutorials [57-60]. However, none of these references gives sufficient details for developing the decision tree, as we do below.

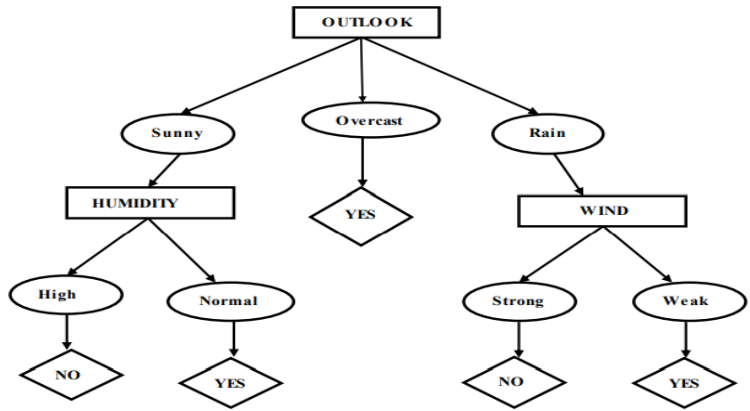

Figure 10.12 A decision tree example

In the figure, we represent each attribute name in a rectangle, each attribute value in an oval, and each decision (class) in a diamond with *No or N for negative*, and *Yes or P for positive*. "Outlook" represents the entire dataset, and it is *a root node (or a parent node)* from which the decision tree starts; it is split into three *branches or sub-trees*, namely, sunny, overcast and rain labeled in ovals. Upon calculations of some splitting criteria that we will introduce shortly, the "sunny" outlook is split further into two branches or sub-trees based on humidity being high or normal; likewise, the "rain" outlook is split further into two branches or sub-trees based on wind being strong or weak. Finally, we see that normal humidity leads to *a leaf node or final output node*, labeled positive or yes within a diamond; and high humidity leads to a leaf node labeled negative or no within a diamond. The same interpretations apply to weak wind being positive or yes and strong wind being negative or no within diamonds on the branches or sub-trees split from the wind rectangle. Finally, we note that we may call the subsequent nodes in different branches split from the root node or parent node *the child nodes.*

There are three basic algorithms to help with the development of decision trees: (1) CART (Classification and Regression Trees) algorithm by Breiman [55]; (2) ID3 algorithm (Iterative Dichotomiser 3) by Quinlan [56]; and (3) C4.5 algorithm by Quinlan [57]. The CART algorithm produces only binary trees with non-leaf nodes always having two children (that is, questions have only yes/no answers). Other algorithms, such as ID3, can produce decision trees with nodes that have more than two children. In the following, we first define some terminologies that we use as criteria for splitting trees, and then demonstrate their applications to the problem specified by Figure 10.12 and Table 10.2 following the ID3 and C4.5 algorithms.

Table 10.2 Training dataset for the decision tree example of Figure 10.12

| Day | Attribute | | | | Class |
|-----|---------|-------------|----------|------|------------------------|
| | Outlook | Temperature | Humidity | Wind | N(Negative); P( Positive) |
| 1 | sunny | hot | high | weak | N |
| 2 | sunny | hot | high | strong | N |
| 3 | overcast | hot | high | weak | P |
| 4 | rain | mild | high | weak | P |

| 5 | rain | cool | normal | weak | P |
| 6 | rain | cool | normal | strong | N |
| 7 | overcast | cool | normal | strong | P |
| 8 | sunny | mild | high | weak | N |
| 9 | sunny | cool | normal | weak | P |
| 10 | rain | mild | normal | weak | P |
| 11 | sunny | mild | normal | strong | P |
| 12 | overcast | mild | high | strong | P |
| 13 | overcast | hot | normal | weak | P |
| 14 | rain | mild | high | strong | N |

## (2) Terminologies for Tree Splitting Criteria

All decision tree algorithms require splitting criteria for splitting a node to form a tree. In many cases, an internal node is split based on the value of a single attribute, and the algorithm searches for the best attribute upon which to split. The main goal of splitting is to reduce the impurity of a node. The node impurity is a measure of the homogeneity of the labels at the node. The current implementation provides two impurity measures for classification (entropy and Gini index) and one impurity measure for regression (variance).

1. *Entropy*:

This term derives from Shannon's entropy (https://en.wikipedia.org/wiki/Entropy_(information_theory)). Any correct decision tree for dataset S will classify objects in the same proportion as their presentation in S. An arbitrary object will be determined to belong to class P (positive or yes) with probability p/(p + n), and to class N (negative or no) with probability n/(p + n). A decision tree is a source of a massage "P" or "N", with the expected information needed to generate this message as [56]:

$$\text{Entropy } I(p,n) = - \left(\frac{p}{p+n}\right) log_2 \left(\frac{p}{p+n}\right) - \left(\frac{n}{p+n}\right) log_2 \left(\frac{n}{p+n}\right) \qquad (10.34)$$

To generalize this expression to the case with multiple instances in a given node t, we write [59,60]:

$$\text{Entropy } I(p,n) = - \sum_{i=1}^{n} p\left(\frac{i}{t}\right) \; log_2 \, p\left(\frac{i}{t}\right) \qquad (10.35)$$

where $p\left(\frac{i}{t}\right)$ denotes the fraction of class *i* instances to the total training instances in a given node *t*.

2. *Gini Index* [8, p. 177; 59]:

The Gini index is a function that determines how well a decision tree was split. Basically, it helps us determine which splitter is best so that we can build a pure decision tree. Gini index has a maximum value of 0.5, which is the worst we can get, and a minimum value of 0 means the best we can get.

$$G_i = 1 - \sum_{i=1}^{n} \left[ p\left(\frac{i}{t}\right) \right]^2 \qquad (10.36)$$

3. *Information Gain* [56, 59,60]:

Information gain, G(S, A)  = Entropy (parent nodes) − Entropy (child nodes)

$$= \text{Entropy}(S) - \sum [\, p(S|A) * Entropy\ (S|A)] \qquad (10.37)$$

For a subset A splitting into child nodes {$C_1$, $C_2$, …. $C_v$}, let us assume that $C_i$ contains $p_i$ objects of class P (positive or yes) and $n_i$ of class N (negative or no), the expected information required for the subtree for $C_i$ is I($p_i$, $n_i$). We find the expected information required for the tree with A as root as the weighted average [56]:

$$\text{Entropy (child nodes) = E(A)} = \sum_{i=1}^{v} \frac{p_i + n_i}{p+n}\ I(p_i, n_i) \qquad (10.38)$$

Substituting Eqs. (10.31) and (10.34) into Eq. (10.37) gives

$$\text{Information gain (A) = I(p, n)} - E(A) \qquad (10.39)$$

The ID3 algorithm examines all candidate attributes and chooses A to maximize information gain (A), forms the tree as above, and then uses the same process recursively to form trees for the residual subsets $C_1$, $C_2$, …. $C_v$ [56].

4. *Gain Ratio* [59]:

The C4.5 algorithm uses the gain ratio as the splitting criterion, which is defined as:

$$\text{Gain ratio = [Information gain (A)]/[Entropy I(p, n)]} \qquad (10.40)$$

**(3) Application of ID3 and C4.5 Algorithms to the Training Dataset S of Table 10.2**

We apply the ID3 and C4.5 algorithms as follows:

1. Calculate the entropy of th dataset.

2. For each attribute/feature, calculate the entropy for all its categorical values by Eq. (10.35). Calculate the information gain for the feature by Eq. (10.39). Calculate the gain ratio for the feature by Eq. (10.40).

3. Find the feature with the maximum information gain for ID3 algorithm, and with the maximum gain ratio for the C4.5 algorithm.

4. Repeat it until we get the desired tree.

(Step 1) For the 14-day data of Table 10.2, we have 9 P (positive or yes) class and 5 N (negative or no) class.

With p = 9, n = 5 and p+n =14, Eq. (10.35) gives the complete entropy of the dataset as:

$$\text{Entropy (S) = Entropy I(p, n)} = - \left(\frac{p}{p+n}\right) log_2 \left(\frac{p}{p+n}\right) - \left(\frac{n}{p+n}\right) log_2 \left(\frac{n}{p+n}\right)$$
$$= -(9/14)*log_2(9/14) - (5/14)*log_2\ (5/14) = \underline{\textbf{\textit{0.94}}}$$

In Steps 2 to 4, we evaluate the four attributes for the dataset, including Outlook, Temperature, Humidity and Wind.

(Step 2) First attribute- Outlook with categorical values of sunny ($p_1$ =2, $n_1$ =3), overcast ($p_2$ =4, $n_2$ =0) and rain ($p_3$ =3, $n_3$=2)

*Entropy(Outlook=sunny) == -(2/5)\*log₂(2/5) – (3/5)\*log₂ (3/5) = 0.971*

*Entropy(Outlook=overcast) == -(4/4)\*log₂(4/4) – (0/4)\*log₂ (0/4) = 0*

*Entropy(Outlook=rain) == -(3/5)\*log₂(3/5) – (2/5)\*log₂ (2/5) = 0.971*

E(outlook), weighted-average entropy for Outlook according to Eq. (10.34)

= p(sunny)\*entropy (outlook=sunny) + p(overcast)\*entropy(outlook=overcast)
  +p(rain)\*entropy(outlook=rain)

= (5/14)\*0.971 + (4/14)\*0 + (5/14)\*0.971 =**0.693**

Information gain according to Eq. (10.39) = Entropy(S) – E(Outlook) =0.94 – 0.693 =**0.247**

Gain ratio according to Eq. (10.40) = 0.247/0.94 = **0.262**

(Step 3) Second attribute – Temperature with categorical values of hot ($p_1$ =2, $n_1$ =2), mild ($p_2$ =3, $n_2$ =1) and cool ($p_3$ =4, $n_3$ =2)

*Entropy(Temperature=hot) = -(2/4)\*log₂(2/4) – (2/4)\*log₂(2/4) = 1*

*Entropy(Temperature = mild) = -(3/4)\*log₂(3/4) – (1/4)\*log₂ (1/4) = 0.811*

*Entropy(Temperature = cool) = -(4/6)\*log₂(4/6) – (2/6)\*log₂ (2/6) = 0.9179*

E(Temperature), weighted-average entropy for Temperature according to Eq. (10.34)

= p(hot)\*entropy (temperature = hot) + p(mild)\*entropy(temperature = mild)
  +p(cool)\*entropy(temperature =cool)

= (4/14)\*1 + (4/14)\*0.811 + (6/14)\*0.9179 =**0.9108**

Information gain according to Eq. (10.39) = Entropy(S) – E(Temperature) =0.94 – 0.9108
  =**0.0292**

Gain ratio according to Eq. (10.40) = 0.0292/0.94 = **0.032**

(Step 4) Third attribute – Humidity with categorical values of high ($p_1$ =3, $n_1$ =4) and normal ($p_2$ =6, $n_2$ =1)

*Entropy(humidity = high) = -(3/7)\*log₂(3/7) – (4/7)\*log₂(4/7) = 0.983*

*Entropy(humidity = normal) = -(6/7)\*log₂(6/7) – (1/7)\*log₂ (1/7) = 0.591*

E(Humidity), weighted-average entropy for Humidity according to Eq. (10.34)

= p(high)\*entropy (humidity = high) + p(normal)\*entropy (humidity = normal)

= (7/14)*0.983 + (7/14)*0.591 =**0.787**

Information gain according to Eq. (10.39) = Entropy(S) – E(Temperature) =0.94 – 0.787
=**0.153**

Gain ratio according to Eq. (10.40) = 0.153/0.787 = **0.1944**

(Step 5) Fourth attribute – Wind with categorical values of strong ($p_1$ =3, $n_1$ =3) and weak ($p_2$ =6, $n_2$ =2)

Entropy(Wind = strong) = *-(3/6)\*log$_2$(3/6) – (3/6)\*log$_2$(3/6) = 1*

Entropy(Wind = weak) = *-(6/8)\*log$_2$(6/8) – (2/8)\*log$_2$ (2/8) = 0.811*

Weighted-average entropy for Wind according to Eq. (10.34)

= p(strong)*entropy (wind= strong) + p(weak)*entropy (wind =weak)

= (6/14)*1 + (8/14)*0.811 =**0.892**

Information gain according to Eq. (10.39) = Entropy(S) – E(Wind) =0.94 – 0.892 =**0.048**

Gain ratio according to Eq. (10.40) = 0.048/0.892 = **0.0538**

Up to now, the attribute with the maximum information gain of 0.247 (for ID3 algorithm) and the maximum gain ratio of 0.262 (for C4.5 algorithm) is Outlook (step 2). Figure 10.13 shows the decision tree built so far.

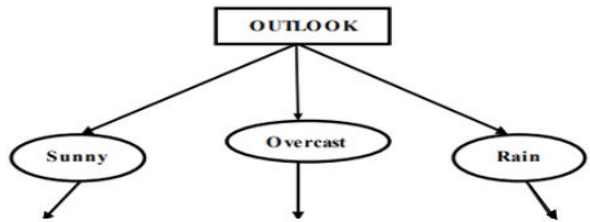

Figure 10.13 Initial decision tree resulting from steps 1 to 5 of ID3 and C4.5 algorithms

Next, we find the best attribute for splitting the data: (1) with Outlook=Sunny values (days 1,2,8,9, and 11); (2) with Outlook =Overcast values (days 3,7,12, and 13); and (3) with Outlook = Rain values (days 4,5,6, 10 and 14). We note that the four instances of Outlook=Overcast (days 3,7,12, and 13) all lead to positive class P or Yes, and there is no need for further analysis. We can direct these Outlook-Overcast instances to a leaf node or final output node of Yes. We can apply the same procedure as in Steps 2 to 5 to analyze the Outlook-Sunny and Outlook-Rain instances to develop the decision tree further.

From step 2, we see: Entropy (Outlook-Sunny) = **0.971**.

In Steps 6 to 8, we analyze the three attributes (temperature, humidity and wind) within Sunny.

(Step 6) First attribute- Temperature with categorical values of hot ($p_1$ =0, $n_1$ =2), mild ($p_2$ =1, $n_2$ =1) and cool ($p_3$ =1, $n_3$ =0)

Entropy(Sunny-Temperature=hot) == $-(0/2)*log_2(0/2) - (2/2)*log_2 (2/2) = 0$

Entropy(Sunny-Temperature=mild) == $-(1/2)*log_2(1/2) - (1/2)*log_2 (1/2) = 1$

Entropy(Sunny-Temperature = cool) == $-(1/1)*log_2(1/1) - (0/1)*log_2 (0/1) = 0$

E(Temperature), weighted-average entropy for Temperature according to Eq. (10.35)

= p(Sunny-Temperature=hot)*entropy (Sunny-Temperature=hot) + p(Sunny-Temperature=mild)*entropy(Sunny-Temperature=mild) +p(Sunny-Temperature = cool)*entropy(Sunny-Temperature = cool)

= (2/5)*0 + (2/5)*1 + (1/5)*0 = **0.4**

Information gain according to Eq. (10.35) = Entropy(Outlook-Sunny) − E(Sunny-Temperature) =0.971− 0.4= **0.571**

Gain ratio according to Eq. (10.36) = 0.571/0.971 = **0.588**

(Step 7) Second attribute- Humidity with categorical values of high ($p_1$=3, $n_1$ =0) and normal ($p_2$ =2, $n_2$ =0)

Entropy(Sunny-Humidity=high) == $-(3/3)*log_2(3/3) - (0/3)*log_2 (0/3) = 0$

Entropy(Sunny-Humidity=normal) == $-(2/2)*log_2(2/2) - (0/2)*log_2 (0/2) = 0$

E(Temperature), weighted-average entropy for Temperature according to Eq. (10.34)

= p(Sunny-Humidity=high)*entropy (Sunny-Humidity=high) + p(Sunny-Humidity=normal)*entropy(Sunny-Humidity=normal)

= (3/5)*0 + (2/5)*0 = **0.0**

Information gain according to Eq. (10.35) = Entropy(Outlook-Sunny) − E(Sunny-Humidity) = 0.971− 0.0= **0.971**

Gain ratio according to Eq. (10.36) = 0.971/0.971 = **1.0**

(Step 8) Third attribute- Wind with categorical values of strong ($p_1$ =1, $n_1$ =1) and weak ($p_2$ =2, $n_2$ =1)

Entropy(Sunny-Wind=strong) == $-(1/2)*log_2(1/2) - (1/2)*log_2 (1/2) = 1$

Entropy(Sunny-Wind=weak) == $-(2/3)*log_2(2/3) - (1/3)*log_2 (1/3) = 0.918$

E(Wind), weighted-average entropy for Temperature according to Eq. (10.34)

= p(Sunny-Wind=strong)*entropy (Sunny-Wind=strong) + p(Sunny-Wind=weak)*entropy(Sunny-Wind=weak)

= (2/5)*1 + (3/5)*0.918 = **0.9508**

Information gain according to Eq. (10.39) = Entropy(Outlook-Sunny) − E(Sunny-Wind) = 0.971− 0.9508= **0.0202**

Gain ratio according to Eq. (10.40) = 0.0202/0.971 = **0.0208**

From steps 6 to 9, the attribute with the maximum information gain of 0.971 (for ID3 algorithm) and the maximum gain ratio of 1.0 (for C4.5 algorithm) is Humidity (step 7). Figure 10.14 shows the decision tree built so far.

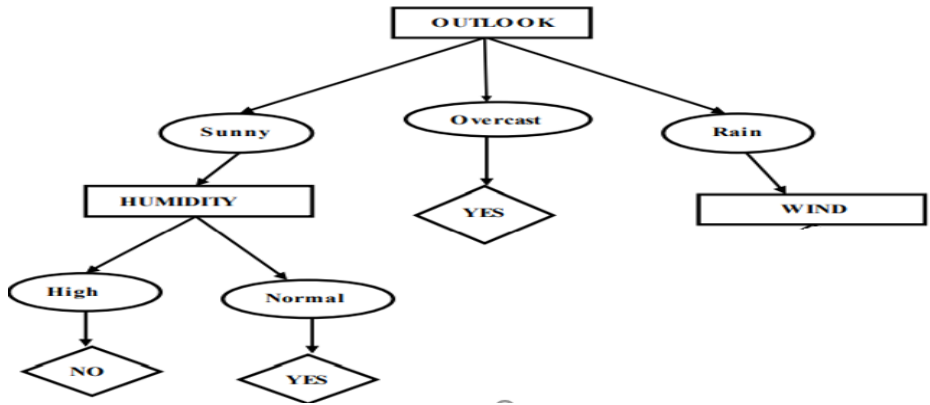

Figure 10.14 Decision tree resulting from steps 6 to 9 of ID3 and C4.5 algorithms

The reader should be able to analyze the two attributes (temperature and wind) within Rain, and obtain the final decision tree, as shown previously in Figure 10.12.

**(4) Other Aspects of Decision Trees**

First, we note that the ID3 algorithm handles categorical feature values (e.g., high and normal) only, and C4.5 algorithm can deal with both categorical and numerical feature values. Refer to [60] for an example of how to apply the C4.5 algorithm to handle numerical values of humidity in the decision tree of Figure 10.12.

Decision tree pruning combats overfitting. Prunning works by deleting nodes that are not clearly relevant. Fr example, a node whose children are all leaf nodes or final output nodes may be unnecessary [8, p. 181]. We can apply a statistically significant test to check if those nodes with only leaf nodes as descendants. See [1, p.663] for more information.

For additional details of the CART algorithm, refer to [55], and [8, pp. 178-179]. Scikit-Learn uses the CART algorithm, which produces only binary trees, meaning that non-leaf nodes always have two children (that is, questions only have yes/no answers). Other algorithms, such as ID3, can produce decision trees with nodes having two more children [8, p.177].

Decision trees are also capable of performing regression tasks. See [8, pp. 182-185] for additional details and the Python implementation. In Appendix B of this book, code B.3 and Table B.1 at the end give the Python implementation of the decision tree algorithm, together with a list of common parameters and their suggested values.

Lastly, we refer the reader to Section 10.3, Enhanced Learning by Ensemble Methods,

**10.2.2e Performance Evaluation Metrics for Classification Models**

We follow [7, pp. 54-58; 73] to present the common evaluation metrics.

(a) *Confusion Matrix.*

We adopt an example from [7, p. 55]. Table 10.4 gives a 2x2 confusion matrix that summarizes the performance of a classification model in predicting examples belong to Spam and Not_Spam categories. Of the 20 examples that are actually spam, the model correctly classifies 18 as true positives, TP = 18. The model incorrectly classifies 2 example as Not_Spam, that is, we have two false negatives, FN = 2. Likewise, 356 examples are Not_Spam or true negatives, TN =356; 9 examples are incorrectly classified, that is, false positives, FP = 9.

Table 10.4   A confusion matrix for classification of spam and not-spam categories predicted by a classification model

|  | Spam (predicted) | Not_Spam (predicted) |
|---|---|---|
| Spam (actual) | *18 (true positives, TP)* | *2 (false negatives, FN)* |
| Not_Spam (actual) | *9 (false positives, FP)* | *356 (true negatives, TN)* |

A confusion matrix has the same number of rows and columns as that of classes. It helps us identify the mistake patterns in some applications.

(b) *Precision and Recall*

Precision is the ratio of true positive (TP) predictions to the total number of true positive (TP) and false positive (FP) predictions:

$$\text{Precision} = TP/ (TP + FP) \qquad (10.41)$$

Recall is the ratio of true positive (TP) to the total number of true positive (TP) and false negative (FN) predictions:

$$\text{Recall} = TP/ (TP + FN) \qquad (10.42)$$

(c) *F1 Score*:

F1 score combine precision and recall using the harmonic mean defined by:

$$F1 = 2 \times 1/[(1/\text{precision}) + (1/\text{recall})] \qquad (10.43)$$

Interested readers may refer to our paper [73] about applications of these metrics to an industrial classification problem.

**10.2.3 Unsupervised Learning for Dimensionality Reduction and Clustering Applications**

**10.2.3a   An Overview**

We begin by quoting the thoughtful statement by Geron [8, p. 235]: "Although most of the applications of ML today are based on supervised learning (and as a result, this is where most of the investments go to), the vast majority of the available data is unlabeled: we have input features **X**, but do not have labels **y**. The computer scientist Yann LeCun famously said that "if intelligence was a cake, unsupervised learning would be the cake, supervised learning would be the icing on the cake, and reinforcement would be the cherry on the cake." In other words, there is a huge potential in unsupervised learning that we have barely started to sink our teeth into".

In Chapter 9, we studied two of the most common unsupervised learning tasks: dimensionality reduction and outlier (anomaly) detection by principal component analysis (PCA). For classification applications, we introduced the concepts of K-means clustering in Section 10.2.2b(2), of P-nearest neighbors in Section 10.2.2b(3), and of kernel trick in Section 10.2.2c(3).

 In this section, we first integrate the kernel trick with PCA and present the kernel PCA to enable the complex nonlinear projections for dimensionality reduction and outlier detection.

We then introduce several unsupervised learning methods for identifying clusters and outliers. Table 10.5 summarizes these methods, as adopted from [77].

Table 10.5 A summary of several unsupervised learning methods for identifying clusters and detecting outliers

| Method | Basis | Model Input | Need the Number of Clusters | Cluster Shaped Identified | Applicable to Outlier Detection |
|---|---|---|---|---|---|
| K-means clustering, | Distance between data and centroids | Actual observations | Yes | Spherical shape; with equal diagonal covariance | No |
| Hierarchical clustering | Distance between data | Pairwise distances between observations | No | Arbitrary shapes | No |
| Density-based spatial clustering of applications with noise (DBSCAN) | Density of region in the data | Actual observations; or pairwise distances between observations | No | Arbitrary shapes | Yes |
| Gaussian mixture model | Mixture of Gaussian distributions of data | Actual observations | Yes | Spherical clusters with different covariance structures | Yes |

### 10.2.3.b  Kernel Principal Component Analysis (KPCA)

We follow [8, pp. 226-229]. First, in Section 10.2.2c, we introduce the "kernel trick" for dealing with inherently nonlinear datasets that cannot be separately by a hyperplane in the original space. In Section 10.2.2c(3), we explain the "kernel trick", using kernel functions of Table 10.1 to implicitly map instances into a higher dimensional space, called *feature space*, and enable nonlinear classification and regression using support vector machines (SVMs). Figure 10.11 shows an example of transforming a linear two-dimensional non-separable case into linear separable case in a three-dimensional space.

We can also apply the kernel trick to PCA by applying a kernel function, making it possible to preserve clusters of instances after nonlinear projections.  This is called the *kernel PCA or kPCA,* and is an effective tool for dimensionality reduction and outlier detection.

Table 10.1 lists the typical kernel functions. For the popular kernel function, Gaussian radial basis function (RBF), that is, $\exp(-\Upsilon\|\mathbf{a} - \mathbf{b}\|^2)$, corresponding to vectors of two instances, $\mathbf{a}$ and $\mathbf{b}$, we see an important superparameter, $\Upsilon$. As kPCA is an unsupervised learning algorithm, there is no specific performance measure to help us choose the best kernel function and superparameter value. Typically, we use repeated trials through a grid search to find the kernel and superparameters. Reference [8, pp. 277-281] gives examples of the Python implementation of this kernel and superparameter optimization.

For classification applications, we could reduce the dimensionality to two using kPCA with grid search for finding the best kernel and superparameter, followed by applying logistic "regression" (actually a classification algorithm) of Section 10.2.2a.

### 10.2.3c K-Means Clustering

In Section 10.2.2.b(2), we discussed K-means clustering in the context of supervised learning by radial basis function (RBF) network. Most of concepts presented previously are applicable to unsupervised learning by K-means clustering. For unsupervised learning when we do not have labels for dependent variables nor the cluster centers (or centroids) for features or independent variables, how do we proceed? We follow [7, pp. 110-114; 8, pp. 240-248] to discuss the highlights.

First, we mention that *it is necessary to scale the input features* before we run K-means clustering. Otherwise, the clusters could be very stretched and K-means will perform poorly.

Basically, we start by placing the cluster centers or centroids randomly, that is, by picking *k* instances at random and using their locations as centroids. We then label the instances, update the centroids, and repeat the tasks of labeling the instances and updating the centroids, and so on until the centroids stop moving. Significantly, computational experience shows that the algorithm is guaranteed to converge in a finite number of iterations (usually quite small) and it will not oscillate forever [8, p. 240].

The efficiency of this clustering scheme depends on the initiation of the centroid locations and of the number of clusters. Heuristically, we could run the algorithm ten times with different random initialization of the centroid locations, and keep the best solution [8, p. 242].

Finding the appropriate number of clusters $k$ is the most important question for applying the K-means clustering method. Geron [8, pp. 246-248] introduces an effective tool, called *silhouette score*, which can be readily implemented in Python libraries. The silhouette score is the average silhouette coefficient over all instances. An instance's silhouette coefficient is equal to $(b-a)$/Max $(a, b)$, where $a$ is the average distance between an instance and all other instances in the same cluster, that is, the average intra-cluster distance; and $b$ is the average distance between an instance and all other instances in the next-closest cluster, that is, the average distance to the instances of the next closest cluster as the one that minimizes $b$, excluding the instance's own cluster. The silhouette coefficient varies between -1 and +1. A coefficient close to +1 means that the instance is well inside its own cluster and far from other clusters. A coefficient close to 0 means that it is close to a closer boundary. Lastly, a coefficient close to -1 means that the instance may have been assigned to the wrong cluster. As an illustration, Figure 10.16 illustrates the silhouette scores for different number of cluster adopted from an example in [8, p. 246]. The figure shows that the number of clusters, $k = 4$, is a very good choice; $k = 5$ is quite good as well, and is much better than $k= 6$ or 7. Fortunately, existing Python libraries have codes to enable us to find the silhouette score when we vary the value of $k$.

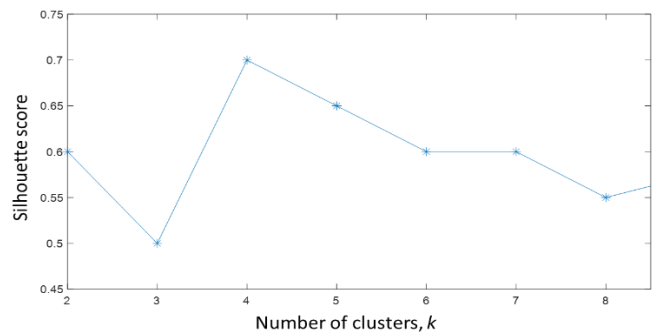

Figure 10.16 Silhouette score versus the number of clusters $k$

Lastly, we note that K-means clustering does not behave very well when the clusters have varying sizes, different densities, or nonspherical shapes. For those with elliptical clusters, Gaussian mixture model that we introduce below would work better [8, p. 248].

### 10.2.3d Hierarchical Clustering

Johnson [78] first proposed the concept of hierarchical clustering. The basic idea is to group data over a variety of scales and create a hierarchical series of nested clusters or a cluster tree, also called *dendrogram*. The cluster tree is not a single set of clusters, but a multilevel hierarchy, where clusters at one level combine to form cluster at the next level. This multilevel hierarchy allows us to choose the level, or scale, of clustering that is most appropriate for our application [78]. Figure 10.17 gives an example of a group of nested clusters (left) and the corresponding dendrogram.

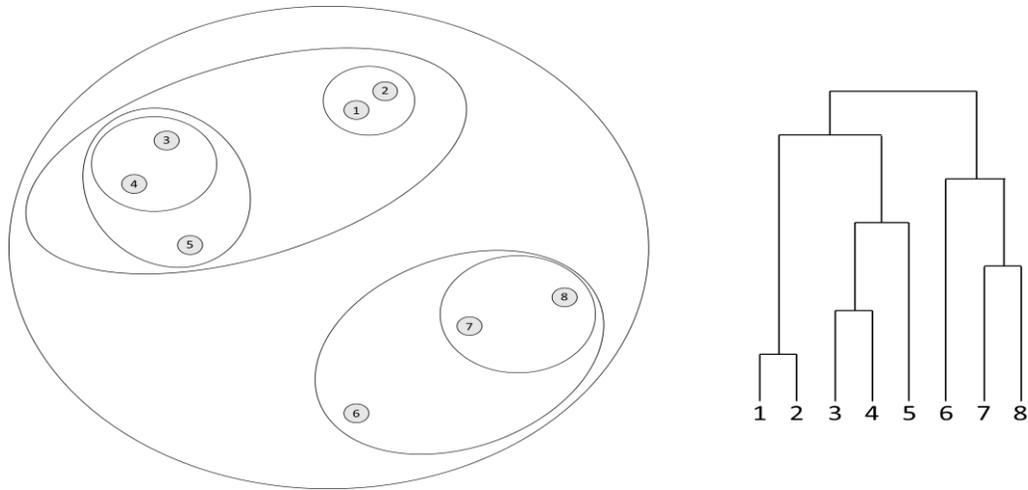

Figure 10.17 Representing nested clusters (left) by a dendrogram (right)

Figure 10.18 illustrates two types of hierarchical clustering. The top diagram in Figure 10.18 illustrates the order that a group of nested clusters is broken up. This represents a top-down view of hierarchical clustering, called *divisive hierarchical clustering (DHC)*. Specifically, we consider all of the data points as a single cluster; and in every iteration, we separate the data points from the clusters which are not comparable. This eventually leads to a number of clusters. The top diagram of Figure 10.18 below demonstrates the concept of divisive hierarchical clustering further. The DHC works best when we have fewer, but larger clusters, and it is computationally expensive and is rarely used.

By contrast, the bottom diagram of Figure 10.18 illustrates the concept of a bottom-up view of hierarchical clustering, called *agglomerative hierarchical clustering (AHC)*. In this case, we consider initially every data point as an individual entity or cluster. At every iteration, we merge these clusters with different cluster until we end up with a single large cluster. The AHC is useful when we have many smaller clusters. It is computationally simpler, more often used and more available.

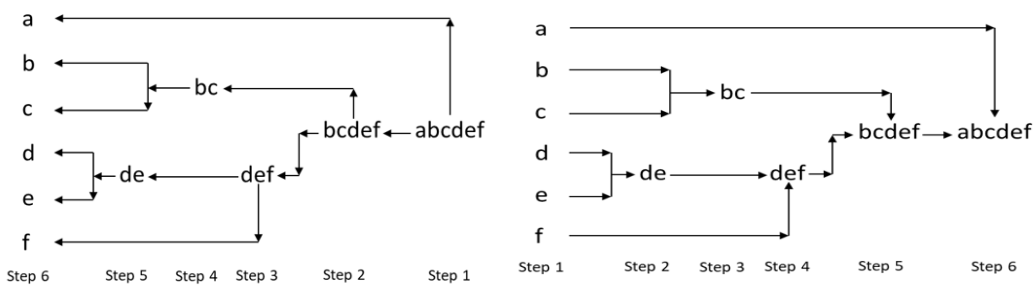

Figure 10.18   Divisive (left) and agglomerative (right) hierarchical clustering

We also caution that hierarchical clustering could lead to a dendrogram that would be wrong. Unless we know the data inside out which is impossible for big datasets, we cannot really avoid this error.  This follows because how we decide to create clusters could lead to significantly different dendrograms, for which we may not be able to tell which result is the most suitable.

The fact that hierarchical clustering would work even if presented with seemingly unrelated data could be a positive as well as a negative. As an illustration, in 2001, medical researchers have applied hierarchical clustering to identify gene expression patterns of breast carcinomas to distinguish tumor subclasses with clinical implications [80]. Interested readers may refer to many online examples of implementing agglomerative and divisive hierarchical clustering using Python.

**10.2.3.e Density-Based Spatial Clustering of Applications with Noise (DBSCAN)**

Easter et al. [79] first proposed a density-based clustering algorithm, called DBSCAN, in 1996. This algorithm overcomes the limitation of the K-means and hierarchical clustering algorithms which fail in creating clusters of arbitrary shapes, and are unable to form clusters based on varying densities. Additionally, when compared to K-means clustering, DBSCAN does not require the number of clusters to be told beforehand.

We follow [7, pp. 112-112] in describing the basic concept of the DBSCAN. Instead of guessing how many clusters we need, we define two hyperparameters, *a limiting distance parameter ε (epsilon) and a limiting number of clusters n*, in DBSCAN. Epsilon is the radius of the cluster to be created around each data point, and n is the minimum number of samples (data points) required inside a cluster for that data point to be classified as a core point of the cluster.

We start by picking an example **x** from our dataset at random and assign it to cluster 1. We then count how many examples (data points) have the distance from **x** less than or equal to ε. If this quantity is greater than or equal to n, then we put all the ε-neighbors to the same cluster 1. We then examine each member of cluster 1, and find their respective e-neighbors. If some member of cluster 1 has n or more ε-neighbors, we expand cluster 1 by adding those ε-neighbors to the cluster. We continue expanding cluster 1 until there are no more examples to put in it. We then pick from the dataset another example not belonging to any cluster and assign it to cluster 2. Continue like this until all examples either belong to some clusters or are marked as *outliers.* An outlier is an example whose ε-neighborhood contains less than n examples.

Figure 10.18 illustrates the circles of an equal radius ε (epsilon) around every data point, with an assumed minimum number of samples required inside a cluster n = 3. We see at least three samples or data points in the middle circles as *the core points (in red)*. There are no other data points around the single date point within the upper right circle; we call this single data point a *noise (in blue)*.

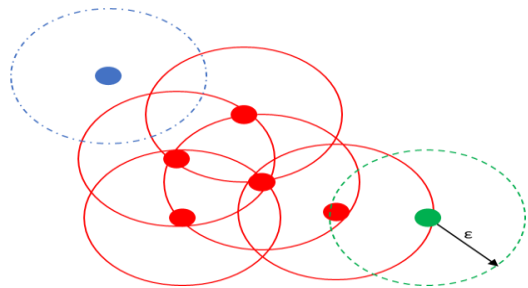

Figure 10.18 An illustration of the core points (red), noise (blue) and
radius (epsilon) in the DBSCAN algorithm

An advantage of DBSCAN over the K-means clustering is that it can build clusters that have an arbitrary shape. However, specifying good values for the two required hyperparameters for DBSCAN could be challenging, and when epsilon is specified, the algorithm cannot effectively deal with clusters of varying density [7, p. 111]. We recommend that interested readers refer to online tutorials and examples of Python implementations of DBSCAN.

**10.2.3f Gaussian Mixture Model (GMM)**

When clusters have different sizes and different correlation structures within them, the K-means clustering algorithm would not be effective. An alternative is to use the Gaussian mixture model (GMM). Here, we follow [1, pp. 738-741; 7, pp. 114-118; 8, pp. 259-274]. A GMM forms clusters as a mixture of multivariate normal density (Gaussian distribution) components. For a given observation, the GMM assigns a posterior probability to each component density (cluster). This probability indicates that the observation has some probability of belonging to the cluster. A GMM can perform hard clustering by choosing the component that maximizes the posterior probability as the assign cluster for the observation. The GMM can also carry out a soft clustering by assigning the observation to multiple clusters based on the scores or posterior probabilities of the observations for the clusters.

There are several variants of the GMM, and we consider the simplest one here. To apply the algorithm, we must know in advance the number of $k$ Gaussian distributions. Note that a standard Gaussian or normal probability distribution function with a mean $\mu$ and a standard deviation $\sigma$ is:

$$f(x) = \frac{1}{\sigma\sqrt{2\pi}} \exp[-\frac{1}{2}(\frac{x-\mu}{\sigma})^2] \qquad (10.44)$$

In Section A.1 of Appendix A of this book, we present the details of a J x K data matrix $\mathbf{X}$, where K is the number of features or independent variables and J is the number of observations or measurements. We also discuss the 1 x K sample mean for each of the K features, $\bar{x}_k$, together with the sample standard deviation $\mathbf{s_k}$ and the sample covariance $s_{jk}$.

Figure 10.19, adopted from [8, p. 260], illustrates the basic structure of the GMM algorithm. We see the feature vector $\mathbf{x}_i$ and the corresponding label $y_i$ (i= 1 to j). If $y_i$= j, it means that the i-th instance has been assigned to the j-th cluster, making the corresponding cluster label as j. The feature vectors within j-th cluster follow a Gaussian (normal) probability distribution with a mean $\mathbf{\mu}_j$ and a covariance matrix $\mathbf{\Sigma}_j$ (j= 1 to k). This is indicated by N($\mathbf{\mu}_j$, $\mathbf{\Sigma}_j$).

In the figure, the circles represent random variables, and the squares represent fixed values. In the two large rectangles, we are to repeat the instance from i= 1 to m, and the cluster from j=1 to k.

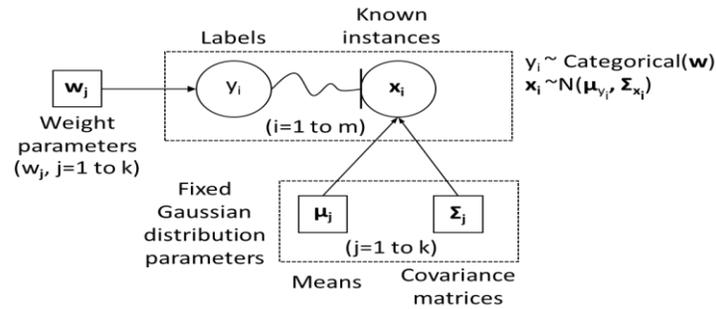

Figure 10.19 The basic structure of the Gaussian mixture model

Given the dataset **X**, we start by estimating the weight factor vector **w** with components $w_j$ (j = 1 to k), and all the Gaussian probability distribution parameters mean $\boldsymbol{\mu}_j$ and covariance matrix $\boldsymbol{\Sigma}_j$ (j = 1 to k). The Python implementation of this procedure is "super easy", according to [8, pp. 261-263], which gives the details of the GaussianMixture code.

Specifically, the algorithm includes two steps: first assigning instances to clusters (called the *expectation* step), and then updating the clusters (called the *maximization* step). Therefore, the core of the GMM algorithm is also called *the EM algorithm*. In the expectation step, the algorithm estimates the probability that each instance belongs to each cluster, based on the current cluster parameters. Then, during th maximization step, the algorithm updates each cluster using all the instances in the dataset, with each instance weighted by the estimated probability that it belongs to the cluster. We call these probabilities the responsibilities of the clusters for the instances, because each cluster's update will be most impacted by the instances it is most responsible for.

Lastly, when applying the GMM algorithm, we cannot use the silhouette score, illustrated previously in Figure 10.15, to choose the appropriate number of clusters, because the clusters in the GMM model are not spherical or have deferent sizes. Reference [8, pp. 267-269] presents the details and Python examples of using different criteria to choose an appropriate number of clusters.

## 10.3 Enhanced Learning by Ensemble Methods

### 10.3.1 An Overview

Conventional learning methods try to develop a single learner from training data. By contrast, *ensemble methods* [61 to 76] train multiple learners to solve the same problem. Specifically, an ensemble contains a number of learners (trained models), called *base learners*, generated from training data by a base learning algorithm. We note that some references call a learned model *a hypothesis* [1, p. 696; 61, p. 2]. The *generalization* of an ensemble refers to the accuracy of the learners or trained models in predicting or classifying new data not used in the training process.

We often apply ensemble methods near the end of a project, once we have already built a few good predictors as our base learners to combine them into an even better predictor [8, p.189].

We focus on ensemble methods for the following reasons: (1) Gradient boosting, an ensemble method that we discuss below, implemented in the popular XGBoost (e**X**treme **G**radient **Boost**ing) package, is routinely used for both large-scale applications in industry for problems with billions of examples, and by the winners of data science competitions [1, p. 702]. In particular, over the past 15 years, ensemble methods have attracted the most attention and won for the most times in the KDD-Cup, which is an annual data mining and knowledge discovery competition sponsored by the ACM special interest group on Knowledge Discovery and Data Mining [61, p. 17]. (2) The winning solutions in machine learning competitions often involve ensemble methods (the most famous in the Netflix Prize competition) [8, p. 189]. (3) According to the experience of Z. H. Zhou, the author of the excellent text, *Ensemble Methods: Foundations and Algorithms,* the best off-the-shelf learning techniques at present are ensemble methods [61, p. 21].

Ensemble methods are appealing mainly because they are able to boost *weak learners*, which are often just slightly better than random guesses, to become strong learners, which can make very accurate predictions, and give almost perfect performance. Schapire [62] has presented a theoretical proof that weak learners can be boosted to become strong learners. As it is easy to obtain base learners or weak learners from training data, but difficult to generate strong learners in practice, ensemble methods present a promising path to generating strong learners.

Additionally, when we try to predict a target variable using any ML technique, the causes of difference in actual and predicted values are *noise, variance and bias*. Ensemble methods do not lower the noise, which is an irreducible error. Depending on the type of ensemble learning algorithm that we use, an ensemble model may result in a lower bias than its component learners, or in a lower variance than its component learners. It would be unlikely for an ensemble model to result in both lower bias and variance than its component learners, because of the well-known bias-variance tradeoff illustrated previously in Figure 10.3.

In general, we develop an ensemble in two steps, first to generate the base learners, which should be accurate and diverse, and then combine them appropriately to become strong learners. In this section, we apply the methods of *bagging, boosting, and stacking* in machine learning practice.

### 10.3.2 Bagging: Random Forest Algorithm

The first method that we apply is *bagging*, which comes the abbreviation of *bootstrap aggregating*, and thus implies two key ingredients of bootstrap and aggregation [65]. In statistics, a sample with replacement is called a *boostrap* [1, p. 697]. Bootstrap involves the random sampling of a small subset of data from the dataset. When sampling is performed with replacement, this method is called *bagging*. When sampling is performed without replacement, it is called *pasting* [8, p. 192]. The method gives an equal probability to selecting all examples in the dataset. Suppose that we have a dataset with N samples and M features. Bagging begins by randomly selecting a sample from the dataset and a subset of features to create a model.

Most of the time, we use a single base learning algorithm so that we have homogenous base learners that are trained in different ways. The algorithm then chooses the features from the subset that gives the best split on the training data, and repeats the same process to create many models and *every model is trained independently in parallel.* Finally, the aggregations of predictions from all the models gives the final prediction.

A popular bagging algorithm is *the random forest algorithm* [67; 1, pp.697-698; 8, pp. 197-199; 61, pp. 57-60], which incorporates a randomized feature selection into the conventional bagging method for decision tree building. During the construction of a component decision tree, at each step of split selection, the algorithm randomly selects a subset of features, and then carries out the conventional split selection procedure within the selected features [61]. To obtain low-bias decision trees, each tree is grown to the maximum size and not pruned back. Random forest yields an ensemble of unstable individual learners, as their structures can drastically change with small changes in input. In general, aggregating the weak learners together can result in *an ensemble model that achieves a lower variance than its component learners, even if the bias could also be reduced.* The algorithm can handle multidimensional datasets and multiclass classification, and work well with noisy data [67].

With bagging, some instances may be sampled several times for any given predictor, while others may not be sampled at all. By default, the predictor samples $m$ training instances with replacement, where $m$ is the size of the training set. This implies that only about 63.2% of the training instances are sampled on average for each predictor. This follows as $m$ grows, this ratio approaches $1 - \exp[-1] = 63.2\%$ [5, p. 212]. The remaining 36.8% of the training instances that are not sampled are called *out-of-bag (OOB) instances*. Note that they are not the same 36.8% for all predictors. A bagging ensemble can be evaluated using the OOB instances without the need for a separate validation dataset [8, pp. 192-193].

Interested readers may refer to many Python implementations of random forest algorithm available online and in [8, pp. 192-198]. In Appendix B of this book, code B.4 and Table B.1 at the end give the Python implementation of the random forest algorithm, together with a list of common parameters and their suggested values.

Workshop 10.1, Section 10.5, illustrates the actual implementation of the random forest algorithm for the prediction of melt index in a slurry HDPE process. In this and other applications, the common procedure is as follows [73].

(1) Use existing libraries like Scikit-Learn to import RandomForestRegressor.

(2) Load the required dataset and preprocess the data by following the steps for data cleaning, integration, and transformation.

(3) Split the dataset into test, validation, and training sets.

(4) Train the model and make baseline predictions.

(5) Tune the hyperparameters using GridSearch if advanced computing resources are available or using iteration methods with the aim of maximizing accuracy, sppeding up processing, and minimizing overfitting. For Random Forest, GridSearch is preferred to select the right hyperparameters.

(6) Evaluate the model based on techniques like k-fold cross validation or out-of-bag (OOB) instances and asses the performance using metrics like root-mean squared error (RMSE) for accuracy.

Our recent article [73] explains more fundamentals and gives the details of each step of this implementation applied to an industrial fermentation process.

### 10.3.3 Boosting: AdaBoost and XGBoost Agorithms

The most popular ensemble method is called *boosting* (originally called *hypothesis boosting*), which refers to a family of algorithms that are able to convert weak learners to strong learners. Briefly, boosting works by building a learner from the training data, and then creating a second learner that attempts *to correct the errors from the first learner*. We add learners *sequentially*, where the later learners focus more on the errors of the earlier learners. We stop adding learners when the training dataset is predicted almost perfectly, or a pre-specified maximum number of learners is reached. Roughly speaking, boosting the weak learners together can result in *an ensemble model that achieves a lower bias than its component learners*, even if variance could also be reduced.

We note that boosting is all about teamwork. Each model that runs will dictate what features the next model will focus on. It utilizes the weighted averages to make weak learners into strong learners. By contrast, bagging has each model run independently, and then aggregates the outputs at the end without giving preference to any model.

A well-known early boosting algorithm is *the adaptive boosting (AdaBoost) algorithm* presented by Freund and Schapire in 1995 [69] initially for classifying datasets. This paper was recognized by the prestigious Gödel Prize in 2003 for outstanding papers in theoretical computer science that is sponsored jointly by the European Association for Theoretical Computer Science (EATCS) and the Special Interest Group on Algorithms and Computation Theory of the Association for Computing Machinery (ACM SIGACT).

When training an AdaBoost classifier, the algorithm first trains a base classifier, such as a decision tree, and uses it to make predictions on the training set. The algorithm then increases the relative weights of misclassified training instances, and again makes predictions on the training set, updates the instance weights, and so on. Basically, the first classifier gets many instances wrong, so their weights get boosted. The second classifier therefore does a better job on these instances, and so on [8, pp. 199 to 200].

Given a training set of N examples (features plus class labels), and a base learning model (e.g., a decision tree), AdaBoost trains a sequence of T base models or weak learners on T different sampling distributions defined upon the training set (D). The algorithm constructs a sample distribution $D_t$ for building the model t by modifying the sample distribution $D_{t-1}$ from the (t-1)-th step. In particular, examples classified incorrectly in the previous step receive higher weights in the new data, attempting to cover misclassified samples. Essentially, AdaBoost fits a sequence of weak learners on repeatedly modified versions of the training dataset, and then combines the predictions from all the weak learners through a weighted sum to give a more accurate, final prediction.

The other popular boosting algorithm is *the gradient boosting algorithm* presented by Friedman [70], who casts the boosting algorithm into a statistical framework as a numerical optimization problem. Just like AdaBoost, gradient boosting works by sequentially adding predictors to an ensemble, each one correcting its predecessors. However, instead of tweaking the instance weights at every iteration like AdaBoost does, gradient boosting tries to fit the new predictor to *the residual errors* made by the previous predictor [8, p. 203].

In the algorithm, one weak learner is added at a time, and existing learners in the model are frozen and left unchanged. This objective is to minimize the loss function of the model by adding weak learners using a gradient-decent-like procedure. The algorithm is attractive because it allows the use of arbitrary differentiable loss functions, thus expanding the boosting technique beyond traditional binary classification problems to support regression, multiclass classification, and more.

Xgboost (eXtreme Gradient Boosting) is an optimized implementation of gradient boosting available in the popular Python library. It was first developed by Chen et. al [71]. The advancements in Xgboost includes regularization, flexibility, parallel processing, better tree pruning and cross validation at each iteration of the boosting process. Xgboost is designed to handle sparse data very well, and is often an important element of the winning entries in ML competitions [8, p. 207].

The reader may refer to [8, pp. 202-207] and many online resources for examples of Python implementations of the AdaBoost and XGBoost algorithms.

The common procedure to implement XgBoost is similar to implementing random forest algorithm. It is as follows [73].

(1) Install XGBoost for use with Python.

(2) Load the required dataset and preprocess the data by following the steps for data cleaning, integration, and transformation.

(3) Split the dataset into test, validation, and training sets.

(4) Train th model and make baseline predictions.

(5) Tune the hyperparameters based on iteration or GridSearch with the aim of maximizing accuracy, accelerating processing, and minimizing overfitting. For XGBoost, a lower learning rate and a higher number of boosting rounds accomplish the task.

(6) Evaluate the model based on techniques like k-fold cross validation, and asses the performance using metrics like root-mean squared error (RMSE) for accuracy.

Our recent article [73] explains more fundamentals and gives the details of each step of this implementation applied to an industrial fermentation process.

We refer the reader to Section 10.6, Workshop 10.1, prediction of HDPE melt index using random forest and extreme gradient boosting ensemble learning methods.

### 10.3.4 Stacking: Stacked Regression

The third method that we apply is stacking [63,64], which is a general procedure where a learner is trained to combine the individual learners. We call the individual learners *the first-level learners,* and name the combined learner *the second-level learner* or *meta-learner.* To carry out stacking, we first train the first-level learners using the original training dataset, and then generate a new dataset for training the second-level learner [66]. In particular, we specify the outputs of the first-level learner as input features to the second-level learner, and retain the original labels as labels of the new training dataset.

One way of applying stacking to regression, or *stacked regression* [64], is to form linear combinations of different predictors to give improved prediction accuracy. Specifically, we use different learning algorithms, with the complete training dataset, to construct *the first-level learners.* Then, a learning algorithm is used to train *a second-level learner* or *a meta-regressor* based on the first-level outputs. The idea is to use cross-validation and least squares under the constraint that all regression coefficients are non-negative, to determine the coefficients in the combination. This non-negativity constraint guarantees that the performance of the stacked ensemble would be better than selecting the single best learner [64].  In his work introducing the concept of *a super learner*, van der Lann [72] theoretically proves that the use of cross validation to create an optimal learner by a weighted combination of many candidate learners. We use open-source machine learning libraries ML extend for stacking regressor model along with the Scikit learn library for the data analysis.

Interested readers may refer to [8, p. 211] and many online resources for Python implementation of the stacking algorithm.

### 10.4 Enhanced Learning by Deep Neural Networks

### 10.4.1   Relevant Concepts of Conventional Neural Networks for Deep Learning Applications

### 10.4.1a   Basic Concepts and Parameters for a Multilayer Perceptron (MLP)

In Section 8.3.2a and Figure 8.101, we introduce the basic processing element (node or neuron) of a neural network. We did this to provide the context to introduce the nonlinear state-space bounded derivative network (SS-BDN) for noninear model-predictive control of polyolefin processes. In the following, we update the discussion in [4] to review the key concepts and parameters in a conventional neural network, *focusing on their limitations and required changes when applied to a deep neural network (DNN)*. The word "deep" in deep learning refers to the number of hidden layers, that is, the depth of the neural network. Basically, every neural network with an input layer, an output layer, and *two or more* hidden layers is *a deep neural network.* In addition to referring to the number of hidden layers as "*depth*", we call the number of nodes or neurons in each layer as "*width*".

Figure 10.20 shows a three-layer perceptron. A perceptron is defined as a neural network with only feedforward interlayer connections. It has no interlayer or recurrent connections, as seen in the connection options in Figure 10.21.

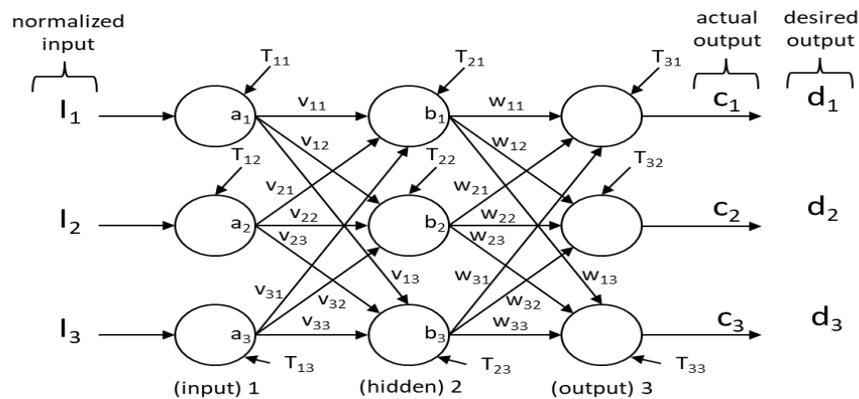

Figure 10.20   A three-layer perceptron.

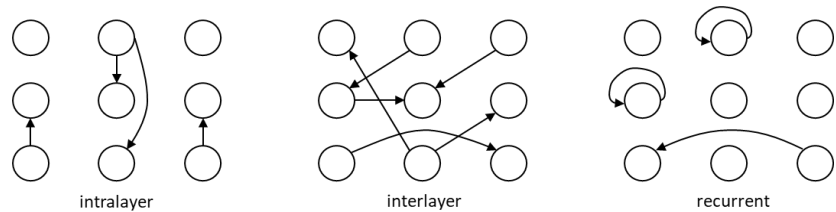

Figure 10.21 The connection options in a neural network.

We consider the neural network of Figure 10.20 for fault diagnosis of process data from a chemical reactor. Table 10.5 lists the input variables (features) and output variables (labels) for the network. For the inputs, we normalize the input values by dividing the actual input variable value by its maximum value, thus limiting each input value to a finite range, [0,1]. As included in Figure 10.20, the desired output, d, from the network is Boolean with 0 indicating no operational fault detected, and 1 indicating an operational fault detected, under the given conditions of input variables.

Table 10.5 Input and output variables for the three-layer perceptron of Figure 10.19

| Input variable (feature) | Output variable (label) |
|---|---|
| $I_1$: reactor inlet temperature, °F | $c_1$ : low conversion |
| $I_2$: reactor inlet pressure, psia | $c_2$ : low catalyst selectivity |
| $I_3$: feed flow rate,  lb/min | $c_3$ : catalyst sintering |

The backpropagation algorithm consists of a forward activation flow and a backward error propagation to optimally select the network parameters, such as weight factors between neurons and internal thresholds of neurons.

We wish to train the network of Figure 10.20 to recognize the following specified conditions: $I_1$= 300/1000 °F = 0.3, $I_2$=100/1000 psia = 0.1, $I_3$=200/1000 lb/min= 0.2; $c_1$= 1 (low conversion), $c_2$ = 0 (no problem), $c_3$= 0 (no problem).

(1) Forward activation flow: Forward output prediction

Step 1. Assume initial values of weight factors, $[v_{ij}]$ and $[w_{ij}]$, and internal thresholds, $[T_{ij}]$, within the range          [-1,+1].

$$[v_{ij}] = \begin{bmatrix} -1 & -0.5 & 0.5 \\ 1 & 0 & -0.5 \\ 0.5 & -0.5 & 0.5 \end{bmatrix} \qquad [w_{ij}] = \begin{bmatrix} -1 & -0.5 & 0.5 \\ 1 & 0 & 0.5 \\ 0.5 & -0.5 & 0.5 \end{bmatrix} \qquad [T_{ij}] = \begin{bmatrix} 0 & 0 & 0 \\ 0.5 & 0 & -0.5 \\ 0 & 0.5 & -0.5 \end{bmatrix}$$

Step 2. Introduce the input vector **I** into the network of Figure 10.20. Calculate the output from the input layer, using a sigmoid transfer function:

$$x_i = I_i - T_{1i} \qquad a_i = f(x_i) = \frac{1}{1+ e^{-x_i}} \qquad (10.45)$$

Substituting the values of $I_1$= 0.3, $I_2$ = 0.1, $I_3$ = 0.3, $T_{11} = T_{12} = T_{13}$ =0, we find $a_1$ = 0.57444, $a_2$ = 0.52498, and $a_3$ = 0.54983.

Step 3. Given the output from the input layer, find the output from the hidden layer:

$$b_j = f\left(\sum_1^3 (v_{ij}\, a_i) - T_{2j}\right) \tag{10.46}$$

where f( ) is the sigmoid transfer function. We note that the terms inside the parenthesis represents *the total activation*; when the weighted sum of the inputs is larger than the internal threshold $T_{2j}$, we have a node firing, which generates a nonzero output from the hidden layer. This equation gives:

$b_1 = f\ (v_{11}a_1 + v_{21}a_2 + v_{31}a_3 - T_{21})$

$\quad = f\ [(-1)(0.57444) + (1)(0.52498) + (0.5)(0.54983)\text{-}0.5]$

$\quad = f\ (-0.27455) = 0.43179$

$b_2 = f\ (v_{12}a_1 + v_{22}a_2 + v_{32}a_3 - T_{22}) = f\ (-0.56214) = 0.36305$

$b_3 = f\ (v_{13}a_1 + v_{23}a_2 + v_{332}a_3 - T_{23}) = f\ (0.79965) = 0.68990$

Step 4. Given the output from the hidden layer, find the output from the output layer:

$$c_k = f\left(\sum_1^3 (w_{jk}\, b_j) - T_{3k}\right) \tag{10.47}$$

This gives:

$c_1 = f\ (w_{11}b_1 + w_{21}b_2 + w_{31}b_3 - T_{31}) = f\ (0.27621) = 0.56862$ (actual value, $d_1 = 1$)

$c_2 = f\ (w_{12}b_1 + w_{22}b_2 + w_{32}b_3 - T_{32}) = f\ (-1.06085) = 0.25715$ (actual value, $d_2 = 0$)

$c_3 = f\ (w_{13}b_1 + w_{23}b_2 + w_{33}b_3 - T_{33}) = f\ (1.24237) = 0.77598$ (actual value, $d_3 = 0$)

(2) Backward error propagation

Step 5. Find improved network parameters ($v_{ij}$, $w_{ij}$ and $T_{ij}$) by minimizing the MSE (mean squared error), commonly called *the loss function* in training neural networks:

$$\text{Loss function} = \text{MSE} = \sum_k \varepsilon_k^2 = \sum_k (d_k - c_k)^2 \tag{10.48}$$

In practice, we slightly modify the MSE equation by incorporating a gradient decent term of the transfer function. Specifically, we multiply the output error $(d_k - c_k)$ by a term of $c_k(1 - c_k)$, which results from differentiating the sigmoid transfer function:

$$f(x_k) = \frac{1}{1 + e^{-x_k}} = c_k$$

The partial derivative is:

$$\frac{\partial f}{\partial x_k} = \frac{e^{-x_k}}{(1 + e^{-x_k})^2} = \frac{1}{1 + e^{-x_k}}\left(1 - \frac{1}{1 + e^{-x_k}}\right) = c_k(1 - c_k) \tag{10.49}$$

For example, when we propagate the output error backward, or backpropagate the output error from the output layer, the error term is:

$\varepsilon_1 = c_1\ (1 - c_1)\ (d_1 - c_1) = 0.10581$

$$\varepsilon_2 = c_2 (1 - c_2)(d_2 - c_2) = \text{-004912}$$

$$\varepsilon_3 = c_3 (1 - c_3)(d_3 - c_3) = \text{-0.13489}$$

Step 6. Continue backpropagation from the output layer to the hidden layer. We find the j-th component of the error vector, $\varepsilon_j$ of the hidden layer relative to each $\varepsilon_k$ using the following equation:

$$\varepsilon_j = b_j (1 - b_j) \left( \sum_1^3 ( w_{jk} \, \varepsilon_k ) \right) \tag{10.50}$$

This equation applies the gradient term, $b_j (1 - b_j)$, to calculate the relative eror.

Applying this equation gives the relative errors of the hidden layer:

$$\varepsilon_1 = b_1 (1 - b_1)(w_{11}\varepsilon_1 + w_{12}\varepsilon_2 + w_{13}\varepsilon_3) = \text{-0.03648}$$

$$\varepsilon_2 = b_2 (1 - b_2)(w_{21}\varepsilon_1 + w_{22}\varepsilon_2 + w_{23}\varepsilon_3) = 0.008872$$

$$\varepsilon_3 = b_3 (1 - b_3)(w_{31}\varepsilon_1 + w_{32}\varepsilon_2 + w_{33}\varepsilon_3) = 0.002144$$

Step 7. Adjust weight factors according to the following formula:

$$\begin{bmatrix} new\ weight \\ factor \end{bmatrix} = \begin{bmatrix} old\ weight \\ factor \end{bmatrix} + \begin{bmatrix} learning \\ rate \end{bmatrix} x \begin{bmatrix} input \\ term \end{bmatrix} x \begin{bmatrix} gradient - descent \\ correction\ term \end{bmatrix} \tag{10.51}$$

or

$$w_{jk,\ new} = w_{jk} + \eta_3\ b_j\ \varepsilon_k\ =\ w_{jk} + \eta_3\ b_j\ [c_k(1 - c_k)(d_k - c_k)] \tag{10.52}$$

where $\eta_3$ is a positive constant with value between 0 and 1, called *learning rate* of the output layer; the input term is $b_j$ and the term within the bracket is the gradient-descent correction term.

Assume a constant learning rate of $\eta = 0.7$. We adjust the weight factors as follows:

$$w_{11,\ new} = w_{11} + \eta\ b_1\ \varepsilon_1\ =\ -0.9680$$

$$w_{12,\ new} = w_{11} + \eta\ b_1\ \varepsilon_2\ =\ -0.5149$$

$$w_{13,\ new} = w_{11} + \eta\ b_1\ \varepsilon_3\ =\ 0.4953$$

Continuing these adjustments, we find the improved weight factors as follows:

$$[w_{jk}] = \begin{bmatrix} -1 & -0.5 & 0.5 \\ 1 & 0 & 0.5 \\ 0.5 & -0.5 & 0.5 \end{bmatrix} \qquad [w_{jk,new}] = \begin{bmatrix} -0.9680 & -0.5149 & 0.4593 \\ 1.0269 & -0.0125 & 0.4657 \\ 0.5511 & -0.5237 & 0.4349 \end{bmatrix}$$

Step 8. Adjust the internal thresholds for the output layer.

$$T_{3k,\ new} = T_{3k}\ + \eta_3\ \varepsilon_k \tag{10.53}$$

With $[T_{31}\ T_{32}\ T_{33}] = [0\ \ 0.5\ \ \text{-0.5}]$ and a constant learning rate $\eta_3 = 0.7$, $\varepsilon_k$ values from step 5, we find:

$[T_{31,\ new}\ \ T_{32,\ new}\ T_{33,\ new}] = [0.0741\ \ 0.04656\ \ \text{-0.5944}]$.

Step 9. Adjust weight factors, $v_{ij}$, connecting the input and hidden layers according to:

$$v_{ij, \text{new}} = v_{ij} + \eta_2 \, a_i \, \varepsilon_j \qquad\qquad （10.4）$$

With $\eta_2$ of 0.7, we find:

$$[v_{ij}] = \begin{bmatrix} -1 & -0.5 & 0.5 \\ 1 & 0 & -0.5 \\ 0.5 & -0.5 & 0.5 \end{bmatrix} \qquad\qquad [v_{ij,\text{new}}] = \begin{bmatrix} -1.0147 & -0.4964 & 0.5009 \\ 0.9866 & 0.0033 & -0.4992 \\ 0.4860 & -0.4966 & 0.5008 \end{bmatrix}$$

Step 10. Adjust the thresholds for the hidden layer, $T_{2j}$ in the hidden layer, following the equation:

$$T_{2j, \text{new}} = T_{2j} + \eta_2 \, \varepsilon_j \qquad\qquad （10.55）$$

This gives

$$T_{21, \text{new}} = T_{21} + \eta_2 \varepsilon_1 = 0.4745$$

$$T_{22, \text{new}} = T_{22} + \eta_2 \varepsilon_2 = 0.0062$$

$$T_{23, \text{new}} = T_{22} + \eta_2 \varepsilon_2 = -0.4985$$

Step 11. Repeat steps 2 to 10 until the MSE or the output-error vector $\boldsymbol{\varepsilon}$ is zero or sufficiently small. It takes 3860 steps for deviation from the desired output $d_k$ to be less than 2%. We find:

Desired output: $[d_1 \, d_2 \, d_3] = [1 \ 0 \ 0]$     Actual output: $[c_1 \, c_2 \, c_3] = [0.9900 \ 0.0156 \ 0.0098]$

Percent error: $[\varepsilon_1 \, \varepsilon_2 \, \varepsilon_3] = [1.00\% \ 1.56\% \ 0.98\%]$

**10.4.1b Incorporation of a Momentum Coefficient**

An improvement of the basic backpropagation algorithm illustrated in Example 10.1 is to use a technique known as *momentum* to speed up the training. Momentum is an extra weight factor added onto the weight factors when they are adjusted. By accelerating the change in the weight factors, we improve the training speed. Figure 10.22 illustrates the concept of the momentum [4]. Suppose that we wish to reach the hill bottom B which represents our global minimum. On the left plot of Figure 10.22, we encounter a rise in our downward path, and we may not be able to continue going downhill and get stuck at the local minimum A. One way to overcome this is to introduce an external momentum to help push us over the rise and eventual reach the hill bottom B, our global minimum, as illustrated in the right plot of Figure 10.22. To implement this concept, we add a momentum term to Eq. (10.41) as follows:

$$\begin{bmatrix} new\ weight \\ factor \end{bmatrix} = \begin{bmatrix} old\ weight \\ factor \end{bmatrix} + \begin{bmatrix} learning \\ rate\ \eta \end{bmatrix} \text{x} \begin{bmatrix} input \\ term \end{bmatrix} \text{x} \begin{bmatrix} gradient - descent \\ correction\ term \end{bmatrix} +$$

$$\begin{bmatrix} momentum \\ coefficient \\ \alpha \end{bmatrix} \begin{bmatrix} previous \\ weight \\ change \end{bmatrix} \qquad (10.56)$$

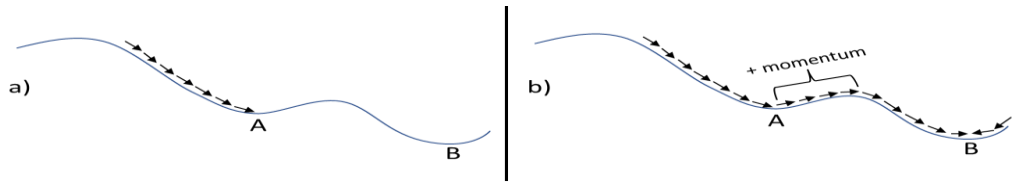

Figure 10.22   An illustration of the concept of adding momentum to help reach a global minimum

Both the learning rate $\eta$ and momentum coefficient $\alpha$ have values greater than zero, and less than or equal to one. *Heuristically, good values to use are: $\eta$ = 0.4 to 0.7; $\alpha = 0.4$ [4].* We introduce their concepts here, as they are important "hyperparameters" in the training of deep neural networks. Not that we refer to properties in the learning algorithm that must be set *before* training as *hyperparameter*, such as the learning rate in the backpropagation algorithm. Hyperparameter is different from a *parameter* that the algorithm is trying to learn and optimize during training; for example, those parameters are the weights and biases in a neural network model. ML model. In fact, Geron [8, p. 325] claims that the learning rate is arguably the most important hyperparameter for training a multilayer perceptron (MLP).

### 10.4.1c Alternative Basic Network Configuration Using Bias Inputs Instead of Internal Thresholds

In Eqs. (10.46) to (10.47), we subtract the internal thresholds $T_{ij}$ (i,j = 1 to 3) from the weighted sum of inputs to each node or neuron, as the input to the transfer function to generate the output from each node. Figure 10.23 shows an alternative basic network configuration that has an extra "bias" input from a "bias" node that is fixed to +1 and weight factors ($v_{01}$, $v_{02}$, $v_{03}$, $w_{01}$, $w_{02}$ and $w_{03}$) going from that node to the nodes in the hidden and output layers.  By doing so, we may re-write Eq. (10.46) as follows:

$$b_j = f\left(\sum_1^3 (v_{ij}\, a_i) - T_{2j}\right) \tag{10.46}$$

$$b_j = f\left(\sum_1^3 (v_{ij}\, a_i) + v_{0j}\, a_0 \right) \tag{10.57}$$

$$b_j = f\left(\sum_0^3 (v_{ij}\, a_i) \qquad (a_0=1,\ T_{2j} = -v_{0j})\right) \tag{10.58}$$

Following Eq. (10.58), we can re-write Eq. (10.47) in a similar way.

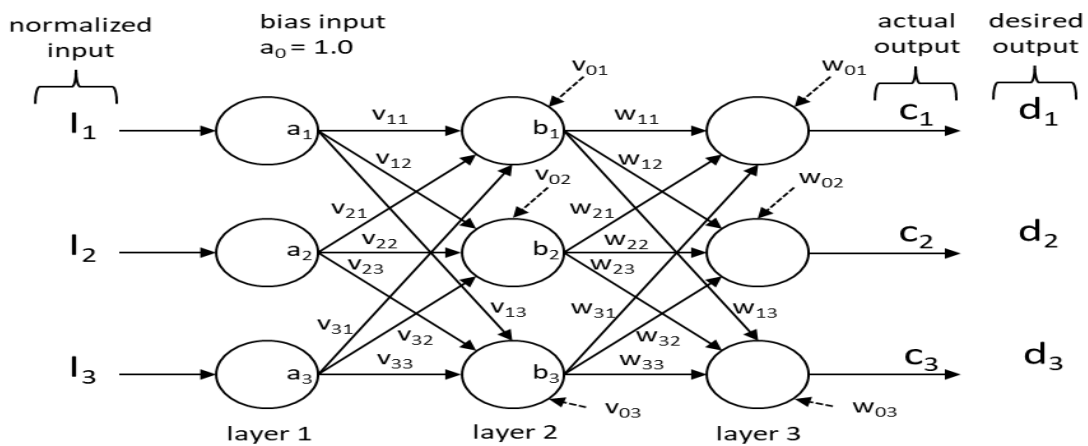

Figure 10.23 Alternative basic network configuration using bias inputs instead of internal thresholds. Weight factors ($v_{01}$, $v_{02}$, $v_{03}$, $w_{01}$, $w_{02}$ and $w_{03}$) connect the bias node to the nodes directed by the dashed arrow.

What is the advantage of using the bias inputs, instead of the internal thresholds?   By setting the initial weight factors ($v_{01}$, $v_{02}$, $v_{03}$, $w_{01}$, $w_{02}$ and $w_{03}$) to be nonzero, the total weighted sum of inputs to each node will be nonzero, even when the outputs of the preceding layer are all zero [1, p.752].

Based on Eqs. (10.57) and (10.58), we can write the output of a fully connected layer in a vector-matrix form:

$$h_{\text{W, b}}(\text{X}) = \varphi(\text{XW} + \text{b}) \qquad (10.59)$$

In the equation,

 **X** = the matrix of input features. It has one row per instance and one column per feature (independent variable);

**W** = the weight matrix, containing all the connection weights, except for the ones from the bias node (neuron). It has one row per input neuron and one column per artificial neuron in the layer.

**b** = the bias vector, containing all the connection weights between the bias neuron and the artificial neurons. It has one bias term per artificial neuron.

**Φ** = the activation function for the neurons in the layer.

Eq. (10.59) will be useful when we discuss the structure of recurrent neural networks below.

### 10.4.1d   Selection of Activation Functions, Vanishing Gradient and Exploding Gradient Problems

Table 10.6 summarizes four activation functions that are commonly used in both conventional and deep neural networks. Figure 10.24 illustrates these functions and their derivatives. We note that for the sigmoid activation function, when the input (x) become large (negative or positive), the output (y) saturates or becomes constant at 0 or 1.   We call this type of activation function *a saturating activation function*. By contrast, the ReLU activation function is *a nonsaturating activation function*. This point is relevant to our discussion in Section10.4.2a relating to fighting unstable gradient problems in the training of deep neural networks.

Table 10.6   Activation Functions and Their Derivatives

| Activation function | Function, y = f(x) | Derivative, y'(x) =f'(x) |
|---|---|---|
| 1.Sigmoid (Logistic) | $\dfrac{1}{1+e^{-x}}$ | $f(x)[1-f(x)]=e^{-x}/(1+e^{-x})^2$ |
| 2. Hyperbolic tangent (tanhx) | $\dfrac{e^x - e^{-x}}{e^x + e^{-x}}$ | $1-[f(x)]^2$ |
| 3. Rectified linear unit (ReLU) | f(x) =0, if x<0; f(x) = x, if x≥ 0 | f'(x) = 0, if x<0; f'(x) = 1, if x≥0 |
| 4. Softplus | $\ln(1 + e^x)$ | $\dfrac{1}{1+e^{-x}}$ (sigmoid) |

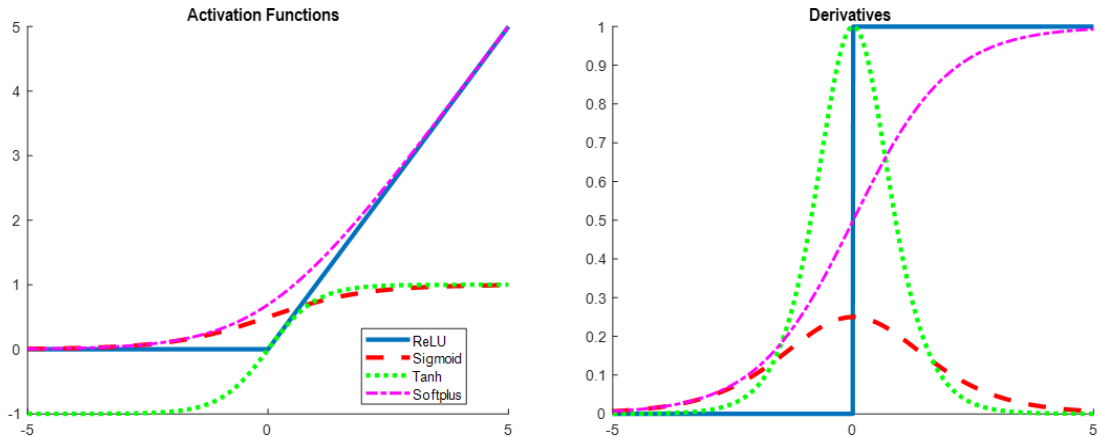

Figure 10.24   An illustration of common activation functions and their derivatives

For the first 25 years of research with multilayer neural networks (roughly 1985-2010), internal nodes used sigmoid and hyperbolic tangent (tanh) activation functions, almost exclusively. From around 2010 onwards, the rectified linear unit (ReLU) and softplus activation functions become more popular, particularly because they have better derivative properties and they can avoid the problems of vanishing/exploding gradients observed when using the gradient descent method to train the networks [1, p. 759].

As demonstrated in the example, during the backward error flow, the backpropagation algorithm works by going from the output layer to the input layer, propagating the error gradient along the way. Once the algorithm has computed the gradient of the loss function (the mean squared error) regarding each parameter (weight factor and internal threshold or bias input) in the network, it applies the gradient to update each parameter with a gradient descent step using, for example, Eqs. (10.4). Unfortunately, the gradient tends to get smaller and smaller as the algorithm progresses approaching the lower layers (i.e., approaching the input layer). Consequently, the gradient descent update leaves the parameters of the lower layers virtually unchanged, and training never converges to a good solution. We call this *the vanishing gradient problem* [8, p. 331].

*The exploding gradient problem* is the inverse of the vanishing gradient and occurs when large error gradients accumulate, resulting in extremely large updates to neural network model weights during training. As a result, the model is unstable and incapable of learning from training data.

As shown in Table 10.7 and in Figure 10.23, when using the ReLU activation function and its smooth variant, softplus activation function, with their positive function and derivative values, the output from a node will always be positive. Using them as activation functions helps avoid the vanishing gradient problem. Research since 2010 has suggested that when training neural networks, particularly deep neural networks with two or more hidden layers, *ReLU is the most used activation function*. Many software libraries and hardware accelerators provide ReLU-specific optimizations. If training speed is your top priority, ReLU should be your best choice as the activation function to use [8, p. 338].

### 10.4.1e   Batch Normalization

We follow [8, p. 338] to introduce the concept of batch normalization, proposed by Ioffe and Szegedy [89], to improve neural network training. While using ReLU activation function and its smooth variant, Softplus, can significantly reduce the vanishing/exploding gradients at the beginning of the training, it does not guarantee that they won't return during training. Batch normization helps avoid these problems by adding an operation in the model just before or after the activation of each hidden layer.

This operation simply zero-centers and normalizes each input (we discuss standardized data by mean-centered data scaled by standard deviation in Section A.1.7 of Appendix A of this book). Batch normalization then scales and shifts the result using two new parameter vectors per layer: on for scaling, and the other for shifting. Essentially, the operation lets the model learn the optimal scale and mean of each layer's inputs. In many cases, you can implement batch normalization (BN) by adding a BN layer as the very first layer of your neural network, and you do not need to standardize your training set (such as using Python code, StandardScaler in Section 10.5.3.c), the BN layer will do it for you.

Ioffe and Szegedy [89] have demonstrated that batch normalization considerably improves all the deep neural networks they experimented with, leading to a huge improvement in image classification network and a significant reduction of the vanishing gradient problem. Batch normalization can also reduce the risk of *overfitting*, that we discussed in Section 10.2.1a. Constraining a model to make it simpler and reduce the risk of overfitting is called *regularization* [8, p. 27], which we discuss more below.

**10.4.1f  Avoid Overfitting through Regularization: Weight Decay and Dropout**

Regularization attempts to limit the complexity of a model, as we want to find the right balance between fitting the training data perfectly and keeping the model simple enough to ensure that it will generalize well with validation data. An approach to regularization is called *weight decay*, which involves adding add a penalty $\lambda(\sum_{i,j}(v_{i,j}^2 + w_{i,j}^2))$, where $v_{i,j}$ and $w_{i,j}$ are weight factors defined in Eqs. (10.46) and (10.47), to the loss function (MSE), Eq. (10.48). $\lambda$ is a hyperparameter of the ML training algorithm controlling the strength of the penalty and the penalty term is to sum the squares of all the weights in the network. Using $\lambda$=0 is equivalent to not using weight decay, while using a larger value of $\lambda$ encourages the weights to become small. A heuristuc value is for $\lambda$ to be near $10^{-4}$ [1, p. 771].

Another popular regularization technique is to use *dropout* [1, p.772; 8, p. 365; 85]. It is a fairly simple algorithm. At every training step, every neuron (including the input neurons, but always excluding the output neurons) has a probability *p* of being temporary "dropped out", meaning it will be entirely ignored during this training step, but it may be active during the next step. The hyperparameter *p* is called *the dropout rate*, and *its heuristic value to use is between 10% to 50%.* For the recurrent neural networks discussed below, *p* is about 20-30%; and for convolutional neural networks discussed below, *p is about 40-50%.* After training, neurons don't get dropped anymore. It has been proven that dropout improves the accuracy of state-of-the-art neural networks by 1-2%. In practice, this is significant, because if a model already has 95% accuracy, getting a 2% accuracy boost means dropping the error rate by almost 40% (going from 5% error to roughly 3%) [8, p. 365].

Finally, Zhou et. al [98] present a detailed hyperparameter optimization study of deep neural networks for absorption, distribution, metabolism, and excretion (ADME) properties of pharmaceutical industries. They tested the following hyperparameters:  learning rate (0.01, 0.1 and 1), weight decay (0, 1E-6, 1E-5, 1E-4, and 1E-3), dropout rate (0, 0.2, 0.4 and 0.6), activation function (ReLU and sigmoid), and batch normalization (with and without). We have already incorporated their observations in our preceding discussion, together with suggestions for heuristic hyperparameter values.

### 10.4.1g  Using a Faster Optimizer

Training a very deep neural network can be painfully slow. There are ways to speed up the training. We refer the reader to [8, pp. 350-359] for a detailed discussion of these methods. We did discuss some of them, such as using a good activation function, ReLU, and applying batch normalization. Another improvement to training soeed comes from using a faster optimizer, instead of the regular gradient-descent algorithm. In particular, we recommend the reader to glance through a classic paper by Kingma and Ba [84] introducing Adam (<u>Ada</u>ptive <u>M</u>oment Estimation), an algorithm for first-order gradient-based optimization of stochastic objective function or loss function.

*Adam optimizer is the recommended scheme for minimizing the loss function for training a deep neural network.* Without going through the specific equations for the algorithm (see [8], p. 356), we note that the default values for three hyperparameters in the algorithm are: learning rate, lr or $\eta$= 0.001; momentum decay parameter $\beta_1$= 0.9; scaling decay hyperparameter $\beta_2$=0.999. Implementing these parameters in Python using Keras is simple:

Optimizer = keras.optimizer.Adam(lr = 0.001, beta_1 =0.9, beta_2 = 0.999)

We demonstrate in Section 10.6.3 how to apply the Adam optimizer using Python in Keras to predict the HDPE melt index using a deep neural network.

### 10.4.1h  Recommended Multilayer Perceptron (MLP) Architecture for Deep Learning Applications

We adopt and expand the recommendations of [8, p. 292] and present Table 10.7 to summarize the typical MLP architecture for deep learning applications.

Table 10.7   Typical architecture for regression MLP for deep learning applications

| Item | Typical value or selection |
|---|---|
| 1. # of input neurons | 1. One per feature (independent variable) |
| 2. #of hidden layers | 2. Typically 2 to 5 |
| 3. # of neurons per hidden layer | 3. Typically 10 to 100 |
| 4. # of output neurons | 4. One per label (dependent variable) |
| 5. Activation function, hidden layers | 5. ReLU or Softplus |
| 6. Output activation function | 6. None |
| 7. Loss function | 7. MSE (mean squared error) |
| 8. Hyperparameter optimization algorithm | 8. Adam (Adaptive momentum estimation) |

| 9. Hyperparameters | 9. Momentum coefficient, learning rate, weight decay, dropout, batch normalization |
| --- | --- |

In Appendix B of this book, code B.6 and Table B.1 at the end give the Python implementation of the deep neural network algorithm, together with a list of common parameters and their suggested values.

We refer the reader to Section 10.7, **Workshop 10.2**, prediction of HDPE melt index by a deep neural network.

Lastly, we share an important empirical finding about deep learning in choosing the number of layers and the number of neurons [1, p. 769]. When comparing two networks with similar numbers of weight factors, a deeper network with a higher number of layers usually gives better generalization performance, showing a higher accuracy in predicting the validation data that the network has not seen previously in its training, than a shallow network with a smaller number of layers.

### 10.4.2 Deep Learning with Recurrent Neural Networks (RNNs)

### 10.4.2a Recurrent Neural Networks for Predictive Modeling of Time-Dependent Processes

Recurrent neural networks (RNN) are mostly used to deal with sequential or time-dependent data types like time-series data. Our previous book [4, pp. 228-364] gives a detailed report of applying RNNs to process forecasting, modeling, and control of time-dependent processes in chemical engineering. References [104 to 120] presents deep learning with RNNs with applications to chemical industries. In Section 10.8, we present Workshop 10.3, demonstrating the prediction of time-dependent HDPE melt index using deep recurrent neural networks.

We can train RNNs by backpropagation through time., which is a popular method for training conventional neural networks [4]. In the following, we follow [8, pp. 499-502] to present the key concepts of deep RNNs.

The left part of Figure 10.25 shows that in recurrent connection, the output from a node feeds into itself. On the very right part of the figure, we see that the recurrent neuron receives the input vector **x**(t) and its own output vector from the previous time step, **y**(t-1). There is no previous output at the first time step, it is set to zero. On the right part of Figure 10.25, we see this recurrent neuron evolved over time.

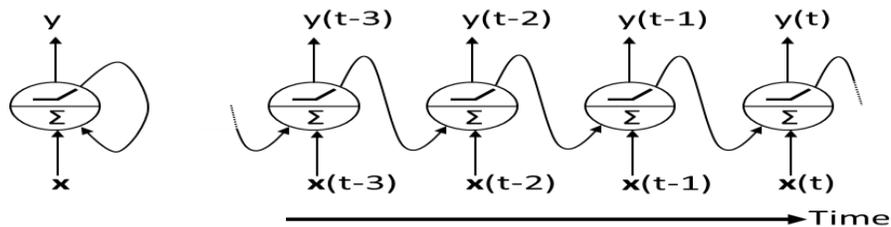

Figure 10.25 The recurrent connection (left) and recurrent neuron evolved over time (right)

For each recurrent neuron, we have two sets of weight factors (say, $\mathbf{w}_x$ and $\mathbf{w}_y$): one for the input vector **x**(t), and the other for the output vector from the previous time step, **y**(t-1). When considering the whole recurrent layer, instead of just one recurrent neuron, we can place all the weight factors in two weight matrices, $\mathbf{W}_x$ and $\mathbf{W}_y$. Following the concept of Eqs. (10.57) to (10.59), we can represent the output vector of the whole recurrent layer for a single instance (i.e., one sample) as follows:

$$\mathbf{y}(t) = \boldsymbol{\varphi}(\ \mathbf{W}_x^T \mathbf{x}(t) + \ \mathbf{W}_y^T \mathbf{y}(t-1)\ + \mathbf{b}) \qquad (10.60)$$

In this equation, $\boldsymbol{\varphi}$ is the activation function, and $\mathbf{b}$ is the bias vector consisting of the bias inputs to all neurons in the recurrent layer.

We can further extend this representation for a layer of recurrent neurons for all instances (samples) in a mini-batch by placing all the input vectors for different instances (samples) at time step t in an input matrix $\mathbf{X}(t)$ and all the output vectors for difference instances (samples) from time step t-1 in an output matrix $\mathbf{Y}(t-1)$. Therefore, we represent the outputs of a layer of recurrent neurons for all instances in a mini batch as follows:

$$\mathbf{Y}(t) = \boldsymbol{\varphi}(\ \mathbf{W}_x^T \mathbf{X}(t) + \ \mathbf{W}_y^T \mathbf{Y}(t-1)\ + \mathbf{b})$$

$$= \boldsymbol{\varphi}(\ \mathbf{X}(t)\mathbf{W}_x + \ \mathbf{Y}(t\text{-}1)\ \boldsymbol{W}_y + \mathbf{b})$$

$$= \boldsymbol{\varphi}[\ \mathbf{X}(t)\ \ \mathbf{Y}(t\text{-}1)]\mathbf{W} + \mathbf{b}) \qquad (10.61)$$

with $\mathbf{W} = [\mathbf{W}_x \ \ \mathbf{W}_y]^{\mathbf{T}}$. Assume that $m$ is the number of instances (samples) in the mini-batch, $n_{neurons}$ is the number of neurons, and $n_{inputs}$ is the number of input features. In Eq. (10.61):

$\mathbf{Y}(t)$ = an $m \ x \ n_{neurons}$ matrix containing the recurrent layer's outputs at time step t for each instance in the mini-batch

$\mathbf{X}(t)$ = an $m \ x \ n_{inputs}$ matrix containing the recurrent layer's inputs at time step t for all instances in the mini-batch

$\mathbf{W}_x$ = an $n_{inputs} \ x \ n_{neurons}$ matrix containing the weight factors for the inputs of the current time step

$\mathbf{W}_y$ = an $n_{neurons} \ x \ n_{neurons}$ matrix containing the weight factors for the output of the previous time step

$\mathbf{b}$ = a vector of size $n_{neurons}$ containing each neuron's bias input

$\mathbf{W} = [\mathbf{W}_x \ \ \mathbf{W}_y]^{\mathbf{T}}$. The weight factor matrices $\mathbf{W}_x$ and $\mathbf{W}_y$ are linked together vertically in a single matrix W of dimension $(n_{inputs} + n_{neurons})\ x \ n_{neurons}$

$[\ \mathbf{X}(t)\ \ \mathbf{Y}(t\text{-}1)]$ = the horizontal concatenation of matrices $\mathbf{X}(t)$ and $\mathbf{Y}(t-1)$

Note that $\mathbf{Y}(t)$ depends on $\mathbf{X}(t)$ and $\mathbf{Y}(t-1)$, which is a function of $\mathbf{X}(t-1)$ and $\mathbf{Y}(t-2)$, which is a function of $\mathbf{X}(t-2)$ and $\mathbf{Y}(t-3)$, and so on. Therefore, $\mathbf{Y}(t)$ depends on all the inputs since time= 0, that is, $\mathbf{X}(0)$, $\mathbf{X}(1)$, $\mathbf{X}(2)$,… $\mathbf{X}(t)$. At time=0, the there are no previous outputs, which are assumed to be all zeros.

Let us extend the backpropagation algorithm discussed in Section 10.4.1 to a time-dependent RNN. Figure 10.26 illustrates the concept of backpropagation through time for training a RNN.

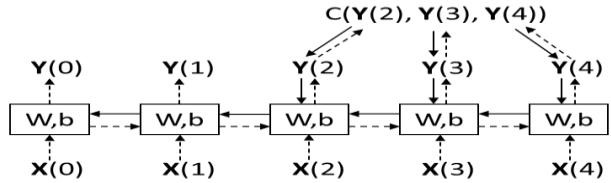

Figure 10.26   An illustration of backpropagation through time

We begin with a forward activation pass through the evolved network from the left to the right, represented by the dashed arrows. We then evaluate the output sequence using a cost or loss function, C [ **Y**(0), **Y**(1), **Y**(2),....**Y**(T)],  where T is the maximum time step.  Next, we proceed with the backward error propagation. The gradients of the cost function are propagated backward through the evolved network (represented by solid arrows). We update the model parameters using the gradients computed during the backpropagation through time.

We compute the cost function using the last three outputs of the network, **Y**(2), **Y**(3) and **Y**(4), so the gradient flows through these three outputs, but not through **Y**(0) and **Y**(1). In other words, the gradient flow background through all the outputs used by the cost function, not just the final output. Lastly, as the weight factor matrix **W** and the bias vector **b** are used at each time step, backpropagation should sum over all time steps.

There are several issues regarding the applicability of the recommendations for deep neural networks discussed in Sections 10.4.1 to 10.4.1h to training RNNs [8, pp. 511-512].  Training an RNN on long sequences with many time steps tends to make the evolved network (see the right part of Figure 10.27) a very deep network. This means that training RNNs may suffer the same unstable (vanishing/exploding) gradient problems that we discussed with deep neural networks.

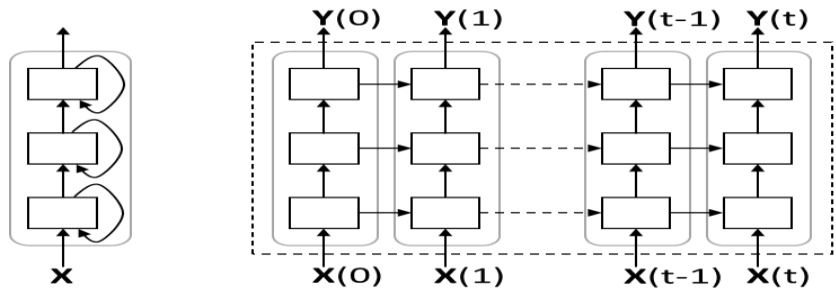

Figure 10.27   A deep RNN (left) evolved through time (right).

Many of the techniques that we discussed in Section 10.4.1 for deep neural networks to overcome the unstable gradient problems are still applicable, but with some cautions and exceptions. First, we can still apply a faster optimizer (Section 10.4.1g) and dropout (Section 10.4.1f). However, we should apply batch normalization (Section 10.4.1e) only between recurrent layers (vertically in Figure 10.27), but not between time steps (horizontally in Figure 10.27). In other words, we could add a batch normalization layer before each recurrent layer, but do not expect too much from it, as research [112] has shown that batch normalization was only slightly beneficial when applied to the inputs at the current time step, but not to the outputs from the previous time step.

For developing deep RNNs, nonsaturating activation functions such as ReLU discussed in Section 10.4.1d may not help much in fighting the unstable gradient problems. In fact, they may actually lead the RNN to be more unstable during training. Why? Suppose that the gradient descent updates the weight factors in a way that increases the output slightly in the first time step. Because the algorithm uses the same weight factors at every time step, the algorithm may slightly increase the outputs at the second time step. This increase in outputs continues in subsequent time steps, and the increasingly large output could eventually "explode". We could reduce this risk by using a smaller learning rate, and by using a saturating activation function like *hyperbolic tangent (tanh), which is recommended as the default activation function for deep RNNs* [8, p.511].

Glorot and Bengio [85] suggest that a better initialization of weight factors could also help in alleviating the unstable gradient problems in RNNs. For both the forward activation flow and the backward error propagation to move properly, the authors argue that we need the variance of the outputs of each layer to be equal to the variance of its inputs, and we need the gradients to have equal variance before and after flowing through a layer in the reverse direction. In practice, we cannot satisfy this requirement unless the layer has an equal number of inputs and neurons (these numbers are called the *fan-in* and *fan-out* of the layer). Glorot and Bengio propose a good weight-factor initialization scheme, called *the Glorot initialization* in Python Keras code that works well in practice. Specifically, initialize the weight factors following a normal distribution with mean 0 and variance $\sigma^2 = 1/fan_{avg} = 2/(fan_{in} + fan_{out})$ [8, p.333].

Bengio et at. [112] proposed the concept of "scaling down the gradient" or "gradient clipping" to tackle vanishing/exploding gradients. Gradient clipping involves clipping the derivatives of the loss function (mean squared error) to have a given value if a gradient value is less than a negative threshold, or more than a positive threshold. For example, we could specify a norm of 0.5, meaning that if a gradient value was less than -0.5, it is set to 0.5; and if it larger than 0.5, then it will be set to 0.5.

Let $\|\mathbf{g}\|$ be the norm of the gradient vector, we replace c ($\mathbf{g}/\|\mathbf{g}\|$) by $\mathbf{g}$, where c is a hyperparameter called the threshold, $\mathbf{g}$ is the gradient, and $\mathbf{g}/\|\mathbf{g}\|$ is a unit vector. After r-scaling, the new gradient will have $\|\mathbf{g}\|$ = c. If $\|\mathbf{g}\| < c$, then we do not need to do anything.

Alternatively, we can visualize gradient clipping as forcing the gradient values (element-wise) to a specific minimum or maximum value if the gradient exceeded an expected range. When the traditional gradient descent method tends to make a large step, the gradient clipping intervenes to reduce the step size to be small enough that it is less likely to go outside the region where the gradient indicates the direction of approximately steepest descent.

In addition to the unstable gradient problems, deep RNNs suffer from a short-term memory problem, which we explain in the following section.

## 10.4.2b Long Short-Term Memory (LSTM) RNNs

This section introduces the concept of the long short-term memory (LSTM) RNNs. We refer the reader to [115 to 117] for three reported applications of the LSTM RNNs to chemical processes, and to Section 10.7 for a workshop for predicting the melt index in a grade transition in a HDPE process.

Because of the data transformation through the transfer function when input data travel through an RNN, the network tends to lose some information at each time step. After a while, the RNN's state could contain virtually no trace of the first inputs. To tackle this problem, there are useful changes to the architecture of the conventional RNN that enable the data information to be preserved over many time steps.

We follow [1, p.775; 8, pp. 514-517; 114] to introduce a special kind of RNN with memory units, called *the long short-term memory (LSTM) cell*, as illustrated in Figure 10.28. Basically, this cell includes three special gates: *a forget gate* to select memory or to forget the irrelevant part of the previous state; *an input gate* to update memory or to selectively update the cell-state values; and *an output gate* to select output or to output certain parts of the cell state. A cell state vector **c**(t-1) goes through these three gating operations, the surviving memory becomes a state vector **c**(t), for which we call it *a long-term state vector*.

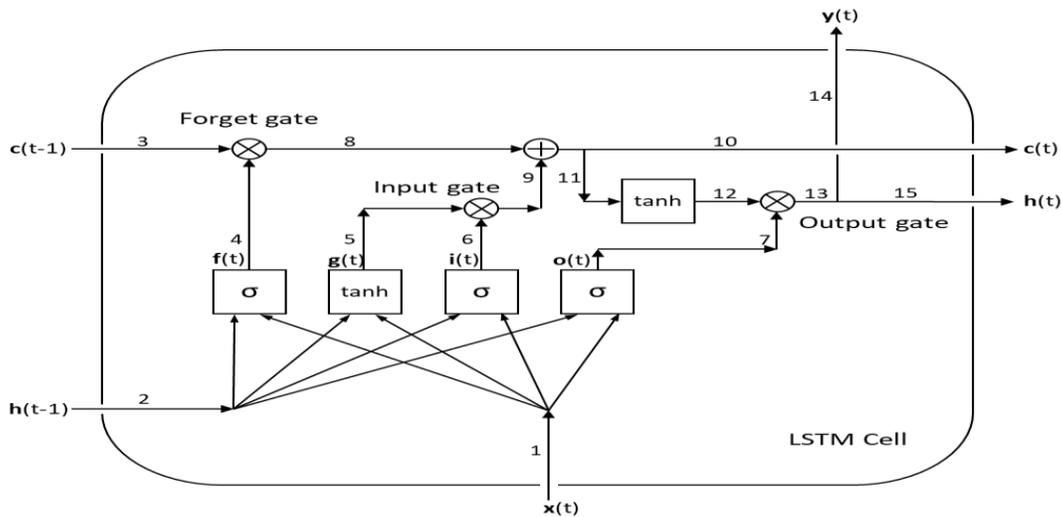

Figure 10.28   An illustration of a LSTM cell.

We see that the input vector **x**(t) at time t (labeled vector 1 in the figure) and the output vector from the previous time step, **h**(t-1) (labeled vector 2) both enter a filtering unit with a sigmoid activation function, giving an output vector **f**(t) (labeled vector 4), which becomes an input vector to the forget gate. Following Eq. (10.60), we represent **f**(t) as:

$$\mathbf{f}(t) = \sigma[\mathbf{W}_{xf}^T \, \mathbf{x}(t) + \mathbf{W}_{hf}^T \, \mathbf{h}(t-1) + \mathbf{b}_f] \qquad (10.62)$$

where σ is the sigmoid activation function with an output vector with scalar components bounded between 0 and 1 (see Figure 10.24). In Eq. (10.62) and the subsequent Eqs. (10.63) to (10.66):

**W**$_{xi}$, **W**$_{xf}$, **W**$_{xo}$, **W**$_{xg}$ are the weight factor matrices for each of the three gates (input, forget, output) and gating unit for their connection to the input vector **x**(t).

*W*$_{hi}$, **W**$_{hf}$, **W**$_{ho}$, **W**$_{hg}$ are the weight factor matrices for each of the three gates (input, forget, output) and gating unit for their connection to the previous short-term state or output vector from the previous time step, **h**(t-1).

**b**$_i$, **b**$_f$, **b**$_o$, **b**$_g$ are the bias terms for each of the three gates (input, forget, output) and gating unit.

Next, we can represent a similar filtering operation to generate an output vector **i**(t) serving as an input vector to *the input gate* (labeled vector 6):

$$\mathbf{i}(t) = \sigma[\mathbf{W}_{xi}^T \mathbf{x}(t) + \mathbf{W}_{hi}^T \mathbf{h}(t-1) + \mathbf{b}_i] \qquad (10.63)$$

We also represent a similar operation to generate an output vector $\mathbf{o}(t)$ which serves as an input vector to *the output gate* (labeled vector 7):

$$\mathbf{o}(t) = \sigma[\mathbf{W}_{xo}^T \mathbf{x}(t) + \mathbf{W}_{ho}^T \mathbf{h}(t-1) + \mathbf{b}_o] \qquad (10.64)$$

Both $\mathbf{x}(t)$ and $\mathbf{h}(t-1)$ also enter *a gating unit* with a hyperbolic tangent (tanh) activation function, giving an output vector $\mathbf{g}(t)$ (labeled vector 5) which becomes an input vector to the input gate. We represent $\mathbf{g}(t)$ as:

$$\mathbf{g}(t) = \tanh[\mathbf{W}_{xg}^T \mathbf{x}(t) + \mathbf{W}_{hg}^T \mathbf{h}(t-1) + \mathbf{b}_g] \qquad (10.65)$$

where tanh is the hyperbolic tangent activation function with an output vector with scalar components bounded between -1 and 1 (see Figure 10.24).

The key idea is that when the long-term state $\mathbf{c}(t-1)$ (labeled vector 3) flows through the network from left to right, it first goes through ***a forget gate f,*** which selects the information that needs to be discarded by combining the previous cell state $\mathbf{c}(t-1)$ (labeled vector 3) to the output vector $\mathbf{f}(t)$ from a filtering unit (labeled vector 4), defined by Eq. (10.58) above, through an element-wise multiplication (as indicated by symbol $\otimes$ below the words "Forget gate" in the figure). This results in an output vector 8 in the figure. Note that we covered the element-wise multiplication, called Hadamard product in Section 8.1.4a, Eq. (8.40).

Next, we drop some memories and then add new memories selected by ***an input gate i***. Specifically, we carry out an element-wise multiplication of output vector $\mathbf{g}(t)$ (labeled vector 5), defined by Eq. (10.55), and output vector $\mathbf{i}(t)$ (labeled vector 6), defined by Eq. (10.53). This results in an output vector from the input gate, labeled vector 9. We then merge the output vector from the forget gate, vector 8, and the output vector from the input gate, vector 9. This gives an output vector that has survived through the operations within the forget and input gates. In the figure, we represent this output vector as a long-term state vector $\mathbf{c}(t)$, labeled vector 10 in the figure. We present this operation as:

$$\mathbf{c}(t) = \mathbf{f}(t) \otimes \mathbf{c}(t-1) + \mathbf{i}(t) \otimes \mathbf{g}(t) \qquad (10.66)$$

As we will send shortly the same information for the long-term state vector $\mathbf{c}(t)$ to an output gate for processing, we label that vector as vector 11 in the figure.

Vector 11 then goes through a hyperbolic tangent activation function with an output vector with scalar components bounded between -1 and 1 (see Figure 10.24). This results in an output vector labeled vector 12. This vector then multiplies with output vector 5 from the output gate element-wise to give an output vector, labeled vector 13, which is identical to the output vector $\mathbf{y}(t)$ at time step t, and to the short-term state vector, $\mathbf{h}(t)$. The latter will proceed to the next time step (t+1) as an input vector $\mathbf{h}(t)$ to work with the input vector $\mathbf{x}(t+1)$. Both will go through the same steps illustrated above involving the forget gate to select memory, the input gate to update memory and the output gate to select output.

### 10.4.2c Gated Recurrent Unit (GRU)

This section presents the concept of the gated recurrent unit (GRU), which is a simplified version of the LSTM memory cell. Burkov [7, p. 74] emphasizes that *GRU is the most effective RNN used in practice.* In

**Workshop 10.3**, Section 10.8, we show that GRU performs more accurately than other deep RNNs in predicting time-dependent melt index during grade transition in an industrial HDPE process.

Cho et al. [125] in their 2014 paper introducing the encoder-decoder network that we will discuss later, actually presented the concept of gated recurrent unit (GRU) for the first time.

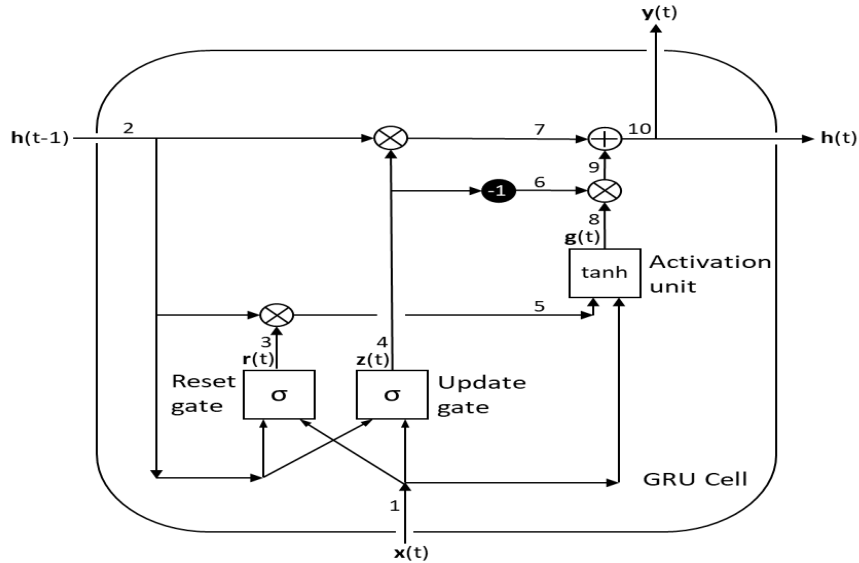

Figure 10.29 A GRU memory cell [8].

Referring to Figure 10.29, we see the input vector **x**(t) at time t (labeled vector 1 in the figure) and the output vector from the previous time step, **h**(t-1) (labeled vector 2) both enter a filtering unit, called *the reset gate*, with a sigmoid activation function, giving an output vector **r**(t) (labeled vector 3). Following Eq. (10.62), we represent **r**(t) as [8, p.520]:

$$\mathbf{r}(t) = \sigma[\mathbf{W}_{xr}^T \mathbf{x}(t) + \mathbf{W}_{hr}^T \mathbf{h}(t-1) + \mathbf{b}_r] \qquad (10.67)$$

where σ is the sigmoid activation function with an output vector with scalar components bounded between 0 and 1 (see Figure 10.24). The reset gate decides if the output vector from the previous time step, **h**(t-1), will be ignored. Specifically, when the reset gate is close to 0, **r**(t) depends primarily on **x**(t) only [114]. The reader can interpret the relevant weight factor matrices and the bias term in Eq. (10.67) and subsequent Eqs. (10.56) - (10.69) in the same way as those in Eqs. (10.62) to (10.64).

The same input vector **x**(t) at time t and the output vector from the previous time step, **h**(t-1) also enter another filtering unit, called *the update gate*, with a sigmoid activation function, giving an output vector **z**(t) (labeled vector 4). Following Eq. (10.67), we represent **z**(t) as [8, p.520]:

$$\mathbf{z}(t) = \sigma[\mathbf{W}_{xz}^T \mathbf{x}(t) + \mathbf{W}_{hz}^T \mathbf{h}(t-1) + \mathbf{b}_z] \qquad (10.68)$$

The update gate controls how much information from the output vector from the previous time step, **h**(t-1), will carry over the output vector from the current time step, producing an updated output vector **h**(t). This acts similarly to the memory cell in the LSTM network, and helps the RNN to remember long-term information.

In Figure 10.28, the activation unit has two inputs: the first is the input vector **c**(t) (labeled vector 1), and the second one is vector 5, which results from the element-wise multiplication of the output vector from the reset gate **r**(t) (labeled vector 3) and the output vector from the previous time step, **h**(t-1), that is, **r**(t)⊗**h**(t-1). Vectors 1 and 5 enter the activation unit with a hyperbolic tangent activation function with an output vector with scalar components bounded between -1 and 1 (see Figure 10.24). Following Eq. (10.67), we represent the output vector from this activation unit as **g**(t) (labeled as vector 8) [8, p.520]:

$$\mathbf{g}(t) = \tanh[(\mathbf{x}(t) + \mathbf{W}_{hg}^{T} \ [\mathbf{r}(t) \otimes \mathbf{h}(t\text{-}1)] + \mathbf{b}_g) \tag{10.69}$$

Referring to Figure 10.28, we look at the element-wise multiplication between g(t) (labeled as vector 8) and vector 6, which results from applying the operator "1-" to vector 4, that is, the output vector from the update gate, **z**(t). The latter simply means [1-**z**(t)]. Therefore, we represent vector 9 resulting from this element-wise multiplication as vector [1-**z**(t)]⊗**g**(t).

Next, we represent the element-wise multiplication of vector 4, that is, output vector z(t) exiting the update gate, with the output vector from the previous time step, **h**(t-1), labeled vector 2, as **z**(t)⊗**h**(t-1). The results in vector 7 in the figure.

Finally, the element-wise addition of vector 7 and vector 9 gives vector 10, representing the output vector from the GRU at time t, **y**(t), which is also sent to the next GRU cell to work with the input vector from the next time step, **x**(t+1), as inputs to the following GRU memory cell. In the figure, we represent the same vector as **h**(t), also labeled as vector 10. We represent **y**(t) or **h**(t) as:

$$\mathbf{y}(t) = \ \mathbf{h}(t) = \mathbf{z}(t) \otimes \mathbf{h}(t\text{-}1) + [1\text{-}\mathbf{z}(t)] \otimes \mathbf{g}(t) \tag{10.70}$$

which is similar to Eq. (10.66) in the LSTM memory cell.

We pause to recommend that the reader takes time to study an instructive, step-by-step illustrated guide to LSTM and GRU by Phi [128]. LSTM and GRU are one of the main reasons behind the success of RNNs. They can handle much longer sequences, time-series data, or dynamic process data with approximately 100 time steps better than simple RNNs [8, p. 520].

**10.4.2d Bidirectional RNNs**

Schuster and Paliwal [118] introduce the concept of bidirectional RNN. Figure 10.30 illustrates a bidirectional RNN. We see two hidden layers in the network, one forward layer and one backward layer. Each layer may consist of LSTMs or GRUs. The network essentially consists of two separate subnetworks, and they serve as separate "experts" for the specific problem on which the networks are trained. One way of merging the opinions of forward experts and backward experts is to assume the opinions to be independent, which leads to arithmetic averaging for regression and to geometric averaging (or arithmetic averaging in the log domain) for classification [118].

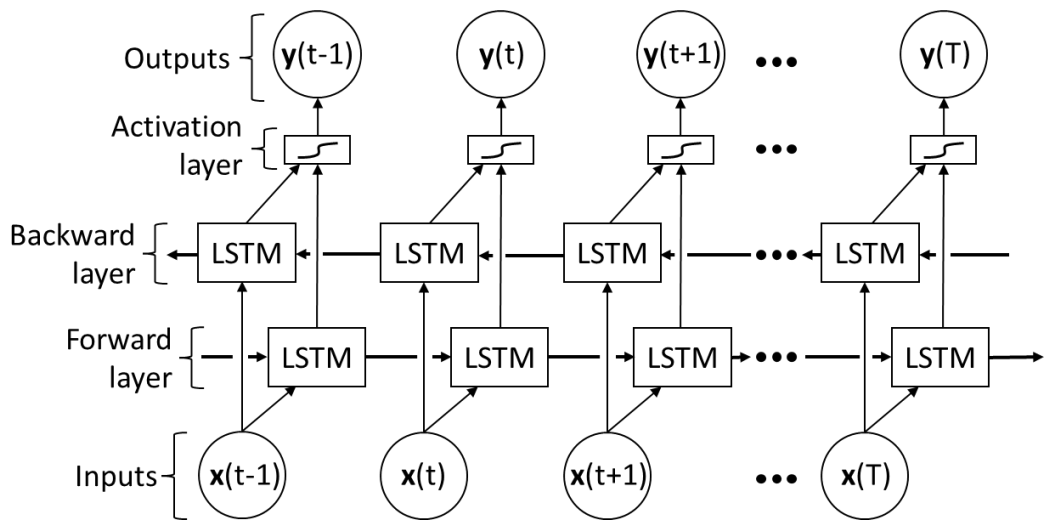

Figure 10.30 An illustration of a bidirectional RNN.

Zhang, et. al. [119] apply bidirectional LSTM and GRU RNNs to a problem in data-driven fault detection and diagnosis of the well-known Tennessee Eastman process [4, pp. 269-291; 120, 121]. They give the details of the application and demonstrate that bidirectional LSTM and GRU RNNs exhibit a dramatically improved performance over other methods for chemical process fault detection and diagnosis. With an additional backward RNN, bidirectional RNNs facilitate the detections of variable deviations over all time points, particularly when a fault has just happened. However, as with all data-based methods, the directional RNNs require adequate historical process data.

We refer the reader to the details of bidirectional RNNs [118] and to many available online tutorials on the training of bidirectional RNNs. In **Workshop 10.3**, Section 10.8, we compare the performance of two types of deep RNNs (LSTM and GRU) in predicting time-dependent melt index during grade transition in an industrial HDPE process.

### 10.4.3 Convolutional Neural Networks (CNNs)

Since its development by LeCun et al. in 1990 [134], convolutional neural networks (CNNs) have become a key tool for identifying satellite images, processing medical images, forecasting time series, detecting data anomalies, and recognizing characters such as ZIP codes and digits. In chemical industries, we see growing applications of CNNs to polymer property predictions from chemical structure [135], classification of LC-MS spectral peaks [136], prediction of thermophysical properties of chemical components from their surface charge distributions (called sigma profiles) [137, 138], computer-aided molecular design and screening methodology for fragrance molecules [140], performance prediction of protein-exchange membrane fuel cells [141], image analytics for classifying industrial catalyst pallets [102], and identifying color changes in thermal food processing [142], among others. In Section 10.8, we present Workshop 10.4 for polymer property prediction from molecular structure using convolutional neural networks [135].

We follow [1, pp.760-764; 97, 139] to explain a few key concepts of CNNs. Our goal is to give sufficient background to enable the reader to understand the cited references for additional details of applications. First, we know that a camera scans photographs to encode the images into pixels. To represent an image, we cannot use a simple vector of input pixel values, as the adjacency of pixels that correlate with each other really matters. In particular, we must consider an important property of images, called *local spatial invariance*. Specifically, we can say roughly that anything that is detectable in one small region of the image (e.g., an eye) would look the same if it appeared in another small local region of the image. We can achieve this local spatial invariance by constraining weight factors connecting a local region to a unit in the hidden layer to be the same for each hidden unit. Therefore, a convolutional neural network (CNN) contains spatially invariant, local connections, at least in the first few layers, and has patterns of weight factors that are replicated across units in each layer. Applying this local spatial invariance characteristic significantly reduces the number of parameters in a deep neural network with many units without losing too much in the quality of the model [7].

Figure 10.31, adopted from [1, p. 761], illustrates the concepts of *kernel and stride* that are important in characterizing a convolution layer. For simplicity, we limit this discussion to a one-dimensional convolution operation. The reader may refer to [8, pp. 446-460] for multi-dimensional examples. First, a *kernel (*also called *a filter)* is a pattern of weight factors that is replicated across multiple local regions. *Convolution* is the process of applying the kernel to the pixels of the image, or to spatially organized units in a subsequent layer.

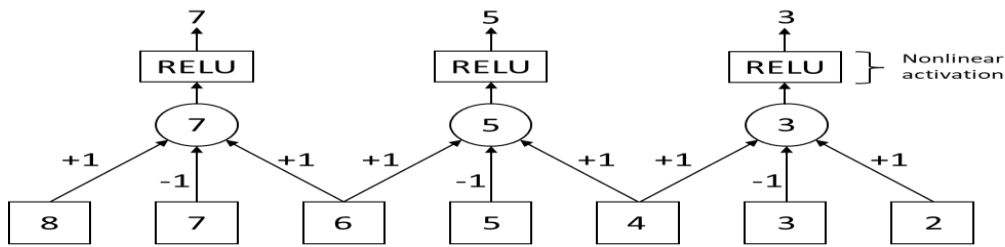

Figure 10.31  An illustration of a one-dimensional convolution operation with a kernel of size $i$ = 3 and a stride $s$=2.

We see in the figure that the kernel vector $\boldsymbol{k}$ = [+1 -1 +1], and this vector is replicated three times. We say that this kernel is of size $i$= 3. In this example, the pixel on which the kernel $\boldsymbol{k}$ = [+1 -1 +1] is entered, is separated by a distance of two pixels; we say that the kernel is applied with *a stride* of $s$ =2. We can understand this more easily by using a single matrix operation as follows:

$$\begin{bmatrix} +1 & -1 & +1 & 0 & 0 & 0 & 0 \\ 0 & 0 & +1 & -1 & +1 & 0 & 0 \\ 0 & 0 & 0 & 0 & +1 & -1 & +1 \end{bmatrix} \begin{pmatrix} 8 \\ 7 \\ 6 \\ 5 \\ 4 \\ 3 \\ 2 \end{pmatrix} = \begin{pmatrix} 7 \\ 5 \\ 3 \end{pmatrix}$$

In this matrix, the 3-element kernel (*i*=3) appears in each row, shifted two positions to the right according to the stride (*s*=2) relative to the previous row.

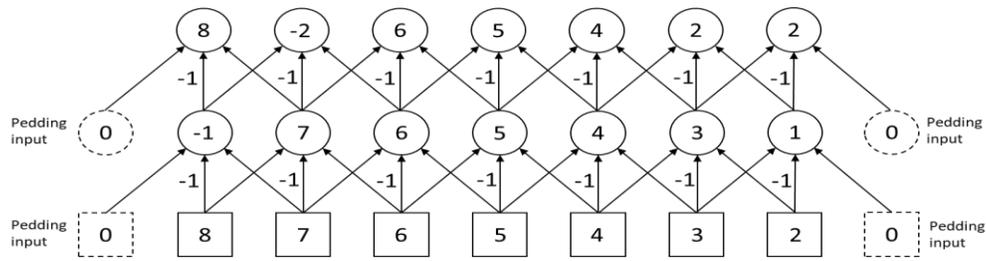

Figure 10.32   An illustration of padding inputs of zero on the left and right ends of hidden layers.
In the plot, all slopped arrows have kernel value of +1.

Figure 10.32 illustrates the concept of *padding* [1, p.762]. In this example, we have a kernel size *i*=3, and a stride *s*=1. Padding inputs of zero are added to the left and right ends of the two hidden layers to keep the hidden layers the same size as the input.

We could also look at the concept of kernel and stride from the view of a receptive field originated in models of the visual cortex in neuroscience. Specifically, *the receptive field* of a neuron is the portion of the sensory inputs that can affect the activation of the neuron. In a convolutional network, *the receptive field of a unit in the first hidden layer is just the size of the kernel, that is, i pixels.* It is possible to connect a large input layer to a small layer by spacing out the receptive fields or kernels. We call the shift from one receptive field to the next *the stride*, such as *s*=2 in the current example.

Figure 10.31 shows the output coming out of the rectified linear unit (ReLU), [7 5 3]. We call this *a rectified feature map*, which then feeds into a new layer, called *the pooling layer*. The goal of pooling is to reduce the computational load, the memory usage, and the number of parameters (thus limiting the risk of overfitting). Basically, a pooling layer in a convolutional network summarizes a set of adjacent units from the preceding layer by a single value. This layer works like a convolutional layer with the same kernel *l* and stride *s*, but there is no activation function to process the output from the layer. There are two types of pooling operations: (1) average pooling computes the average value of its *l* inputs. This is equivalent to a convolution with a uniform kernel vector, $k$ = [1/*l*, 1/*l*,....1/*l*]. The effect of this operation is to coarsen the resolution of the image (i.e., "to downsample" or "to shrink") by a factor of *l*. (2) Max-pooling computes the maximum value of its *l* inputs. For the outputs from the ReLU in Figure 10.31, we see that average pooling gives an output of (7 + +5 +3)/3 = 5; and max-pooing gives an output if Max [7 5 3] = 7.

Following [131], we give another example of pooling and flattening in Figure 10.33.

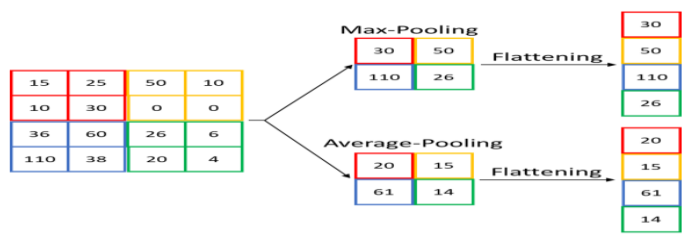

Figure 10.33   An example of max-pooling and average-pooling, and flattening.

Lastly, *a fully connected layer* (also called *a dense layer*) forms when the flattened matrix from a pooling layer is fed as an input, which classifies and identifies the image.

Applying all the concepts that we have explained thus far, we show in Figure 10.34 a block diagram of the CNN that Abranches, et al. [137] used in their work on predicting physiochemical properties from sigma profiles [138] that we discussed previously in Section 10.1.2.b.

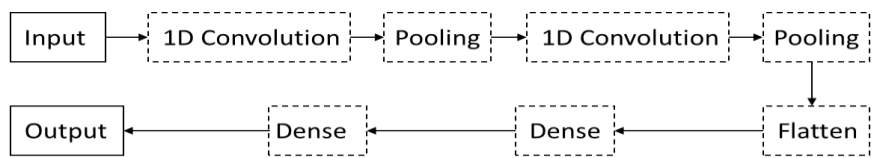

Figure 10.34.  A block diagram of the CNN for predicting physiochemical properties from sigma profiles.

We refer the reader to Section 10.9, **Workshop 10.4**, prediction of polymer property based on molecular structure using CNN, together with the transformer network introduced in the next section.

**10.4.4 Attention Is All You Need: Transformer Model**

In a 2017 groundbreaking paper, entitled "Attention Is All You Need", Ashiwani Vaswani and his colleagues [143] proposed *the transformer architecture*, which is a neural network that learns context and thus meaning by tracking relationships to *sequential data* like the words in a sentence (similar to *time-dependent process variable values* in a dynamic chemical process). A transformer model applies an evolving set of mathematical technique, called *attention mechanisms*, to detect subtle ways that distant data elements in a series can influence and depend on each other [144].

According to Merritt [144], transformer models are in many cases replacing recurrent neural networks (RNNs) and convolutional neural networks (CNNs), which were the most popular types of deep neural networks (DNNs) for sequence-to-sequence ("seq2seq") learning just five years ago. Transformer models overcome a significant deficiency of the sequential nature of both RNNs and CNNs that *prevents parallel processing*. We recommend some online tutorial articles on transformer models [145-147]. Selected examples of the rapidly growing applications of the transformer models in chemical process industries include molecular optimization for new drug discovery [148-149], computational chemistry [150], COSMO-SAC sigma profile modeling for high-throughput solvent screening [151], and dynamic soft sensor modeling for polypropylene melt index prediction [152].

We introduce the transformer model by referring to an application to *new drug discovery* for a protein target reported by Grechishnikova [149]. Previously, most methods for protein-specific new drug generation require prior knowledge of protein binders, their physiochemical characteristics, or the three-dimensional structure of the protein. By adopting the transformer model, the proposed method generates novel molecules with predicted ability to bind a target protein *by relying on amino acid sequence only*. Grechishnikova considers the target-specific new drug design as a "translational" problem between the amino acid "language" and the Simplified Molecular Input Line Entry System (SMILES) representation of the molecule [122]. In Section 10.1.2b, we briefly introduced SIMLES and its extensions [122 to 124] as an effective molecular representation as inputs to deep neural networks, and SMILES uses a string of characters to represent atoms, bonds, branches, cyclic structures, disconnected structures, and aromaticity with coding rules. In Section 10.9, we apply the SIMLES representation [122] to predict polymer property using convolutional neural networks [135].

We follow [7, pp. 541-547; 8, pp. 541-562; 143-147] to introduce the basic concepts of the transformer model. Figure 10.35 shows the original architecture of the transformer model [143], and we discuss the key concepts below. In the figure, we see that both the left and right parts of the model use a stack of N (typically 6; indicated by Nx in the figure) identical encoder layers and a stack of N identical decoder layers. As the input sequences to the encoder and decoder are processed separately and in parallel, the transformer model reduces the number of computations required in comparison to a model which processes both sequences together.

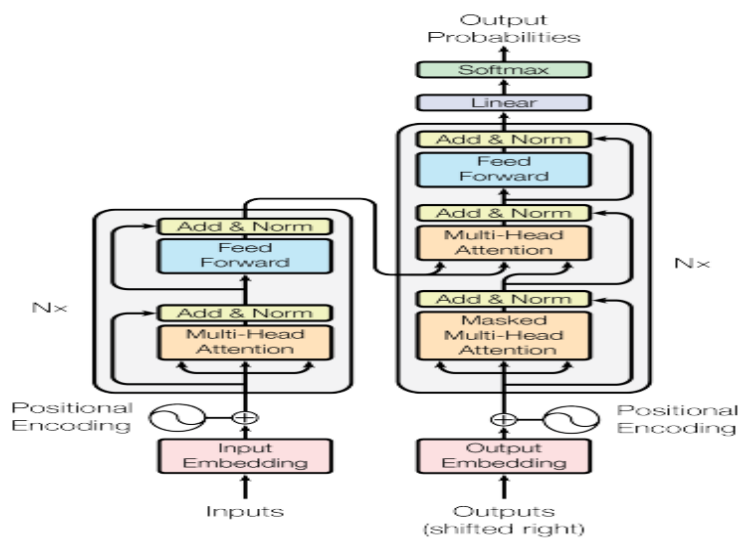

Figure 10.35 The architecture of the transformer model [143]

### 10.4.4a   Encoder-Decoder Stacks

We follow [7, pp. 87-92] to introduce *the encoder-decoder architecture* adopted by the transformer models.  The left part of Figure 10.35 represents *an encoder*, and the right part is *a decoder*. Briefly, the encoder reads the input and generates some sort of state (similar to the state in recurrent current networks that we discussed in Figure 10.28) that can be seen as a numerical representation of the *meaning* of the input that the machine can work with. In ML, we refer to the meaning of some entity, whether it be an image, a text or a video, as *a vector or a matrix that contains real numbers*. We call this vector or matrix the embedding of the input, as we see the term *input embedding* on the left part of Figure 10.35.

The decoder takes an embedded input from the encoder and generates a sequence of outputs. Because the transformer encoder has no recurrence nor convolution, in order for the transformer model to make use of the order of the sequence, we must add some information about the positions into the input embeddings. This is done using *positional encoding*, as illustrated in the bottom of Figure 10.35. Let *pos* be the position of the word in a sequence, $d_{model}$ be the length of the encoding vector (same as the embedding vector; typically, $d_{model}$ =512), and *i* be the index value into this vector, Vaswani, et al. [143] proposed to use sine and cosine functions of different frequencies to characterize the positional embedding (PE):

$$PE(pos,2i) = sin(pos/ 10000^{2i/d_{model}}) \qquad (10.71)$$

$$PE(pos,2i+1) = cos(pos/ 10000^{2i/d_{model}}) \qquad (10.72)$$

Here, sine values correspond to all even indexes, and cosine values represent all odd indexes.

Conceptually, the decoder takes a start-of-sequence input feature vector **x**(0), produces the first output **y**(1), updates its state by combining the embedded input **x**(0) with the output **y**(1) to produce its next input **x**(1), which is similar to the architecture illustrated previously with RNNs in Figure 10.24. The resulting vector or matrix embedding the output, as we see the term *output embedding* on the right part of Figure 10.35.

For the new drug discovery problem [149], the encoder maps a protein acid sequence ($x_1$, $x_2$, ...$x_n$) to a sequence of continuous representation **z**= ($z_1$, $z_2$, .... $z_n$). Given **z**, the decoder then generates an output sequence ($y_1$, $y_2$, ... $y_m$) one element at a time. In this case, the output sequence is a SMILES string [122].

Referring to the representation of a single encoder layer (as a part of a stack of 6 identical encoder layers) displayed in the left part of Figure 10.35, we see two sublayers, namely: (1) *a multi-head attention mechanism*; and (2) *a fully-connected feedforward network*, both of which are explained shortly below. The transformer model uses *a residual connection* [154] around each of the two sublayers, meaning that the encoder provides the data an alternative path to reach latter parts of the network by skipping a sublayer. Additionally, the encoder employs *layer normalization* [155]. Unlike batch normalization discussed in Section 10.4.1e, layer normalization directly estimates the normalization statistics from the summed inputs to the neurons within a hidden layer so the normalization does not introduce any new dependencies between training cases.

Turning now to the right part of Figure 10.35, we see that within a single decoder layer (as a part of a stack of 6 identical decoder layers), the decoder has actually *a "masked" multi-head attention* that receives the output from the encoder stack, in addition to *a multi-head attention* that receives inputs from both the encoder and the decoder, and *a feedforward network*. The decoder stack also applies *residual connections* around each of the sublayers, followed by *layer normalization*. We have previously explained the concept of peddling in Figure 10.32. Here, masking within the decoder layer through *the "masked" multi-head attention* attempts to zero attention output where there is paddling in the input sequences to ensure that paddling does not contribute to the self-attention. We discuss the attention mechanisms below.

### 10.4.4b  Self-Attention and Attention Mechanisms within the Encoder-Decoder Stacks

As we see in the bottom of both left and right parts of Figure 10.35, transformer models use *positional encoding* to tag data elements coming in and out of the network. Attention units follow these tags, calculating a kind of algebraic map of how each element relates to each other. Attention queries are typically executed in parallel by calculating a matrix of equations in what is called *multi-headed attention*, a term we see within Figure 10.35 [144]. This is actually *a self-attention mechanism*.

What is self-attention and how does it find meaning? Self-attention, sometimes called intra-attention, is an attention mechanism that relates different portions of a single sentence in order to compute a representation of the sequence [143]. Consider, for example, the sentence: "David pours water from the pitcher to the cup until it is full". We know "it" refers to the cup. Being able to identify "it" as the cup is an example of the self-attention mechanism in action [144]. As we described above, the encoder reads the input and generates some sort of state that can be seen as *a numerical representation of the meaning* of the input that the machine can work with. This meaning represents a relationship between things.

The attention layer takes its input in the form of three parameters, known as *Query (Q), Key (K) and Value (V).* In the original formulation of the transformer model involving *sequential data* like the words in a sentence, we can interpret the Query parameter as the word *for which* we are quantifying the attention, and interpret the Key and Value parameters as the words *to which* we are paying attention, that is, how relevant is that word to the Query word [145]. For sequential data, we typically represent Q, K, and V values in a sequential order as rows or columns of a matrix.

In the transformer model, the attention module repeats its computations multiple times in parallel. Each of these computation paths is called *an attention head*. The attention module splits its parameters in multiple ways, and passes each split independently though a separate head. All of these similar calculations are then combined through appropriate probability-based weighting factors to produce *an attention score,* for which we call *multi-head attention* [145].

### 10.4.4c  Softmax Function and Probability-Based Weighting Factors

The Softmax function is a function that turns a vector of J real values into a vector of J real values that sum to 1. The input values can be positive, zero, negative or greater than one, and the Softmax function transform them into values between 0 and 1. Specifically, the Softmax function is:

$$\sigma(\mathbf{z})_i := \frac{e^{z_i}}{\sum_{j=1}^{J} e^{z_i}} \qquad\qquad (10.73)$$

Where do we apply the Softmax function to the transformer model? Let us look at the three steps of an attention-based learning system applying to a sequential data like translating as an English sentence into Spanish.

1) If we are processing an input sequence of words, we first feed the input sequence into an encoder, which will output a vector of every element in the sequence.

2)A list of these vectors, together with the decoder's previous hidden states, will be exploited by the attention mechanism to dynamically highlight which of the input information will be used to generate the output.

3) At each time step, the attention mechanism takes the previous hidden state of the decoder and the list of encoded vectors and use them to generate unnormalized score values that indicate how well the elements of the input sequence align with the current output. As the generated score values should make relative sense in terms of their importance, we normalize the scores by passing them through a Softmax function to generate weights, which will have values between 0 and 1, and the values also add to 1. We can interpret the resulting weights as probabilities. Finally, the encoded vectors are scaled by the computed weights to generate a vector, called *a context vector*, which is then fed to the decoder to generate a translated output.

### 10.4.4d  Attention Mechanisms in Encoder-Decoder Stacks

Within Figure 10.35, we see three attention mechanisms in actions. These are [145]:

1) Self-attention in the encoder, called *multi-head attention*: the input sequence to the encoder pays attention to itself.

2) Self-attention in the decoder, called *masked multi-head attention*: the input sequence to the decoder pays attention to itself.

3) Encoder-decoder attention in the decoder, called *multi-head attention*: the target sequence pays attention to the input sequences from the encoder and from the decoder.

Vaswani, et al. [143] proposed the multi-head attention mechanism on the left part of Figure 10.36, which includes the scaled dot-product attention mechanism that is displayed in more detail on the right part of Figure 10.36.

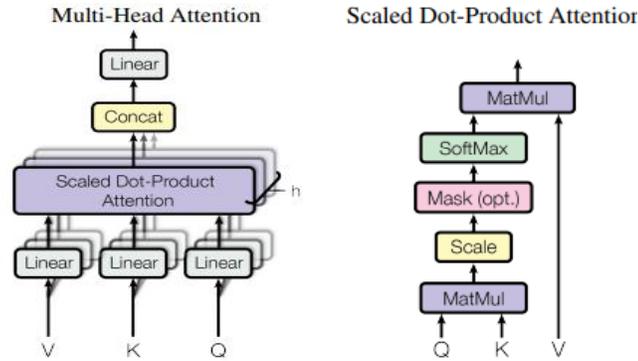

Figure 10.36 Multi-head attention and scaled dot-product attention [143]

We recommend the reader to pause temporarily and go through parts 3 and 4 of Doshi [145] that use simple graphical demonstration of the vector-matrix operations to open up the black boxes involved in the two attention mechanisms of Figure 10.36, and in the brief description of the two attention mechanisms in Vaswani et al. [143].

According to Vaswani et al. [143], the input to the scaled dot-product attention mechanism consists of query and key vectors of dimension $d_k$, and the value vector of dimension $d_v$. The mechanism computes the dot products of queries with all keys and scale the dot products down by dividing each by $\sqrt{d_k}$, and then applies a softmax function to obtain the weights on the values. Here, dividing each dot product by $\sqrt{d_k}$ is to allow for more stable computations, as multiplying large values can have exploding effects.

In practice, the attention mechanism computes the attention score on a set of queries simultaneously, packed together into a matrix Q. The keys and values are also packed together into matrices K and V. We then calculate the attention score matrix by the following softmax function of the matrix operation:

$$\text{Attention(Q, K, V)} = \text{softmax}(\frac{QK^T}{\sqrt{d_k}})*V \qquad (10.74)$$

The reader may refer to Doshi [145] for an explicit graphical vector-matrix demonstration of this calculation.

Referring to the left part of Figure 10.36, Vaswani et al. [143] proposed the following steps for the multi-head attention mechanism. Instead of performing a single attention function with a single-dimension vectors of query, key and value, the attention mechanism linearly projects the queries, keys and values $h$ (e.g., 8) times with different, learned linear projections to $d_k$, $d_k$, and $d_v$ dimensions, respectively (e.g., $d_k$ = $d_v$ = $d_{model}/h$ = 512/8 = 64). On each of these projected versions of queries, keys, and values, the attention mechanism then computes the attention function in parallel, resulting in $d_v$-dimensional values. These values are then concatenated and once again linearly projected, resulting in the final values, as shown on the left part of Figure 10.36.

For multi-head attention mechanism produces *h* different representations of Q, K and V values and computes an attention function for each representation:

$$Head_i = Attention\ (QW_i^Q,\ KW_i^K,\ VW_i^V) \quad i = 1\ to\ h \tag{10.75}$$

The outputs are concatenated and projected one more time, giving the final values:

$$Multi\text{-}Head\ (Q,\ K,\ V) = (head_1,\ head_2,\ ....\ head_h)*W^o \tag{10.76}$$

where the dimensions of matrices are: $W_i^Q$ and $W_i^K$ are both $d_{model}\ x\ d_k$; $W_i^V$ is $d_{model}\ x\ d_v$; $W^o$ is $hd_v$ x $d_{model}$ [149].

Again, we refer the reader to Doshi [145] for detailed graphical demonstration of all the calculation steps involved.

### 10.4.4e   Position-Wise Feedforward Networks

We see in both left and right parts of Figure 10.35, each of the layers in the encoder and decoder contains a fully connected feedforward network, which is applied to each position separately and identically. This consists of two linear transformations with a ReLU activation function in between. Following see Table 10.6 and Figure 10.24, we write the ReLU activation function as f(x) = 0, x<0, and f(x) = x, x≥0, or f(x) = max (0, x). We can then represent the output from the feedforward network as:

$$FNN(x) = max\ (0,\ xW_1 + b_1)\ W_2 + b_2 \tag{10.77}$$

$W_1$ is the vector of weights of the first feedforward network of all the neurons and $b_1$ is its bias vector. The output from the first feedforward network then goes through a ReLU transform, and hence the max function. The transformed feature is again multiplied by the weight vector $W_2$ of the second layer and its bias vector $b_2$ to give the output.

### 10.5 General Guidelines for Choosing Appropriate ML Algorithms
### 10.5.1 Factors to Consider in Selecting Appropriate ML Algorithms

We summarize the key points in [33 to 36], among many other online resources, the key factors to consider in choose appropriate ML algorithms.

1. *Interpretability*: Interpretability is the ability to determine the cause and effect from a ML model. If a model can take the inputs, and routinely get the same outputs, the model is interpretable. If understanding the reason behind the model results is a requirement for our problem, we need to choose an interpretable method (e.g. a linear regression model), instead of a black-box model (e.g., a deep neural network).

2. *Accuracy*: For regression or prediction problems, review the performance evaluation metrics, such as mean squared error (MSE), root mean squared error (RMSE), and coefficient of determination ($R^2$), defined in Section 10.2.1b. Understand the bias-variance tradeoff illustrated in Figure 10.2. For classification problems, review the performance metrics, such as precision, recall and F1 score, defined in Section 10.2.2e.

Figure 10.37, adopted from [168], gives an empirical plot of accuracy versus interpretability of ML algorithms. Highly accurate ML models typically have nonlinear or non-smooth relationships and require long training time, while highly interpretable ML models typically involve linear, smooth, and well-defined relationships, and are easy to train. Decision trees are unique ML models that can have both good accuracy and high interpretability. We also note that for tabular data, some ensemble methods (e.g., XGBoost) could give Comparable accuracy than deep neural networks.

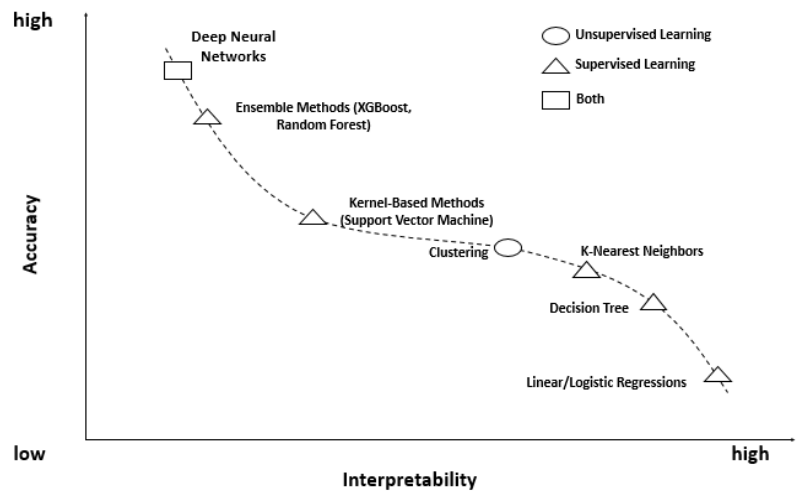

Figure 10.37 An empirical plot of accuracy versus interpretability of ML models: "o" - unsupervised learning; "Δ" – supervised learning; "□" – both.

3. *Training time*: This is the time taken by a ML algorithm to learn and create a model. It depends on whether we wish to update the resulting model continuously, as in the case of stock price prediction. Enhanced learning methods, such as ensemble methods and deep neural networks typically take a much longer training time, while achieving the best accuracy. By contrast, linear regression, and its variants have faster training time, but llower accuracy. Figure 10.38 gives an empirical plot of training time versus accuracy of ML models.

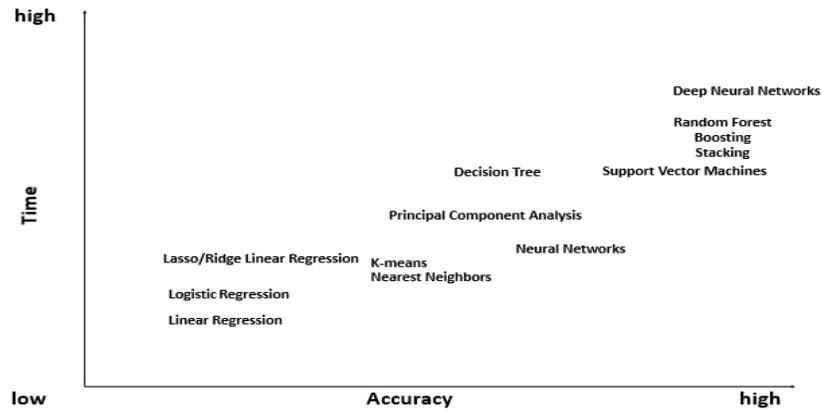

Figure 10.38 An empirical plot of training time versus accuracy of ML models

Table 10.8 summarizes additional key considerations, including:
4. *Dataset*: Size, type of features, number of features and examples, linearity characteristics, preprocessing (mean-centering and scaling of features);
5. *Parameters;*
6. *Constraints*;
7. *Accuracy versus interpretability*;
8. *Training time versus accuracy*.

Table 10.8  Factors to consider in choosing appropriate ML algorithms

| Factors | Characteristics | Suggested Algorithms |
|---|---|---|
| Problem input type | 1. Input data -labeled | Supervised learning |
| | 2. Input data-unlabeled | Unsupervised learning |
| | 3. Inpt data- both labed and unlabeled | Semi-supervised learning |
| | 4. To optimze an objective function by interacting with the environment | Reinforcement learning |
| Problem output type | 1. Numerical vs categorical | Regression/prediction vs classification |
| | 2. A set of input groups | clustering |
| | 3. Outlier or anomaly | Outlier or anomaly detection |
| | 4. Reduced dimension of features | Dimensionality reduction |
| Dataset: size | 1. Small | Favor high bias/low variance model |
| | 2. Large | Favor low bias/high variance model |
| Dataset: type of features | 1. Numerical | Regression models |
| | 2. Categorical | Classification models |
| Dataset: number of features and examples | 1. Small/modest | Support vector machines |
| | 2. Large | Neural networks; ensemble methods |
| Dataset: linearity characteristic | 1. Linear | Linear regression, logistic regression, SVM |
| | 2. Nonlinear | Ensembled models; Deep neural netwoks |
| Dataset: preprocessing (Mean-centering and scaling of features) | 1. Prefer preprocessing | Most models |
| | 2. Preprocessing not required | Decision trees |
| Parameters | 1. With parameters | Parametric models |
| | (a) Set before training | (a) Hyperparameters |
| | (b) Optimized during training | (b) Parameters |
| | 2. Without parameters | Nonparametric models (e.g. decision trees) |
| Constraints | 1. Must update model continuouly (e.g., stock prediction) | Fast training time required |
| | 2. Accuracy is more important than training time (e.g., detecting cancer cells) | Accurate model required |
| Accuracy vs interpretability | See Figure 10.37 | |
| Training time vs accuracy | See Figure 10.38 | |

**10.5.2 A Summary of Selected Machine Algorithms**

To help the readers choose appropriate ML algorithms, we adopt the summary of machine learning algorithms by DataCamp [168] in Tables 10.9 to 10.11 to summarize the linear regression, tree-based and clustering algorithms according to the algorithm name, description, advantages and disadvantages. We also compare deep neural networks (DNNs), including the multilayer perception (MLP), recurrent neural network (RNNs), convolutional neural network (CNN) and transformer network in Table 10.12.

Table 10.9   A summary of selected linear regression algorithms [167]

| Algorithm | Description | Advantages | Disadvantages |
|---|---|---|---|
| Linear regression (Section 10.2.1a) | A simple algorithm that models a linear relationship between inputs and a continuous numerical output variable | Explainable method; Interpretable results by its output coefficient; Fast to train | Assumes linearity between inputs and output; Sensitive to outliers; Can underfit with small, high-dimensional data |
| Logistic regression (Section 10.2.1c) | A simple algorithm that models a linear relationship between inputs and a categorical output (1 or 0) | Interpretable and explainable; Less prone to overfitting with regularization; Applicable for multi-class predictions | Assumes linearity between inputs and outputs; Can overfit with small, high-dimensional data |
| Ridge regression (Section 10.2.1d) | A regression method that penalizes features that have low predictive outcomes by shrinking their coefficients closer to zero. Can be used for classification or regression | Explainable and interpretable; Less prone to overfitting; Best suited where data suffer from collinearity; | All the predictors are kept in the final model; Doesn't perform feature selection |
| Lasso regression (Section 10.2.1.d) | Same as Ridge regression | Less prone to overfitting; Can handle high-dimensional data; No need for feature selection | Can lead to poor interpretability as it can keep highly correlated variables |

Table 10.10   A summary of selected tree-based algorithms [167]

| Algorithm | Description | Advantages | Disadvantages |
|---|---|---|---|
| Decision tree | Decision Tree models make decision rules on the features to produce predictions. Can be used for classification or regression | Explainable and interpretable; <br><br> Can handle missing values | Prone to overfitting; <br><br> Sensitive to outliers |
| Random forest | An ensemble learning method that combines the output of multiple decision trees | Reduces overfitting; <br><br> Higher accuracy compared to other models | Training complexity can be high; <br><br> Not very interpretable |
| Gradient boosting regression | Gradient boosting regression employs boosting to make predictive models from an ensemble of weak predictive learners | Better accuracy compared to other regression models; <br> It can handle multicollinearity <br><br> It can handle non-linear relationships | Sensitive to outliers and can therefore cause overfitting; <br> Computationally expensive and has high complexity |
| Extreme gradient boosting (XGBoost) | Gradient boosting algorithm that is efficient and flexible. Can be used for both classification and regression tasks | Provides accurate results; <br><br> Captures non-linear relationships | Hyperparameter tuning can be complex; <br> Does not perform well on sparse datasets |

Table 10.11  A summary of selected clustering algorithms

| Algorithm | Description | Advantages | Disadvantages |
|---|---|---|---|
| K-means clustering | K-Means is the most widely used clustering approach. It determines K clusters based on Euclidean distances | Scales to large datasets; Simple to implement and interpret; results in tight clusters | Requires the expected number of clusters from the beginning; Has troubles with varying cluster sizes and densities |
| Hierarchical clustering | A "bottom-up" approach where each data point is treated as its own cluster—and then the closest two clusters are merged together iteratively | There is no need to specify the numberof clusters; The resulting dendrogram is informative | Doesn't always result in the best clustering; Not suitable for large datasets due to high complexity |
| Density-based spatial clustering of applications with noise (DBSCAN) | A method to identify cluster based on the density of region in the data | Identify cluster of arbitrary shapes; No need for initial number of clusters; Can identify outliers | Challenging to specify hyperparameters-the radius of the cluster around each data point and the limiting number of clusters |
| Gaussian Mixture model | A probabilistic model for modeling normally distributed clusters within a dataset | Computes a probability for an observation belonging to a cluster; Can identify overlapping clusters and outliers More accurate results compared to K-means | Requires complex tuning; Requires setting the number of expected mixture components or clusters |

Table 10.12 A summary of selected deep neural networks

| Deep Neural Network Type | Description | Input data | Advantages | Disadvantages |
|---|---|---|---|---|
| Multilayer Perception (MLP) | Section 10.4.1a | Tabular data | Capable of learning any nonlinear function | (1) While solving an image classification problem using a MLP, must convert a 2D image into a 1D vector before model losing the spatial features of an image. (2) Vanishing and exploding gradients (Section 10.4.1d) |
| Recurrent neural networks (RNNs): LSTM (long short-term memory) RNN, gated recurrent unit (GRU), bi-directional RNN | Section 10.4.2. Has a recurrent connection on the hidden state. This looping constraint ensures that sequential information is captured in the input data. | Sequence data (text, audio, time series) (Not for tabular data, and image data) | (1) Captures the sequential information present in the input data; (2) Can share the parameters across different time steps (called "parameter sharing"), resulting in fewer parameters to train and decreases the computational cost | (1) Not efficient in handling long sequences. Tends to forget the contents of the distant position, and mixes the contents of adjacent positions. (2) Vanishing and exploding gradients |
| Convolutional neural network (CNN) | Section 10.4.3. Consists of filters (i.e., kernels) to extract relevant features from the input data using the convolution operation. | Image data | (1) Able to capture the spatial features from an image, helping to identify the location of an object, as well as its relation with other objects in an image; (2) Implement parameter sharing with a single filter applied across different parts of an input to produce a feature map | Vanishing and exploding gradients |
| Transformer network | Section 10.4.4. Transformers perceives the entire input sequences simultaneously. It employs the attention mechanism. | Allows processing multiple modalities (e.g., images, videos, text and audio) using similar processing blocks. | (1) Able to process entire sequences in parallel, increasing the speed and capacity of sequential deep learning models. (2) "Attention mechanisms" tracks the relations between words across very long text sequences in both forward and reverse directions. | Take time to understand the encoder-decoder stacks |

**10.5. Decision Chart for Selecting Appropriate ML Algorithms**

The reader can Google the topic "Machine learning algorithm selection cheat sheets-images", and can find several dozen, step-by-step, query-and-answer-type decision charts for choosing appropriate ML algorithms. Most of those charts agree on the basic guidelines for algorithm selection, but differ on the specific recommendations for certain applications. Three popular algorithm selection "cheat sheets" are:

(1) Microsoft Azure: https://docs.microsoft.com/en-us/azure/machine-learning/algorithm-cheat-sheet

(2) Scikit-learn: https://scikit-learn.org/stable/tutorial/machine_learning_map/index.htML

(3) SAS- The Power to Know:
https://blogs.sas.com/content/subconsciousmusings/2020/12/09/machine-learning-algorithm-use/ or https://www.reddit.com/r/learnmachinelearning/comments/r5l17z/brief_overview_which_machine_learning_algorithm/

(4) Accel.AI: https://www.accel.ai/anthology/2022/1/24/machine-learning-algorithms-cheat-sheet

Figure 10.39 shows a decision chart adopted from (4) Accel.AI.

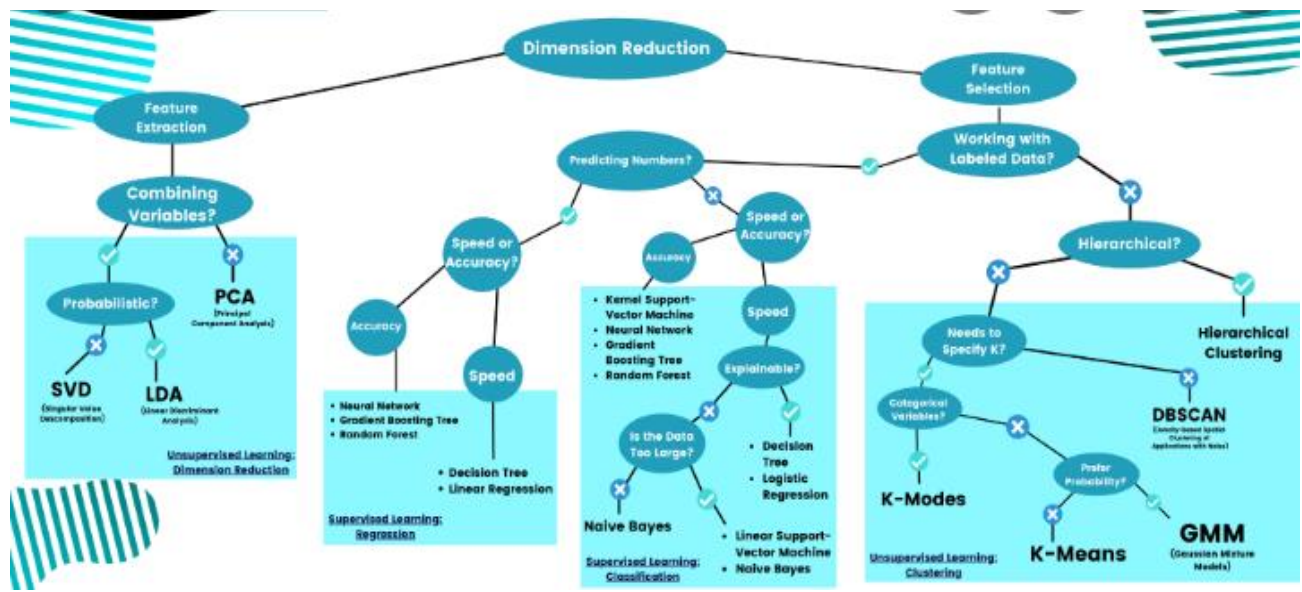

Figure 10.39 An example of a ML algorithm selection decision chart.

**10.6 Workshop 10.1 – Prediction of HDPE Melt Index Using Random Forest and Extreme Gradient Boosting (XGBoost) Ensemble Learning Models**

**10.6.1 Objective and HDPE Process**

The objective of this workshop is to demonstrate the development and application of a predictive model for the HDPE melt index using a random forest ensemble learning model.

We consider a slurry HDPE process at the LG Petrochemicals in South Korea [56] that we illustrated previously in Figure 1.44 with two reactors in parallel. We also showed the simulation flowsheet for one of the two reactors in the process in Figure 9.42.

**10.6.2 Data Collection and Visualization**

Park et. al. [51] correlate the MI data with nine independent variables listed in Table 10.13. We are grateful to Professor Y. K. Yeo, a co-author of [51] for providing us the original data for this workshop. The dataset consists of 5000 observations and 14 main independent variables (X) and one dependent variable (Y), which is the polymer melt index, MI. The complete data appear in an Excel file, **HDPE_LG_Plant_Data.xlsx**, within Workshop 10.1. We choose only 8 independent variables listed in Table 10.13 for the current workshop. Figure 10.40 illustrates the process data varying with time (in minute).

Table 10.13   Process Variables for Workshop 10.1

| Process variable | Description |
|---|---|
| C2 | Monomer ethylene feed flow rate |
| H2 | Chain-transfer agent, hydrogen feed flow rate |
| CAT | Catalyst feed flow rate |
| HX | Hexane solvent feed flow rate |
| C3 | Comonomer propylene feed flow rate |
| T | Temperature of the reactor |
| P | Pressure in the reactor |
| H2/C2 | Feed concentration ratio in the reactor of monomer ethylene to hydrogen |
| C3/C4 | Feed concentration ratio of propylene to 1-butylene monomer |
| MI (dependent or quality variable) | Melt index of polymer |

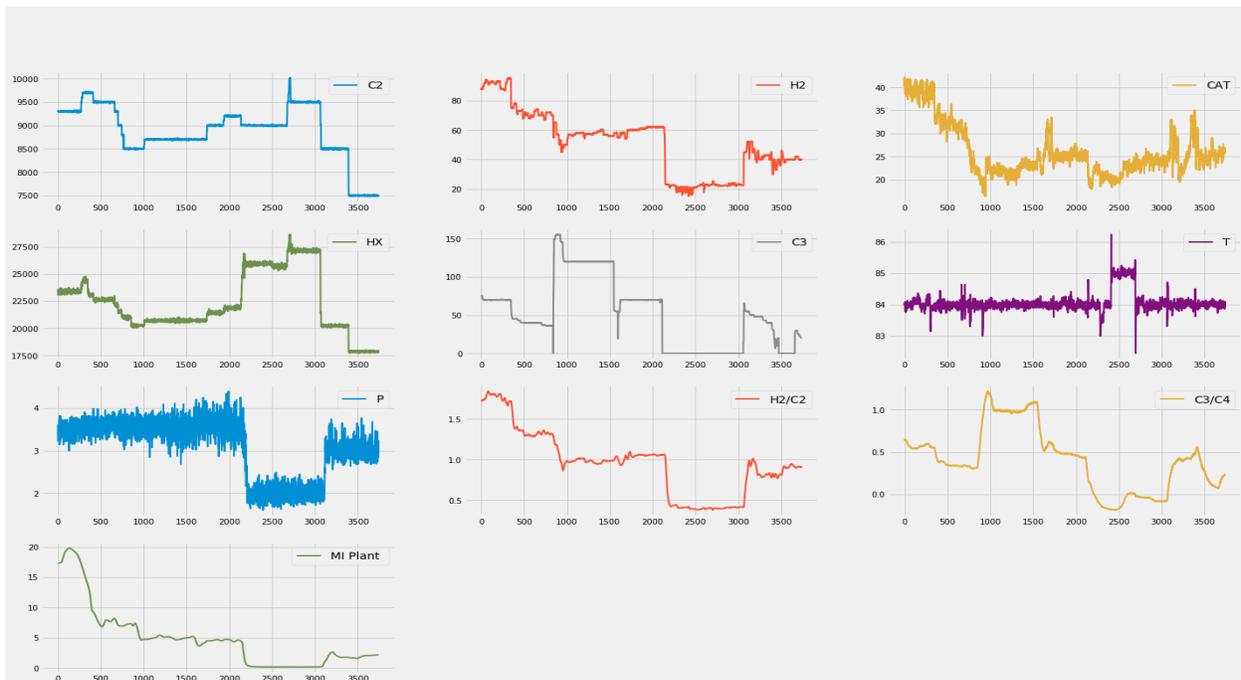

Figure 10.40 Visualization of the HDPE process data

We load the csv data and convert it into data-frames using Pandas library in Python. We then use the Pandas DataFrame to define the independent features/variables (X) and dependent variables (y). Note that Pandas DataFrame is two-dimensional size-mutable, potentially heterogeneous tabular data structure with labeled axes (rows and columns). Figure 10.41 shows the code.

```python
import pandas as pd
df = pd.read_excel('HDPE_LG_Plant_Data.xlsx')
df.head()

#Features
X = df.iloc[:, 1:10]
X.head()

#Dependent Variable
y = df.iloc[:,10]
y.head()
```

Figure 10.41 Defining independent variables (features) and dependent variable

We can also visualize and plot the data using Matplotlib library in Python. See the code in Figure 10.42.

```python
#Data Visualization
p = df.iloc[:,1:11]
p.head()
plt.style.use('fivethirtyeight')
p.plot(subplots=True,
       layout=(6, 3),
       figsize=(22,22),
       fontsize=10,
       linewidth=2,
       sharex=False,
       title='Visualization of the HDPE plant data')
plt.show()
```

Figure 10.42 Visualization of process data displayed in Figure 10.40.

### 10.6.3 Data Cleaning and Preprocessing

In order to handle any missing data, we can remove any of the observations with missing data using some of the functions of Pandas library like: df.dropna( ) .

### 10.6.3a Feature Selection

We can remove any highly correlated variables to make the models more efficient and capture the right causality in the data. We can first calculate the correlation matrix and plot the them. We can use the Seaborn library for plotting the correlation mapping. Seaborn is a data visualization library built on top of matplotlib and closely integrated with Pandas data structures in Python. See Figures 10.43 for the code, and Figure 10.44 for the correlation mapping between all variables.

```
#Feature Selection
corr =X.corr()
import seaborn as sns
sns.heatmap(corr)
```

Figure 10.43 Finding the correlations of the dataset and plotting a correlation map

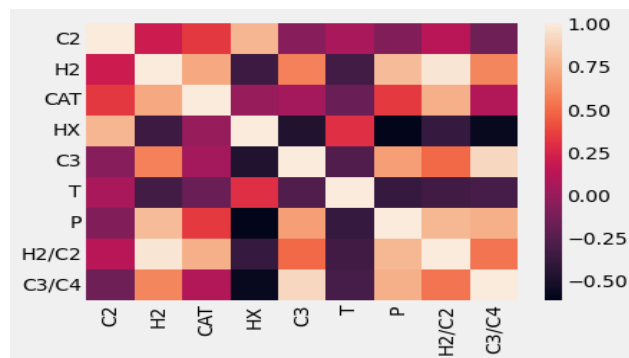

Figure 10.44 Correlation mapping of features (independent variables)

On the right of Figure 10.44, we see the correlation coefficient between two features (independent variables) going from − 0.50 to 1.00.  The figure shows that H2 is highly correlated with C2, having a correlation coefficient close to +1. We can also confirm this by removing those features for which the correlation value is greater than 0.95. See the code in Figure 10.45.

```
#Droppping Highly correlated variables
correlation_matrix = X.corr().abs()
correlated_features = set()

for i in range(len(correlation_matrix .columns)):
    for j in range(i):
        if abs(correlation_matrix.iloc[i, j]) > 0.95:
            colname = correlation_matrix.columns[i]
            correlated_features.add(colname)

print(correlated_features)
```

Figure 10.45 Removing those features (independent variables) for which the correlation coefficient is greater than 0.95.

In a larger dataset, we can drop highly correlated features to improve prediction accuracy; in the current workshop, we retain all features (independent variables).

### 10.6.3.b Defining Training and Evaluation Datasets

We divide the dataset using the Scikit-learn ("sklearn" for short) test train split library in Python. The default test split in the library is 0.25, which means that Python uses 75% of the dataset for training and cross-validation and uses the remaining 25% dataset for testing and evaluation. You may search the Internet for tutorials about train_test_split:

```
X_train, X_test, y_train, y_test = train_test_split(X, y)
```

### 10.6.3.c  Standardization

We then standardize each of the training and test dataset using the sklearn standard-scaler library so that the features and outputs are in the same range and the prediction and model correlations are more accurate. The features are standardized by removing the mean and scaling to unit variance (see Appendix A, Section 1.1.7). See Figure 10.46 for the code.

```
from sklearn.preprocessing import StandardScaler
sc_x = StandardScaler().fit(X_train)
X_train_std = sc_x.transform(X_train)
X_test_std = sc_x.transform(X_test)

y_train = y_train.values.reshape(-1,1)
y_test = y_test.values.reshape(-1,1)
sc_y = StandardScaler().fit(y_train)
y_train_std = sc_y.transform(y_train)
y_test_std = sc_y.transform(y_test)

n_train = np.count_nonzero(y_train)
y_train_std = y_train_std.reshape(n_train,)
```

Figure 10.46 Using the standard-scaler library to standardize the dataset to a zero mean and a unit variance.

### 10.6.4 Build Machine Learning Model

We use the standardized training data to train the ML model. In this case, we consider an ensemble learning random forest model for prediction. We use the sklearn random forest regressor library for model training. We can use cross validation to find the training RMSE. We keep the default hyperparameters (see Table B.1, Appendix B). See Figure 10.47 for the Python code.

```python
#Model Training
from sklearn.ensemble import RandomForestRegressor
rf = RandomForestRegressor(n_estimators=100)

rf.fit(X_train_std, y_train_std)

cv = cross_val_score(rf, X_train_std, y_train_std, cv = 10,scoring='neg_mean_squared_error')
cv_score = cv.mean()

rmse_train = np.sqrt(abs(cv_score))
print(rmse_train)

y_rf_train_std = rf.predict(X_train_std)
y_rf_train_std = y_rf_train_std.reshape(-1,1)
y_rf_train = sc_y.inverse_transform(y_rf_train_std)

rmse_train = np.sqrt(mean_squared_error(y_train, y_rf_train))
print(rmse_train)
```

Figure 10.47 The Python code for ensemble learning random forest regressor.

The training gives R2 = 0.99 and RMSE = 0.09734. The standard deviation of quality variable, y, is 5.
In this case, the default Hyperparameter already performs well, but we will still showcase
hyperparameter tuning below. We can also use cross validation accuracy to prevent overfitting by
finding the optimum hyperparameters that give the highest accuracy on the validation set.
The random forest model is a tree-based model and has many hyperparameters listed below (see also
Table B.1, Appendix B).

- n_estimators = number of trees in the forest
- max_features = max number of features considered for splitting a node
- max_depth = max number of levels in each decision tree
- min_samples_split = min number of data points placed in a node before the node is split
- min_samples_leaf = min number of data points allowed in a leaf node
- bootstrap = method for sampling data points (with or without replacement)

We use grid search with 5-fold cross validation sklearn library to obtain the best hyperparameters. 5-
fold cross-validation involves randomly dividing the dataset into 5 groups (folds) of approximately equal
size. The first group is treated as a test set, and the remaining 4 groups are used to fit the model. We
vary one of the hyperparameters, n_estimators, which represents the number of trees in the forest, to
find the optimum hyperparameter using the cross-validation accuracy as the metric. Figure 10.48 shows
the Python code.

```python
from sklearn.ensemble import RandomForestRegressor
from sklearn.model_selection import GridSearchCV

# Create the parameter grid based on the results of random search
param_grid = {
    'bootstrap': [True],
    'max_depth': [80, 90, 100, 110],
    'max_features': [2, 3],
    'min_samples_leaf': [3, 4, 5],
    'min_samples_split': [8, 10, 12],
    'n_estimators': [100, 200, 300, 1000]
}
# Create a Random Forest model
rf = RandomForestRegressor()
# Instantiate the grid search model

grid_search = GridSearchCV(estimator = rf, param_grid = param_grid,
cv = 3, n_jobs = -1, verbose = 2)

#Find optimum hyperparameters
best_grid = grid_search.best_estimator_

Y_train_best = best_grid.predict(X_train)
rmse_train = np.sqrt(mean_squared_error(y_test, Y_best))
print(rmse_train)
```

Figure 10.48   Using grid search to find optimum hyperparameters

**10.6.5 Analysis of ML Model results**

Figure 10.49shows the code to evaluate the model results. The RMSE of the test dataset with standard deviation is 0.16 and R2 of 0.99 which is a very good prediction for data which has standard deviation of 5.

```python
#Evaluate the results
Y_test_best = best_grid.predict(X_test)
rmse_test = np.sqrt(mean_squared_error(y_test, y_test_best))
print(rmse_test)
```

Figure 10.49 The Python code to evaluate the model result

The random forest ML model also decides the relative importance of different operating variables in reducing the Mean Decrease in "node Impurity" (called MDI), which is a measure of how much each independent (i.e., feature) reduces the variance in the model. Figure 10.50 shows the Python code for finding and plotting the mean decrease in impurity (MDI). Figure 10.51 shows the resulting MDI plot, showing the most important independent variables (features) for prediction of melt index using a random forest regressor model. It indicates that the hydrogen feed flow rate (independent variable or feature) has the most significant impact on the resulting melt index (quality variable) value.

```python
import time
import numpy as np

start_time = time.time()
importances = rf.feature_importances_
std = np.std([
    tree.feature_importances_ for tree in rf.estimators_], axis=0)
elapsed_time = time.time() - start_time

print(f"Elapsed time to compute the importances: "
    f"{elapsed_time:.3f} seconds")

feature_names = [f'feature {i}' for i in range(X.shape[1])]

import pandas as pd
forest_importances = pd.Series(importances, index=feature_names)
fig, ax = plt.subplots()
forest_importances.plot.bar(yerr=std, ax=ax)
ax.set_title("Feature importances using MDI")
ax.set_ylabel("Mean decrease in impurity")
fig.tight_layout()
```

Figure 10.50 The Python code for model evaluation and finding the mean decrease in node impurity (MDI)

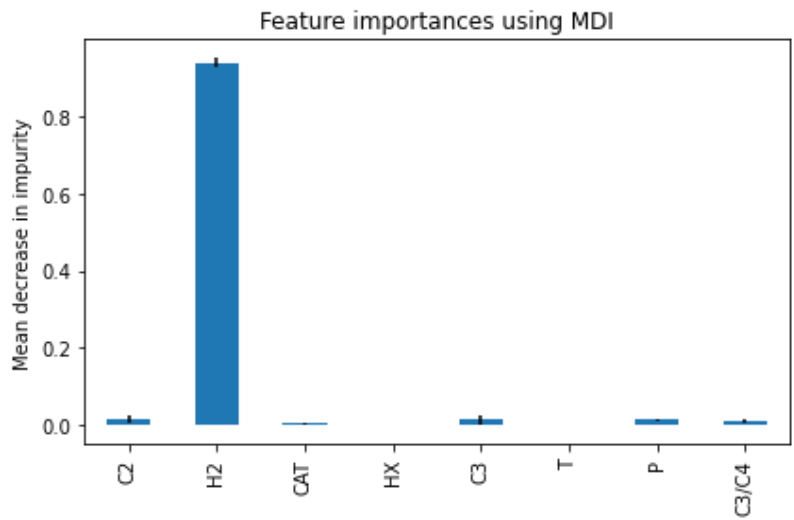

Figure 10.51 Feature (independent variable) importance plot resulting from the random forest regressor.

We can finally plot the final prediction of melt index with time and compare the prediction that with plant data. Figure 10.52 shows the Python code for making the plot, and Figure 10.53 shows the resulting comparison plot. The figure validates the accurate prediction of Random Forest regressor.

```
t = df.iloc[:,0]
import matplotlib.pyplot as plt
plt.style.use('default')
plt.scatter(t,y.iloc[:,1] ,c='green', label = 'Actual' )
plt.plot(t,Y_p[:,1] ,c = 'red', label = 'ML Predicted MI')
plt.xlabel('time')
plt.ylabel('Melt Index')

plt.legend()
plt.show()
```

Figure 10.52 The Python code for comparing the model prediction with plant data for melt index.

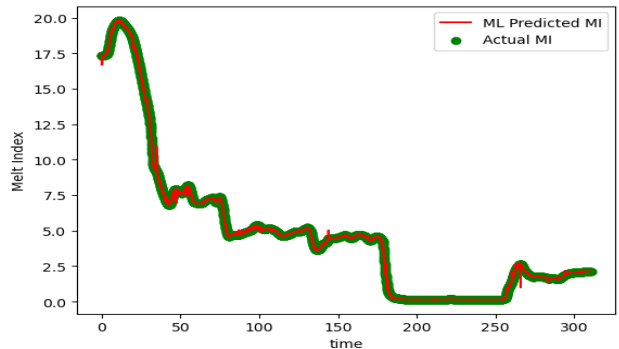

Figure 10.53 Comparison of predicted melt index with plant data resulting from the random forest model.

### 10.6.6 Melt Index Prediction Using XGBoost

We use the same dataset to predict the melt index using XGBoost and follow the same procedure for data preprocessing and feature engineering as in random forest and use the XGBoost ML model for training and prediction.

XGBoost is also a tree-based gradient boosting model, and it has three types of hyperparameters:

- **General parameters** relate to which booster we are using to do boosting, commonly tree or linear model
- **Booster parameters** depend on which booster you have chosen
- **Learning task parameters** decide on the learning scenario. For example, regression tasks may use different parameters with ranking tasks.

For this model, we just use the default hyperparameters for the model. See Figure 10.54.

```
#Model Training
import xgboost as xgb
xgr =xgb.XGBRFRegressor()
xgr.fit(X_train_std,y_train_std)

#Model Prediction
y_xgr_std = xgr.predict(X_test_std)
y_xgr_std = y_xgr_std.reshape(-1,1)
y_xgr = sc_y.inverse_transform(y_xgr_std)
y_xgr = xgr.predict(X_test)
rmse = np.sqrt(mean_squared_error(y_test, y_xgr))
print(rmse)
```

Figure 10.54 Python code to implement XGBoost applied to melt index prediction

The RMSE prediction for the standalone Xgboost model is around 0.25, which is slightly less than the random forest prediction for this dataset.

Similar to Random Forest Xgboost can also give feature importance, which also predicts H2 as the most important feature. See the Python coding in Figure 10.55 and the resulting feature importance plot in Figure 10.56.

```
xgb.plot_importance(xg_reg)
plt.rcParams['figure.figsize'] = [5, 5]
plt.show()
```

Figure 10.55   The Python code to prepare the feature importance plot.

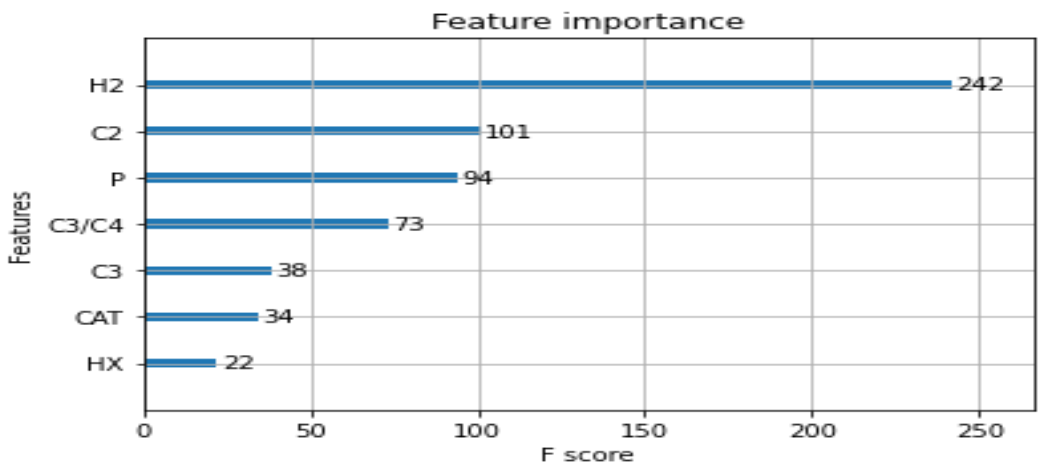

Figure 10.56 The feature importance plot from XGBoost.

This concludes the current workshop.

**10.7 Workshop 10.2 Prediction of HDPE Melt Index Using Deep Neural Networks**

### 10.7.1 Objective

The objective of this workshop is to demonstrate the development and application of a predictive model for the HDPE melt index using deep neural networks.

We consider the same slurry HDPE process at the LG Petrochemicals in South Korea [51] that we illustrated previously in Figure 1.44 with two reactors in parallel. We also showed in Figure 9.42 the simulation flowsheet for one of the two reactors in the process. We demonstrate below how to solve Workshop 10.1 of Section 10.6 using a deep neural network with the Google TensorFlow framework with Keras deep Learning library.

### 10.7.2 Deep Neural Network Configuration

Our deep neural network is similar to that illustrated in Figure 10.57, with has an input layer with 9 nodes, three hidden layers with 64 nodes each, and an output layer with a single node.

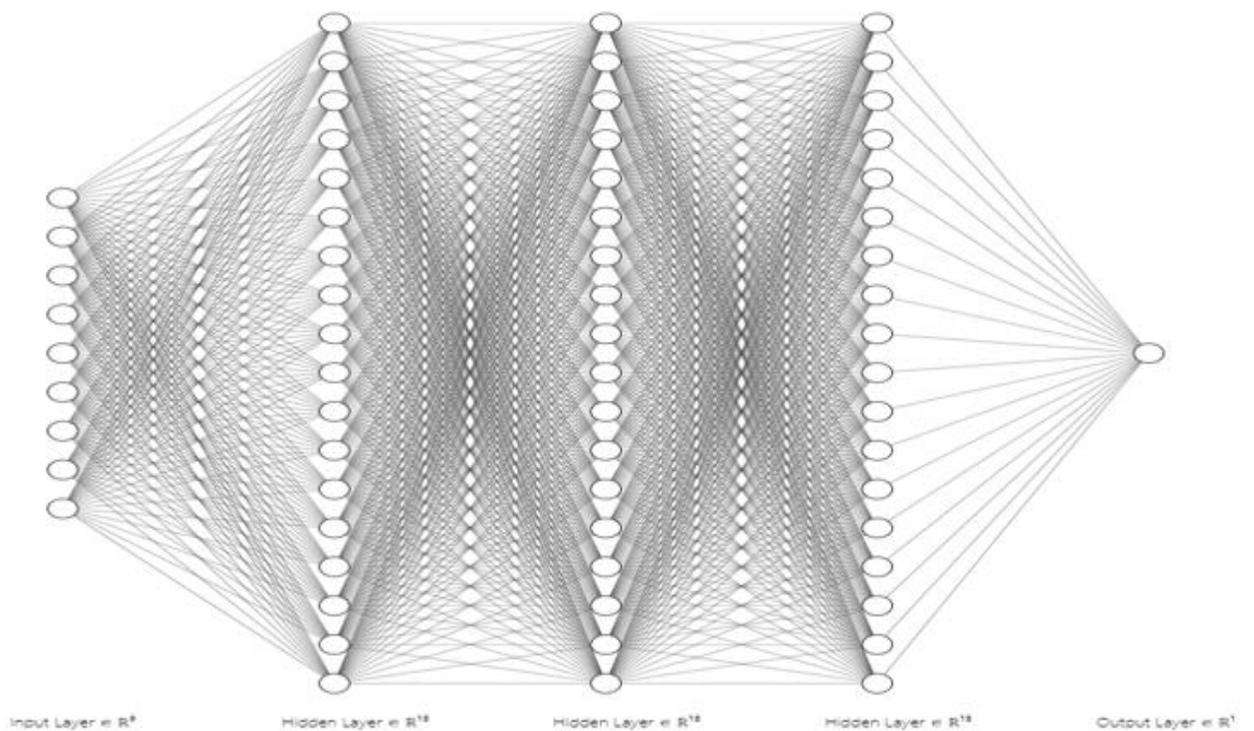

Figure 10.57    An illustration of a deep neural network

We note that the HDPE process flowsheet appears in Figure 9.42; the data file, **HDPE_LP_Plnt_Data.xlsx**, is available in the file for Workshop 10.1; and Table 10.13 gives the list of 9 independent variables (features) for the single quality variable, the melt index. We use the same data prepropossing step described in Section 10.6.3. Specifically, we normalize the dataset both the inputs X and y. Figure 10.58 shows the Python code for data normalization.

```
#Data Normalization

sc_x = StandardScaler().fit(X_train)
X_train_std = sc_x.transform(X_train)
X_test_std = sc_x.transform(X_test)

y_train = y_train.values.reshape(-1,1)
y_test = y_test.values.reshape(-1,1)
sc_y = StandardScaler().fit(y_train)
y_train_std = sc_y.transform(y_train)
y_test_std = sc_y.transform(y_test)
```

Figure 10.58 Python code for data normalization

We use the ReLU activation function of Table 10.6 and Figure 10.24 for the hidden layers and use a linear activation function for the output layer. We use a dense layer type, meaning that each neuron in the dense layer receives inputs from all the neurons of its previous layer. Figure 10.59 shows the Python code to define the deep neural network using Keras. We also use Dropout (Section 10.4.1f) to avoid overfitting with probability of 0.25.

```
model = Sequential()
model.add(Dense(64, input_dim=9, kernel_initializer='normal', activation='relu'))
model.add(Dropout(0.25))
model.add(Dense(64, activation='relu'))
model.add(Dropout(0.25))
model.add(Dense(64, activation='relu'))
model.add(Dropout(0.25))
model.add(Dense(1, activation='linear'))
model.summary()
```

Figure 10.59 The code to implement a deep neural network in Keras

### 10.7.3 Prediction of Neural Network Configuration

We use the Adam optimization of Section 10.4.1g to train the neural network. As this is a regression problem, we choose the MSE, Eq. (10.6), as the loss function to quantify the error between model prediction and data. We then fit the deep neural network to the training data and check the history output for the results. Figure 10.60 shows the Python code to implement this model calculation.

```
#Compiling Model and Prediction
model.compile(loss='mse', optimizer='adam', metrics=['mse','mae'])
history=model.fit(X_train_std,y_train_std, epochs=50, batch_size=150, verbose=1, validation_split=0.2)

#Loss history curve
print(history.history.keys())
# "Loss"
plt.plot(history.history['loss'])
plt.plot(history.history['val_loss'])
plt.title('model loss')
plt.ylabel('loss')
plt.xlabel('epoch')
plt.legend(['train', 'validation'], loc='upper left')
plt.show()
```

Figure 10.60 The code for compiling model prediction and loss curve

Figure 10.61 shows the Python code to compare the model prediction with test data that were not used in developing the prediction model.

```
#Model prediction on test data

y_pred_test_std = model.predict(X_test_std)
y_pred_test = sc_y.inverse_transform(y_pred_test_std)
r2_score(y_test, y_pred_test)
```

Figure 10.61 The code to compare model prediction with test data.

We use inverse transform to scale the predictions to the original scale and then evaluate the predictions. The resulting RMSE, Eq. (10.7), of the melt Index prediction is 0.42 for a dataset with standard deviation of 0.5.

Figure 10.62 illustrates the value of loss function (that is MSE) for training and validation versus the number of iterations (epochs).

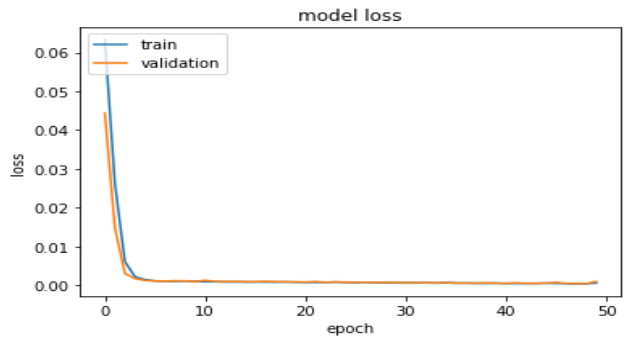

Figure 10.62   Loss function curve of deep neural network for melt index prediction

In Workshop 10.1, Section 10.5, we demonstrate the use of random forest ensemble learning to improve the accuracy and minimize the RMSE of MI prediction. For the present workshop, we can lower the RMSE from a value of 0.42 for our deep neural network to a value of 0.16 using random forest ensemble (Section 10.6.5). By applying the ensemble learning methods, the reader would find that they can generally lower the RMSE of ML models, as reported in prior studies [73 to 76].

## 10.8 Workshop 10.3 Prediction of Time-Dependent HDPE Melt Index Using Dynamic Deep Recurrent Neural Networks

### 10.8.1 Objective

Our objective is to build two different types of deep recurrent neural network structures to predict the melt index for a dynamic process: (1) long short-term memory (LSTM); and (2) gated recurrent unit (GRU). See Sections 10.4.2b to 10.4.2c.

We use the same HDPE dataset used in the previous workshops. The idea here is that we train the data on the model for past time periods and make predictions for a future time. See Sections 10.4.2b to 10.4.2c.

### 10.8.2 Long Short-Term Memory (LSTM) Recurrent Neural Network (RNN)

We use the following steps as a step-by-step workflow to build the network architectures. We use the same preprocess steps as in previous workshops to standardize and normalize the dataset. The only thing different is the way that we define our training, validation, and test sets. We use the observations for the first 1500 min for training and the next 500 min for validation and the next 500 for testing. The training and validation steps are defined below. Since this is a continuous process dataset with no seasonality, we do not need any extra data preparation steps. The input to a LSTM model for TensorFlow needs to be reshaped into a 3D format – [samples, timesteps, features]. For the training dataset, the number of samples are 1500, the features are 9, while the timestep is 1 equal to each observation. Figure 10.63 shows the Python code for the data preprocessing.

```python
sc_x = StandardScaler().fit(X)
X = sc_x.transform(X)
y = y.values.reshape(-1,1)
sc_y = StandardScaler().fit(y)
y = sc_y.transform(y)

# Defining Test, Train and Validation dataset
X_train = X[:1500]
y_train = y[:1500]
X_val = X[1500:1800]
y_val = y[1500:1800]
X_test = X[1800:2400]
y_test = X[1800:2400]

# Reshape into 3D for LSTM input
X_train = X_train.reshape(( X_train.shape[0], 1, X_train.shape[1]))
X_val = X_val.reshape(( X_val.shape[0], 1, X_val.shape[1]))
X_test = X_test.reshape((X_test.shape[0], 1, X_test.shape[1]))
```

Figure 10.63   Data preprocessing and splitting the data into training, validation and test sets for time series prediction by the LSTM/ GRU models

We use a LSTM model with two stacked layers with 64 units and 32 units each and dense layer with 1 unit for output. We also add Dropout to check for overfitting. Figure 10.64 shows the Python code for the LSTM model.

```python
#LSTM Model
model = Sequential()
model.add(LSTM(units = 64,return_sequences = True,input_shape = (1,9)))
model.add(Dropout(0.25))
model.add(LSTM(units = 32,return_sequences = True))
model.add(Dropout(0.25))
model.add(Dense(1))
```

Figure 10.64   LSTM model code

We use the Adam optimizer, and follow by using an inverse transform to reshape and make predictions for the test dataset. Figure 10.65 shows the Python code for the prediction using the LSTM model.

```python
model.compile(loss=MeanSquaredError(), optimizer=Adam(learning_rate=0.0001), metrics=[RootMeanSquaredError()])
history = model.fit(X_train, y_train, validation_data=(X_val, y_val), epochs=10)

#Model predciton
y_pred_std = model.predict(Xest)
y_pred = sc_y.inverse_transform(y_pred_std)
test_predictions = y_pred.flatten()
y_test = sc_y.inverse_transform(y_test)
y_test = y_test.flatten()

test_results = pd.DataFrame(data={'Test Predictions':test_predictions, 'Actuals':y_test})
test_results

import matplotlib.pyplot as plt
plt.plot(test_results['Test Predictions'][1800:2400],c ='green',label = 'LSTM prediction')
plt.plot(test_results['Actuals'][1800:2400],c ='red', label = 'Actual')
plt.xlabel('Time')
plt.ylabel('Melt Index')
plt.legend()
```

Figure 10.65   Code for prediction using the LSTM model.

The LSTM test predictions are plotted and then compared with actual plant data and compared with actual predictions of the melt index as shown in Figure 10.66. The RMSE for the predictions is 1.2.

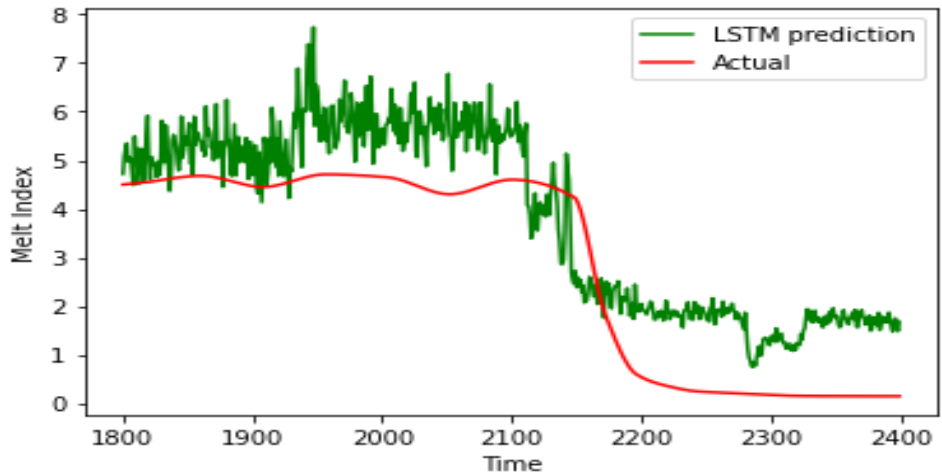

Figure 10.66 Comparison of LSTM model prediction with plant data.

### 10.8.3 Gated Recurrent Unit (GRU)

We repeat the same process using a GRU model in TensorFlow with a similar configuration of 64 and 32 units as shown in Figures 10.67 and 10.68.

```python
#GRU Model
model = Sequential()
model.add(GRU(units = 64,return_sequences = True,input_shape = (1,9)))
model.add(Dropout(0.25))
model.add(GRU(units = 32,return_sequences = True))
model.add(Dropout(0.25))
model.add(Dense(1))
```

Figure 10.67   The Python code for specifying the GRU model

```python
import matplotlib.pyplot as plt
plt.plot(test_results['Test Predictions'][1800:2400],c ='green',label = 'LSTM prediction')
plt.plot(test_results['Actuals'][1800:2400],c ='red', label = 'Actual')
plt.xlabel('Time')
plt.ylabel('Melt Index')
plt.legend()
```

Figure 10.68 The Python code to plot the GRU predictions compared to the actual data

Figure 10.69 compares the GRU model prediction with plan data. The GRU model has a RMSE 0.9 for MI prediction which is better than LSTM predictions as expected. This result confirms the statement by Burkov [7, p. 74] that GRU is the most effective RNN used in practice.

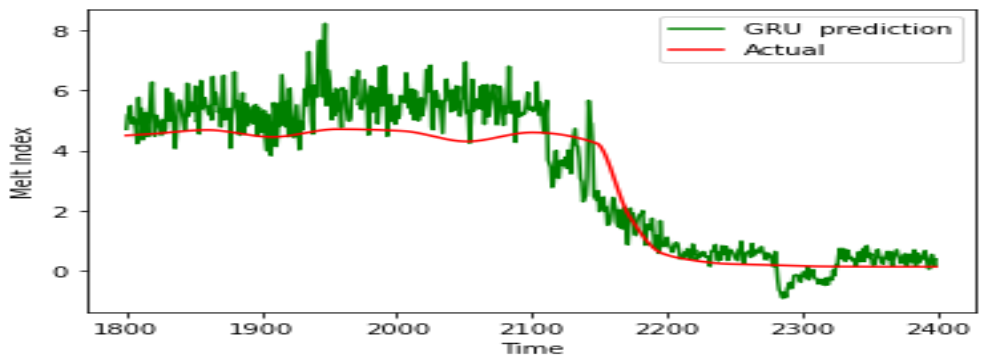

Figure 10.69 A comparison of GRU model prediction and plant data.

## 10.9  Workshop 10.4 Polymer Property Prediction Based on Molecular Structure Using Convolutional Neural Network

### 10.9.1 Objective

The objective of this workshop is to demonstrate how to predict polymer properties, particularly glass transition temperature, based on its molecular structure using a combination of a transformer network (Section 10.4.3) and convolution neural network (CNN) (Section 10.4.2.f). Brownlee [143] has noted that "A CNN or RNN (Section 10.4.2) model is rarely used alone. These types of networks are used as layers in a broader model that also has one or more MLPs (multilayer perceptrons, Section 10.4.1a). Technically, these are a hybrid type of neural network architecture". In this workshop, we illustrate an integration of a transformer network and a CNN.

### 10.9.2 Transformer Network for Deep Learning for Chemical Image Recognition (DECIMER) to Convert a Molecular Structure Image to SMILES Representation

We can convert the molecular structure of a compound to an encoded form of linear strings known as Simplified Molecular Input Line Entry System (SMILES) [122], which can then be used for analysis and prediction of molecular properties based on its structure. In Section 10.1.2b, we briefly introduced SIMLES and its extensions [122 to 124] as an effective molecular representation as inputs to deep neural networks, and SMILES uses a string of characters to represent atoms, bonds, branches, cyclic structures, disconnected structures, and aromaticity with coding rules.

Rajan et. al. [157] showcase a Deep Learning for Chemical Image Recognition (DECIMER) methodology for prediction of SMILES notation from structural image. DECIMER uses a standard transformer network of Figure 10.35 that converts the bitmap of chemical structure depiction into a computer readable format. They use a CNN model to parse the images into extracted features, which is fed into a transformer network. They use four encoder-decoder stacks and eight parallel attention heads (see Section 10.4.3b).

Here, we use the Python library of DECIMER developed by Rajan et. al. [157] and show how to convert a structural image to SMILES notation. We convert the molecular structural image to SMILES using the code below.

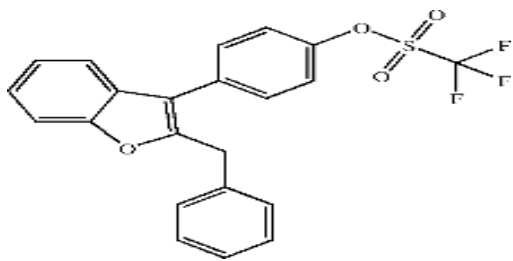

```
from DECIMER import predict_SMILES
image_path = "/content/drive/MyDrive/PolyData/DECIMER_test_0.png"
SMILES = predict_SMILES(image_path)
print(SMILES)

Downloading trained model to /root/.data/DECIMER-V2
/root/.data/DECIMER-V2/models.zip
... done downloading trained model!
C1=CC=C(C=C1)CC2=C(C3=CC=CC=C3O2)C4=CC=C(C=C4)OS(=O)(=O)C(F)(F)F
```

 Figure 10.70   The Python code to convert images to SMILES structure using DECIMER methodology

According to [157], the Python code for DECIMER and the trained models are available at https://github.com/Kohulan/DECIMER-TPU, https://doi.org/10.5281/zenodo.4730515. The data are available as SMILES at: https://doi.org/10.5281/zenodo.4766251.

Thus, using this transformer network, we can generate a database for polymer SMILES encoded structures.

### 10.9.3 Convolutional Neural Network for Predicting Polymer Property Using SMILES Representation

Next, we showcase here the methodology proposed by Miccio and Schwartz [135] to employ chemical structures to predict polymer properties using CNNs. The authors use the SMILES representation of the molecule [122]. The datasets are provided in the Supplement Material of the article (see: https://doi.org/10.1016/j.polymer.2020.122341). These datasets include 218 polymers, their SMILES representation code, and the corresponding glass transition temperatures in K.

In Python, one-hot encoding refers to the representation of categorical variables as binary vectors. These categorical values are first mapped to integer values. Each integer value is then represented as a binary vector that is all 0s (except the index of the integer which is marked as 1). Using one hot-encoding, we can convert the SMILES strings into binary numerical data with matrices with zero and one. We then convert these matrices into binary images, which can then be used for analysis and prediction using CNN. Figure 10.71 illustrates the methodology of converting the SMILES code to an encoded binary image.

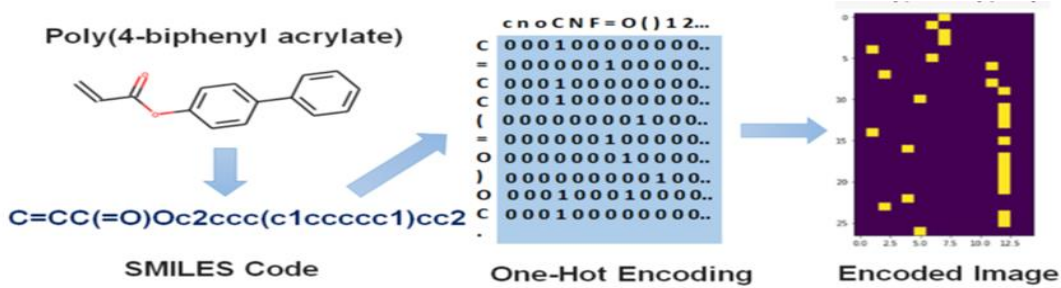

Figure 10.71. The methodology for converting SMILES encoding into a binary image

We first load the dataset using the Pandas library in Python. Figure 10.73 shows a snapshot of dataset, which contains the SMILES structure of polymer and the corresponding glass transition temperature, Tg.

Then, we convert the SMILES dataset to numerical matrix representation using one hot encoding library considering each element of the structure in a dictionary form. The "one hot encoding" then converts it into matrix forms with 0,1 which can also be converted into binary form. The data are converted into numpy input form and the corresponding Tg as the predictions. We then split the dataset into training set and test set.

Figure 10.72 shows the Python code of these operations, and Figure 10.73 illustrates the resulting binary images.

```python
#One Hot encoding of SMILES data
d=[]
n=[['c'], ['n'], ['o'], ['C'], ['N'], ['F'], ['='], ['O'],
        ['('], [')'], ['1'],['2'],['#'],['Cl'],['/'],['S'],['Br']]
e = OneHotEncoder(handle_unknown='ignore')
e.fit(n)
e.categories_
df1=df["SMILES Structure"].apply(lambda x: pd.Series(list(x)))
for i in range(df1.shape[0]):
    x=e.transform(pd.DataFrame(df1.iloc[i,:]).dropna(how="all").values).toarray()
    y=np.zeros(((df1.shape[1]-x.shape[0]),len(n)))
    d.append(np.vstack((x,y)))

# COnverting encoded SMILES to binary images
plt.figure(figsize=(20,100))
for i in range(len(d)):
    plt.subplot(len(d),5,i+1)
    plt.imshow(d[i])

#Dataset
X = np.array(d)
Y=df["Tg"].values

from sklearn.model_selection import train_test_split
X_train, X_test, y_train, y_test = train_test_split(X, Y, test_size=0.15, random_state=0)
```

Figure 10.72   Converting the SMILES code to binary images.

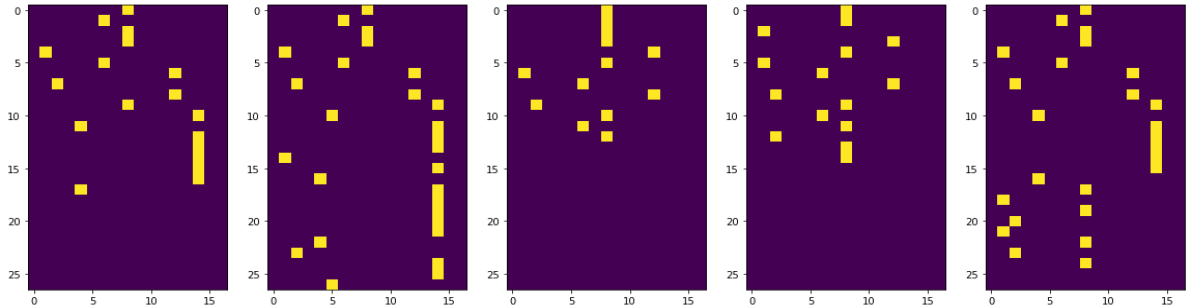

Figure 10.73   The binary images of the dataset.

Figure 10.74 illustrates a schematic of a convolutional neural network used in the original work by Miccio and Schwartz [135]. The reader may refer to our discussion of deep neural networks about ReLU activation (section 10.4.1d), batch normalization (section 10.4.1.e), dropout (section 10.4.1f), Adam optimizer (Section 10.4.1g), kernel size and stride (Figure 10.31), zero padding (Figure 10.32), max pooling and flattening (Figure 10.33), dense or fully connected (FC) layer (Figure 10.34), and CNN block diagram (Figure 10.34).

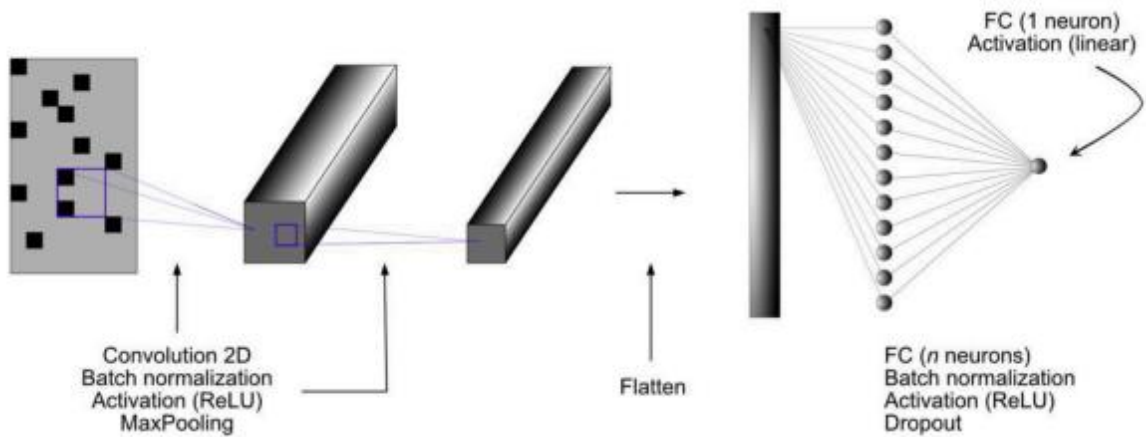

Figure 10.74 A schematic of a convolutional neural network used in Miccio and Schwartz [135]

We analyze the binary images using CNN. The CNN architecture we use has two convolution layers and two dense layers, and an output. The first convolution 2D layer has 128 filters with a kernel size (3,3). The input numpy shape of the dataset is (65,17,1), since this is a binary image data with one channel. The activation function ReLU is used in all layers. The layer is flowed by Max Pool Layers of size (3,3) which reduces size of the inputs to speed computations. The layer is followed by Batch Normalization which makes the training process stable and faster. We add a similar Convolution layer with 64 filters and followed by Max Pool and Batch Normalization layers, and the input is then flattened though a flatten layer, which is followed by two dense layers of 128 and 64 neurons and the 1 output dense layer. We also use dropout tool to prevent overfitting.  Figure 10.75 shows the Python code of this CNN implementation.

```
model=Sequential()
model.add(Conv2D(128,(3,3), activation="relu",input_shape=(65,17,1)))
model.add(MaxPool2D(pool_size=(2,2)))
model.add(BatchNormalization()),
model.add(Conv2D(64,(3,3), activation="relu"))
model.add(MaxPool2D(pool_size=(2,2)))
model.add(BatchNormalization()),
model.add(Flatten())
model.add(Dense(128,activation="relu"))
model.add(Dropout(0.1))
model.add(Dense(64,activation="relu"))
model.add(Dropout(0.1))
model.add(Dense(1))
```

Figure 10.75 The Python code to implement a convolutional neural network

We then compile the model using Adam optimizer, train the model and make predictions. See the Python code in Figure 10.76. The model accuracy for the predictions using structural image is around 82 % for the given CNN configuration. The loss curve for the predictions showcases good accurate predictions as shown in Figure 10.77.

```
model.compile(loss='mse', optimizer='adam', metrics=['mse','mae'])
history=model.fit(x=X_train,y=y_train,epochs=500,batch_size=16,validation_split=0.1)

y_predtrain=model.predict(X_train)
y_predtest=model.predict(X_test)
rmse_test = np.sqrt(mean_squared_error(y_test, y_predtest.reshape(y_test.shape)))
print(rmse_test)

r2_score(y_test, y_predtest.reshape(y_test.shape))
```

Figure 10.76 The Python code to do model optimization and prediction.

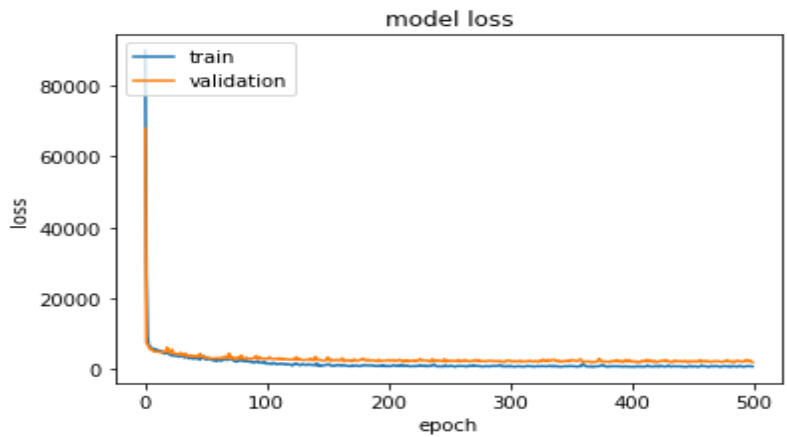

Figure 10.77  Loss curve for CNN training and validation.

Hence, with reasonably accurate predictions, we can use the polymer structure to predict its properties like glass transition temperature using CNN.

**10.10 Workshop 10.5 Melt Index Prediction Using Automated Machine Learning**

The objective of this workshop is to demonstrate the use of an automated ML tool to guide the development of a predictive model for the same melt index prediction problem in Workshop 10.1. We consider the same process dataset as in Workshop 10.1 and perform the same data preprocessing steps.

AutoML is the process of automating the task of ML model development, including feature selection, model selection. It makes the model more scalable and efficient, and is easier for people who have limited knowledge of ML. In this workshop, we use the H2O AutoML library in Python which automates the ML workflow which includes the automatic training and tuning of ML models [129,130]. The model training can be stopped at user defined time.

We import the dataset into H2O DataFrame using Python, and divide it into test and train data. A DataFrame is a two-dimensional, size-mutable, potentially heterogeneous tabular data. Data structure also contains labeled axes (rows and columns). Arithmetic operations align on both row and column labels. The Python code appears in Figure 10.74.

```
#Data loading in H2O
h2o.init()

df = h2o.import_file(path = '/content/drive/MyDrive/PolyData/HDPE_LG_Plant_Data.csv')

df.describe(chunk_summary=True)

train, test = df.split_frame(ratios=[0.8], seed = 1)
```

Figure 10.78 The Python code to import the dataset into H2O DataFrame

We choose the number of models that are evaluated by H2O. The current version of H2O AutoML trains and cross-validates the following algorithms: three pre-specified XGBoost GBM (Gradient Boosting Machine) models, a fixed grid of GLMs (Generalized Linear Models), a default Random Forest (DRF), five pre-specified H2O GBMs, a near-default Deep Neural Net, an Extremely Randomized Forest (XRT), a random grid of XGBoost GBMs, a random grid of H2O GBMs, and a random grid of Deep Neural Nets. Interested readers may refer to the documentation for H2O.ai [105] for details of these algorithms. Figure 10.79 illustrates the Python codes to specify the maximum number of ML models to be tested as 25.

```
#Train AutoML
aml = H2OAutoML(max_models =25,
                balance_classes=True,
                seed =1)

train.head()

aml.train(training_frame = train, y = 'MI_Plant')

preds = aml.predict(test)
preds = aml.leader.predict(test)
```

Figure 10.79 The Python code to specify the maximum number of models to be tested

The AutoML object includes a "leaderboard (lb)" of models that were trained in the process, including the 5-fold cross-validated model performance. Figure 10.76 shows the Python code to display the specific algorithms tested, and Figure 10.77 shows the results.

```
lb = h2o.automl.get_leaderboard(aml, extra_columns = "ALL")
lb

# Get the best model using the metric
m = aml.leader
# this is equivalent to
m = aml.get_best_model()
```

Figure 10.80 The Python code to display the specific ML algorithms tested in the model development

| model_id | rmse | mse | mae | rmsle | mean_residual_deviance | training_time_ms | predict_time_per_row_ms | algo |
|---|---|---|---|---|---|---|---|---|
| StackedEnsemble_BestOfFamily_1_AutoML_1_20220522_183810 | 0.102337 | 0.0104728 | 0.0540606 | 0.0200499 | 0.0104728 | 2064 | 0.13802 | StackedEnsemble |
| StackedEnsemble_AllModels_1_AutoML_1_20220522_183810 | 0.104057 | 0.010828 | 0.0556452 | 0.0203781 | 0.010828 | 4460 | 0.699585 | StackedEnsemble |
| GBM_5_AutoML_1_20220522_183810 | 0.108873 | 0.0118534 | 0.0615105 | 0.0240461 | 0.0118534 | 764 | 0.047608 | GBM |
| GBM_2_AutoML_1_20220522_183810 | 0.117893 | 0.0138986 | 0.0663535 | 0.0226625 | 0.0138986 | 1239 | 0.044936 | GBM |
| XRT_1_AutoML_1_20220522_183810 | 0.119005 | 0.0141621 | 0.0611941 | 0.0223284 | 0.0141621 | 1973 | 0.03827 | DRF |
| GBM_3_AutoML_1_20220522_183810 | 0.120173 | 0.0144416 | 0.0656911 | 0.0246946 | 0.0144416 | 1342 | 0.048857 | GBM |
| XGBoost_grid_1_AutoML_1_20220522_183810_model_14 | 0.121767 | 0.0148272 | 0.0563099 | 0.0222907 | 0.0148272 | 604 | 0.008824 | XGBoost |
| DRF_1_AutoML_1_20220522_183810 | 0.129768 | 0.0168396 | 0.0619152 | 0.0228055 | 0.0168396 | 3032 | 0.031713 | DRF |
| GBM_4_AutoML_1_20220522_183810 | 0.131085 | 0.0171834 | 0.0672527 | 0.0246612 | 0.0171834 | 1447 | 0.055435 | GBM |
| XGBoost_grid_1_AutoML_1_20220522_183810_model_4 | 0.135872 | 0.0184611 | 0.0654633 | 0.0263721 | 0.0184611 | 480 | 0.007541 | XGBoost |

Figure 10.81 Display of results from AutoML

In Figure 10.81, RMSE represents room mean squared error; MSE is mean squared error; MAE is mean absolute error; and RMSLE is root mean squared log error. The figure shows the algorithms used and their RMSE on the test dataset. The results show that as expected, the stacked ensemble model of all the mentioned models gave the most accurate prediction.

We can also use the H2O AutoML tool for help in interpreting the model. We apply the tool, SHAP (Shapley Additive Explanations) proposed by Lundberg and Lee [127]. Figure 10.82 shows a SHAP summary plot. In the plot, the y-axis represents the independent variable (feature), where the variables with the highest importance are placed at the top, and those with the lowest importance are placed at the bottom. For each feature, we see the positive and negative SHAP values of the feature (independent variable). For example, H2/C2 flow rate ratio, the SHAP values are mostly positive between 1 to 9, indicating that H2/C2 has mainly a positive impact on the dependent variable, melt index. Likewise, SHAP values for H2 flow rate are mostly positive between 2 to 4, indicating that H2 has mainly a positive impact on the melt index.

There are some advantages of SHAP-based plotting over traditional importance plots [131]: (1) SHAP-based plots can highlight both the importance of the independent variables (features) and the positive and negative relationships of the independent variables with the dependent variable; (2) the SHAP-based plot includes every single observation as shown by each dot in the plot; and (3) SHAP-based plots work with sparse datasets. Traditional importance plots based on partial least square (PLS) only show the trend on a generalized basis and do not account for individual cases.

Lastly, we mention that there are a growing number of automated machine learning algorithms to help the user identify the right ML tool to deal with feature selection, preprocessing, model development, hyperparameter tuning, etc. This is important as the number of available ML algorithms is getting huge and almost exhaustive. Interested readers may refer to our recent article [132] applying an automated ML algorithm, called TPOT (tree-based pipeline optimization tool) to an industrial fermentation operation.

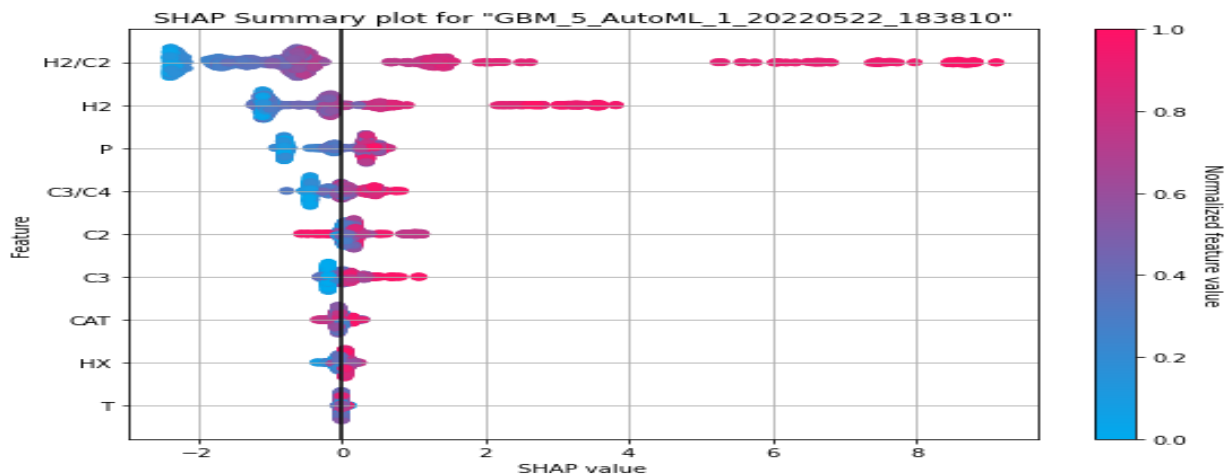

Figure 10.82   SHAP summary plot of the H2O Auto ML predictions

## 10.11. Limitations of Stand-Alone Data-Based Models

The predictive ML models often cannot identify the important features correctly, hence they often cannot quantify the model sensitivities accurately. Based on the knowledge of polyolefin reaction kinetics, we know that melt index (MI) is highly dependent on the hydrogen flow rate and a small change in hydrogen flow leads to a significant change in MI.

Table 10.14 compares the actual plant values of melt index (MI) at different hydrogen flow rate for a first-principle-based model, a causal partial least square (PLS) model, and a data-based ML model. While the ML predictions for the base case (MI = 4.7 and 4.9) are close to the actual plant value of 5, the ML models do not predict MI values accurately at an increased hydrogen flow of 90 m³/hr. By increasing the hydrogen flow rate, we observe much less change in ML prediction compared to a first-principl-based model. We find that the first-principle-based model is better able to predict the change since it is built on accurate reaction kinetics.

We also test the data beyond the operating range at a hydrogen flow of H2 of 180 m$^3$/hr for which actual plant data are not available. We find that that the first-principle-based model and the causal model are more accurate in their extrapolative predictions outside the range of hydrogen flow rate in the original plant data, when compared to a predictive ML model.

Table 10.14. Model comparison based on MI values at varying hydrogen flow

| Model | MI (H2 = 60 m$^3$/hr) | MI (H2 = 90 m$^3$/hr) | MI (H2 = 180 m$^3$/hr) |
|---|---|---|---|
| Actual Plant value | 5 | 17.5 | - |
| First-principle-based model prediction | 5.3 | 20.3 | 56 |
| Data-based model prediction by causal partial least square (PLS) | 4.7 | 12 | 33 |
| Data-based model prediction by ensemble machine learning | 4.9 | 10 | 19 |

In Chapter 11, we explore a hybrid science-guided machine learning (SGML) approach to modeling chemical processes by integrating first-principle-based models with a data-based machine learning models. We show that the integrated approach can significantly improve both the interpolative and extrapolative accuracies of the resulting model.

you have to buy it). See: http://ema.cri-info.cm/wpontent/uploads/2019/07/2019BurkovTheHundred-pageMachineLearning.pdf)

8. Geron. A. (2022). *Hands-On Machine Learning with Scikit-Learn, Keras, and Tensor Flow: Concepts, Tools and Techniques to Build Intelligent Systems*, 3rd edition, O'Rilley, Sebastopol, CA.

9. Marsland, S. (2022) Machine Learning: An Algorithmic Perspective. 2nd edition, Chapman & Hill/CRC Press, Boca Raton, FL.

10. Grus, J. (2019). *Data Science from Scratch: First Principles with Python*. 2nd edition, O'Rilley, Sebastopol, CA.

11. Chollet, F. (2021). *Deep Learning with Python*. Simon and Schuster, New York.

12. Raschka, S.; Mirjalili, V. (2019). *Python Machine Learning: Machine Learning and Deep Learning with Python, Scikit-Learn and Tensor Flow2*. 3rd edition, Packt Publishing, Bermingham, UK.

**(Coding)**

13. Pedregosa, F.; Varoquaux, G.; Gramfort, A.; Michel, V.; Thirion, B.; Grisel, O.; Blondel, M.; Prettenhofer, P.; Weiss, R.; Dubourg, V. (2011). Scikit-Learn: Machine Learning in Python. *Journal of Machine Learning Research*, 12, 2825.

14. Abadi, M.; Barham, P.; Chen, J.; Chen, Z.; Davis, A.; Dean, J.; Devin, M.; Ghemawat, S.; Irving, G.; Isard, M. (2016). *Tensorflow: A System for Large-Scale Machine Learning*, *12th {USENIX} Symposium on Operating Systems Design and Implementation ({OSDI},* pp 265-283.

**(Review Artcles)**

15. Qin, S. L. (2003). Statistucal Process Monitoring: Basics and Beyond. *J. Chemometrics*. **17**, 480.

16. Qin, S. J. (2014). Process Data Analytics in the Era of Big Data. *AIChE Journal*, **60**, 3092.

17. Chemical Engineering Progress (2016). Special Section on Big Data Analytics, March, 27-50.

18. Chiang, L. H.; Lu, B.; Castillo, I. (2017). Big Data Analytics in Chemical Engineering. *Annu. Rev. Chem. Biomol. Eng*., **8**, 63.

19. Ge, Z.; Song, Z.; Ding, S. X.; Huang, B. (2017). Data Mining and Analytics in the Process Industry: The Role of Machine Learning.*IEEE Access,* **5,** 20590.

20. Goldsmith, B. R.; Esterhuizen, J.; Liu, J. X.; Bartel, C. J.; Sutton, C. (2018). Machine Learning for Heterogeneous Catalyst Design and Discovery. *AIChE Journal*, **64**, 2311.

21. Zendehboudi, S.; Rezael, N.; Lohi, A. (2018). Applications od Hybrid Models in Chemical, Petroleum, and Energy Systems: A Systematic Review. *Applied Energy*, **228**, 2539.

22. Lamoureux, P. S.; Winther, K. T.; Torres, J. A. G.; Streibel, V.; Zhao, M.; Bajdich, M.; Abild-Pedersen, F.; Bligaard, T. (2019). Machine Learning for Computational Heterogeneous Catalysis. *ChemCatChem*, **11**, 3581.

23. Qin, S. J.; Chiang, L.H. (2019). Advances and Opportunitues in Machine Learning for Process Data Anayltics. *Computersand Chemical Engineering*, **126,** 465**.**

24. Mater, A. C.; Coote, M. L. (2019). Deep Learning in Chemistry. *Journal of Chemical Information and Modeling,* **59**,245.

25. Venkatasubramanian, V. (2019). The Promise of Artificial Intelligence in Chemical Engineering: Is It Here, Finally? *AIChE Journal*. **65**, 466.

26. Dobbelaere, M. R.; Plehiers, P. P.; Van de Vijver, R.; Stevens, C. V. (2021). Machine Learning in Chemical Engineering: Strengths, Weaknesses, Opportunities and Threats. *Engineering*, **7**, 2101.

27. Qin, S. J.; Guo, S.; Li, Z.; Chaing, L. H.; Castillo, I.; Braun, B. (2021). Integration of Process Energy and Statistical Learning for the Ddow Data Challeneg Problem. *Computers and Chemical Egineering*, **153**, 10741.

28. Trinh, X.; Maimaroglou, D.; Hoppe, S. (2021). Machine Learning in Chemical Product Engineering: The State of the Art and a Guide for Newcomers. *Proceses.* **9**, 1456.

29. Khaleghi, M. K.; Savizi, I. S. P.; Lewis, N. E.; Shojaosadati, S. A.. (2021). Synergisms of Machine

**(Deep Learning with Recurrent Neural Networks - Multilayer Perceptron, MLP; Long Short-Term Memory, LSTM; Gated Recurrent Unit, GRU)**